\documentclass[11pt,letterpaper,twoside,reqno,nosumlimits]{amsart}
\pdfoutput=1

\usepackage{subeqnarray}

\newtheorem{definition}{Definition}[section]
\newtheorem{proposition}{Proposition}[section]

\newtheorem{theorem}{Theorem}[section]
\usepackage{etoolbox}
\usepackage{amssymb}
\patchcmd{\section}{\scshape}{\bfseries}{}{}
\makeatletter
\renewcommand{\@secnumfont}{\bfseries}
\makeatother
\usepackage[toc,page]{appendix}

\usepackage[dvipsnames]{xcolor}
\patchcmd{\section}{\normalfont}{\normalfont\color{MidnightBlue}}{}{}
\patchcmd{\subsection}{\normalfont}{\normalfont\color{MidnightBlue}}{}{}

\makeatletter
\def\subsubsection{\@startsection{subsubsection}{3}%
\z@{.5\linespacing\@plus.7\linespacing}{-.5em}%
{\normalfont\bfseries}}
\makeatother

\usepackage{fancyhdr}
\usepackage{amsmath,amsfonts,amsbsy,amsgen,amscd,mathrsfs,amssymb,amsthm,mathtools,tensor}

\usepackage[a4paper,margin=2.0cm]{geometry}
\usepackage{overpic}

\usepackage{stmaryrd}

\usepackage{amsopn}

\usepackage{array} 
\usepackage{booktabs}
\usepackage{makecell}
\usepackage{caption}
\usepackage{subcaption}
\usepackage{graphicx}
\usepackage{color}
\usepackage{url}
\usepackage{epstopdf}
\usepackage{pdfsync}
\usepackage[colorlinks]{hyperref}
\usepackage{comment}
\usepackage{mathabx}
\usepackage{upgreek}
\usepackage{parskip}

\usepackage{mathrsfs}
\usepackage{cleveref}
\usepackage{yfonts}

\hypersetup{
    linkcolor=blue,
}

\newlength{\fixboxwidth}
\usepackage{epsfig,amsbsy,graphicx,multirow}

\usepackage{ algorithm, algorithmic}
\renewcommand{\algorithmiccomment}[1]{\bgroup\hfill//~#1\egroup}

\usepackage[all,cmtip]{xy}

\usepackage{accents}

 \numberwithin{equation}{section}

\def\RR{\mathbb{R}}

\def\K{\kappa}

\def\restrict#1{\raise-.5ex\hbox{\ensuremath|}_{#1}}

\def\<{\big\langle}
\def\>{\big\rangle}

\def\det{\operatorname{det}}

\def\Hc{\mathcal{H}}

\def\RR{\mathbb{R}}

\definecolor{red}{rgb}{0.9, 0, 0}

\definecolor{green}{rgb}{0.0, 1.0, 0.0}

\makeatletter
\newcommand{\oset}[3][0ex]{%
  \mathrel{\mathop{#3}\limits^{
    \vbox to#1{\kern-2\ex@
    \hbox{$\scriptstyle#2$}\vss}}}}
\makeatother

\makeatletter
\newcommand{\uset}[3][0ex]{%
  \mathrel{\mathop{#3}\limits_{
    \vbox to#1{\kern-2\ex@
    \hbox{$\scriptstyle#2$}\vss}}}}
\makeatother

\usepackage{aurical}
\usepackage[T1]{fontenc}

\usepackage{pgfplots}
\pgfplotsset{compat=1.18}

\usepackage{tikz}
\usetikzlibrary{shadows}
\usetikzlibrary{arrows}
\usetikzlibrary{shapes}

\usepackage[numbers]{natbib}

\begin{document}

\title[Kernel SOS]{Kernel Sum of Squares for Data Adapted Kernel Learning of Dynamical Systems from Data:\\ A global optimization approach}

\author{Daniel Lengyel$^*$,  Boumediene Hamzi$^{**,***}$, Houman Owhadi$^{**}$, Panos Parpas$^*$}
\email[Daniel Lengyel]{d.lengyel19@imperial.ac.uk}
\email[Boumediene Hamzi]{boumediene.hamzi@gmail.com}

\address{$^*$Department of Computing, Imperial College London}
\address{$^{**}$Department of Computing and Mathematical Sciences, Caltech, Pasadena/CA, USA}

\address{$^{***}$The Alan Turing Institute, London, UK}


\begin{abstract}
    This paper examines the application of the Kernel Sum of Squares (KSOS) method for enhancing kernel learning from data, particularly in the context of dynamical systems. Traditional kernel-based methods, despite their theoretical soundness and numerical efficiency, frequently struggle with selecting optimal base kernels and parameter tuning, especially with gradient-based methods prone to local optima. KSOS mitigates these issues by leveraging a global optimization framework with kernel-based surrogate functions, thereby achieving more reliable and precise learning of dynamical systems. Through comprehensive numerical experiments on the Logistic Map, Henon Map, and Lorentz System, KSOS is shown to consistently outperform gradient descent in minimizing the relative-$\rho$ metric and improving kernel accuracy. These results highlight KSOS's effectiveness in predicting the behavior of chaotic dynamical systems, demonstrating its capability to adapt kernels to underlying dynamics and enhance the robustness and predictive power of kernel-based approaches, making it a valuable asset for time series analysis in various scientific fields.
\end{abstract}

\maketitle
 \tableofcontents
\section{Introduction}

The widespread presence of time series data across numerous scientific fields has spurred the development of a wide range of statistical and machine learning methods for forecasting \cite{kantz97,CASDAGLI1989, yk1, yk2, yk3, yk4, survey_kf_ann,Sindy,jaideep1,nielsen2019practical,abarbanel2012analysis}.

Among the various learning approaches, kernel-based methods offer substantial advantages in terms of theoretical analysis, numerical implementation, regularization, guaranteed convergence, automation, and interpretability \cite{chen2021solving, houman_cgc}. Notably, reproducing kernel Hilbert spaces (RKHS) \cite{CuckerandSmale} have established solid mathematical foundations for the analysis of  dynamical systems \cite{5706920, yk1, bhcm11,bhcm1, lyap_bh, bh_sparse_kfs, BHPhysicaD, hamzi2019kernel, bh2020b,klus2020data,ALEXANDER2020132520,bhks,bh12,bh17,hb17,mmd_kernels_bh, boumedienehamzi2022note} and surrogate modeling (cf. \cite{santinhaasdonk19} for a survey) as well as analyzing neural networks \cite{smirnov2022mean}. However, the accuracy of these emulators relies on the choice of base kernel, and insufficient attention has been given to the challenge of selecting an appropriate kernel.

In recent studies conducted by Hamzi and Owhadi and their collaborators, it has been demonstrated that kernel flows (KFs)  \cite{Owhadi19}, a cross-validation technique that can also be viewed as a compression method, can effectively reconstruct the dynamics of chaotic dynamical systems in both regular \cite{BHPhysicaD} and irregular \cite{lee2021learning} time sampling scenarios. The parametric variant of KFs involves utilizing a parameterized kernel function and minimizing the regression relative error between two interpolants represented by the kernel. One interpolant is obtained using all data points, while the other is derived using half of the data points. This approach can be considered as a variation of the cross-validation method. It can also be viewed as a method of data compression using Kolmogorov complexity in the context of Algorithmic Information Theory (AIT) \cite{bh_huter_kfs_ait}.
 Subsequently, several research works have extended the concept of KFs. For instance, a non-parametric version of kernel flows \cite{Owhadi19}, employing kernel warping, has been employed to approximate chaotic dynamical systems in \cite{bhkfnp}. Another version of KFs has been developed for stochastic differential equations (SDEs) \cite{bhkfsdes}, as well as for systems with missing dynamics \cite{bh_kfs_missing} and Hamiltonian dynamics \cite{bh_hamiltonian}. From an application perspective, KFs have been employed in the context of machine learning for classification tasks \cite{yoo2021deep}, and in geophysical forecasting \cite{hamzimaulikowhadi}. A recent version of kernel flows named \emph{Sparse Kernel Flows} \cite{bh_sparse_kfs} has been applied to 133 chaotic dynamical systems  from various fields such as biochemistry, fluid mechanics, and astrophysics.

It is important to note that while the kernel learning algorithms mentioned earlier can derive optimal parameters from data, they still require a base kernel. This limitation makes it challenging for practitioners to select an appropriate kernel function for their specific practical problem. Furthermore, machine learning methods have been successfully applied to various long and short-term prediction tasks, and the choice of the machine learning algorithm should depend on the objective at hand. In previous work, Hamzi and Owhadi  and their collaborators introduced different variants of Kernel Flows (KFs) that cater to specific dynamic objectives. One variant, based on Lyapunov exponents, aims to capture the long-term behavior of the system \cite{BHPhysicaD}. Another variant, based on the Maximum Mean Discrepancy, focuses on capturing the statistical properties of the system and its potential connection to the Frobenius-Perron operator \cite{BHPhysicaD}.  Another variant is based on choosing a kernel that allows to reconstruct attractors and that was named \emph{Hausdorff Metric Kernel Flows} (HMKFs) \cite{hmkfs}.

These versions can be viewed as different approaches to learning dynamical systems from data, each with its own specific "dynamic objective" in mind. In these different versions of KFs, the kernel was learned through local optimization of the corresponding objective functions. In this paper, we consider the problem of \emph{global optimization} of some of these objective functions using kernel sum of squares (kernel SOS) where the objective function is represented as a sum of kernels that could be then represented as a sum of squares in some suitable RKHS.

\section{Statement of the Problem}

Given a time series ${x}_1,\cdots, {x}_n$ from a deterministic dynamical system in $ \mathbb{R}^d$, our goal is to forecast the evolution of the dynamics from historical observations.

A natural solution to forecasting the time series is to assume that the data are sampled from a discrete dynamical system
\begin{equation}
	{x}_{k+1} = f^\dagger({x}_k,\cdots,{x}_{k-\tau^\dagger +1})
\end{equation}
where ${x}_k \in \mathbb{R}^d$ is the state of the system at time $t_k$, $f^\dagger$ represents the unknown vector field and $\tau^\dagger \in \mathbb{N}^*$ represents the delay embedding or delay\footnote{In this paper, we fix $\tau^\dagger$ a priori. Selection strategies of $\tau^\dagger$  are detailed in \cite{BHPhysicaD}),}. 

In order to approximate $f^\dagger$, given $\tau \in \mathbb{N}^*$,  the problem of the dynamical system approximation can be recast as a kernel interpolation  problem 
\begin{equation}
	{Y}_k = f^\dagger({X}_k), ~k = 1,\cdots,N
\end{equation}
with ${X}_k:=\left({x}_{k+\tau-1},\cdots, {x}_k \right)$, ${Y}_k:={x}_{k+1}$ and $N = n-\tau$ for $\tau \in \mathbb{N}^*$ . 

Given a reproducing kernel Hilbert space\footnote{A brief overview of RKHSs is given in the Appendix.} of candidates $\mathcal{H}$ for $f^\dagger$, and using the relative error in the RKHS norm $\|\cdot\|_\mathcal{H}$ as a loss, the regression of the data $({X}_k,{Y}_k)$ with the kernel $K: \mathbb{R}^n \times  \mathbb{R}^n \rightarrow \mathbb{R}$ associated with $\mathcal{H}$ provides a minimax optimal approximation \cite{owhadi_scovel_2019}  of  $f^\dagger$ in $ \mathcal{H}$.  This regressor (in the presence of measurement noise of variance $\lambda>0$) is
\begin{equation}\label{mean_gp}
	f({x}) = K({x},{X})\left(K({X},{X}) + \lambda{I} \right)^{-1}{Y}
\end{equation}
where ${X} = \left( {X}_1, \cdots, {X}_N \right), ~{Y} = \left( {Y}_1, \cdots, {Y}_N \right) $, $K({x},{X})$ is the $N \times N$ matrix with entries $K({x},{X}_i)$, $K({X},{X})$ is the $N \times N$ matrix with entries $K({X}_i,{X}_j)$, ${I}$ is the identity matrix and $\lambda \ge 0$ is a hyper-parameter that ensures the matrix $K({X},{X}) + \lambda{I}$ invertible. This regressor has also a natural interpretation in the setting of Gaussian process (GP) regression:  (\ref{mean_gp}) is the conditional mean of the centered GP $\xi \sim \mathcal{N}(0,K)$  with covariance function K conditioned on $\xi({X}_k) = {Y}_k+\sqrt{\lambda}{Z}_k$ where the ${Z}_k$ are centered i.i.d. normal random variables. 

\subsection{On Kernel Learning}\label{subseckf}
The accuracy of any kernel-based method depends on the kernel $K$.  Here, we follow the parametrized KFs algorithm to learn a "good" kernel in the sense that there will be no significant loss in accuracy if the number of regression points can be halved. The intuition is that if the induced function does not change much when the data is halved, then the kernel induces the appropriate bias for the given function class. Consequently, the induced function should generalize well and be preferred. Notably, this is reminiscent of a cross-validation approach to obtain optimal hyperparameters.

To make this more precise, let  $K_\theta ({x},{x}')$ be a family of kernels parameterized by ${\theta}$, and let $K_\theta ({X},{X})$ be the corresponding gram matrix associated with a vector of points ${X}$. We also denote by $\Theta$ the space of valid parameters for $K$. Finally, let $f_{\theta}^{(X, Y)}$ be the functions induced by the kernel $K_{\theta}$ on data $(X, Y)$. While there has been a range of metrics introduced that follow the above cross-entropy approach, in this paper, we consider the \textit{relative-$\rho$} metric\footnote{Check Appendix B for other variants of KFs and corresponding $\rho$ functions.}. This is because it closely follows the idea of measuring the distance between the induced function on the complete and on the reduced data set. To define $\rho$ we prepare a data vector ${X}^b = \left( {X}_1,\cdots,{X}_N \right) $ and  ${Y}^b=\left( {Y}_1,\cdots,{Y}_N \right) $. We then sub-sample half of the data at random and define this data as ${X}^c$ and ${Y}^c$. Then $\rho$ is given as
\begin{equation} \label{kf_original}
    \begin{aligned}
        {\rho}({\theta}) &= 
        \frac{\|f^b_{\theta}-f^c_{\theta}\|^2_\mathcal{H}}{\|f^b_{\theta}\|^2_\mathcal{H}} 
        = 1 - \frac{{{Y}^c}^\top\left(  K_{\theta}({X}^c,{X}^c) +\lambda{I}\right)^{-1} {Y}^c}{{{Y}^b}^\top \left( K_{\theta}({X}^b,{X}^b)+\lambda{I} \right) ^{-1} {Y}^b}
    \end{aligned}
\end{equation}
which is the squared relative error (in the RKHS norm $\|\cdot\|_{K_\theta}$ defined by $K_\theta$) between the interpolants $f^b_{\theta}$ and $f^c_{\theta}$ obtained from the two nested subsets of the time series. Since we work with no noise in the paper, we assume that $\lambda = 0$.

Finding the minimizer $\theta^*$ of $\rho$ is however not straightforward. For one, $\rho$ is generically not a convex function and can hence have a range of minima which can be difficult to find. Furthermore, evaluating $\rho$ may be expensive when there is a large amount of data to fit. Every evaluation of $\rho$ corresponds to having to produce a new best fit based on $\K_{\theta}$ and the dataset $(X, Y)$. Therefore, we want to find an algorithm that is able to efficiently find good candidate solutions for $\theta^*$. Such an algorithm will have to explore the space to avoid getting stuck at sub-optimal minima and saddle-points.

\subsection{Gradient-based methods}
The most commonly used method to find $\theta^*$ candidates is via gradient-descent based algorithms. The advantage of such methods is the simplicity and the guarantee of converging to some local minimum. However, if the function has a range of sub-optimal minima or saddle-points, it may get quickly stuck at such points. Furthermore, every update step requires an evaluation of $\rho$, potentially making this method prohibitively expensive.

\subsection{Kernel Sum of Squares as a Global Method}
We contrast the gradient-based method with a global optimization method, which promises avoidance of local minima and allows for tighter control of the number of function evaluations. The kernel sum of squares (\emph{KSOS}) is an extension of the classic sum of squares \cite{Lasserre} and solves for a global minimum via the use of kernel-based surrogate functions.  

To make this idea more precise, consider the following optimization problem 
\begin{align*}
    \max_{c \in \mathbb{R}} c \quad \text{such that} \quad \rho(\theta) - c \geq 0 \text{ for } \theta \in \Theta. 
\end{align*}
While this problem is convex, there are uncountably many constraints that need to be satisfied, making it generally intractable. However, under some circumstances, this verification can be feasible.

Assume that for a given $\rho$ and $c$ there exists a non-negative $h_c$ such that $\rho(\theta) - c = h_c(\theta)$. Then trivially, the constraint $\rho(\theta) - c \geq 0$ for $\theta \in \Theta$ is satisfied. An instance where this is feasible is when $\rho$ is given by a polynomial. Then one can use semi-definite programming and the Positivstellensatz to verify the constraint efficiently \cite{stengle1974nullstellensatz, ahmadi2018sum}. When $\rho$ is not a polynomial, this may be difficult when $\rho$ is not well approximated by polynomials as it can lead to instability \cite{cheney2009course}. Instead, one may introduce a similar approximation by using functions where an appropriate kernel provides a more flexible approach adapting to the structure of $\rho$. 

Let $\phi(\theta)$ be a feature representation and introduce the function $\langle \phi(\theta), A \phi(\theta) \rangle$ for some positive-semi definite operator $A$. This represents a rich function class and a low-complexity universal function approximator on all non-negative functions. Hence, if there exists $A$ such that $\rho(\theta) - c = \langle \phi(\theta), A \phi(\theta) \rangle$ for $\theta \in \Theta$ then the constraint is satisfied \cite{marteau2020non}. While this function form is flexible, it remains infeasible to ensure that it is exact for all $\theta \in \Theta$. Hence, the constraint is sub-sampled and enforced on a finite subset of parameters $\mathcal{T} \subset \Theta$ with $\mathcal{T} = \{\theta_i\}_{1 \leq i \leq N^{(SOS)}}$. This is valid as the introduced function is a universal function approximator, that is the accuracy can be improved arbitrarily with more samples \cite{rudi2024finding, marteau2020non}. It then only remains to find $A$ such that $\rho(\theta_i) - c = \langle \phi(\theta_i), A \phi(\theta_i) \rangle$ for $1 \leq i \leq N^{(SOS)}$. However, when an $A$ exists to satisfy the sub-sampled constraints, it is often not unique. To ensure that $\langle \phi(\theta), A \phi(\theta) \rangle$ is sufficiently regular to be a good approximation of $\rho(\theta) - c$, a regularizer on the trace of $A$ is added to the objective function. 

To introduce the optimization problem in \emph{KSOS}, we note that it has been shown that instead of working in the infinite-dimensional space of $\phi(\theta)$, we may work in a finite-dimensional space \cite{rudi2024finding, marteau2020non}. Then, instead of defining an infinite dimensional feature map, it suffices to work with a kernel function. Let $K^{(SOS)}$ be a kernel and the gram matrix over $\mathcal{T}$ be given by $K^{(SOS)}(\mathcal{T}, \mathcal{T}) = R R^T$ with $R$ being upper-triangular. Furthermore, let $\Phi_j$ be the $j$-th column of $R$ and represent the $j$-th feature vector. We then write the \emph{KSOS} optimization problem as 
\begin{align}
    \max_{c \in \mathbb{R}, B \succeq 0} c - \lambda Tr(B) \quad \text{such that} \quad \rho(\theta_i) - c = \Phi_i^T B \Phi_i \text{ for } 1 \leq i \leq N^{(SOS)}, \label{eq:kernelSOS}
\end{align}
for some $\lambda > 0$. An increased $\lambda$ will then lead to more weight on the regularity of the approximation of $\rho(\theta) - c$. Conversely, if $\lambda$ is zero, we have $c = \min_{1 \leq i \leq N} \rho(\theta_i)$ and hence the lowest function value over $\mathcal{T}$ is proposed as optimal. This is because $c \geq \min_{1 \leq i \leq N} \rho(\theta_i)$ and since there is no penalty for $B$ being irregular, it suffices to choose the matrix which achieves this \cite{rudi2024finding}.

We note that the number of function evaluations on $\rho$ is given directly by $N^{(SOS)}$. This can be controlled a-priori and specifies the accuracy of the approximation of $\rho - c$ via $\Phi_i^T B \Phi$. This may be beneficial when $\rho$ is expensive to evaluate due a large data-set, specifically compared to gradient descent where the number of function evaluations until convergence is more difficult to deduce. This does not come without some added complexity, as \emph{KSOS} incurs the cost of solving the optimization problem in Equation \ref{eq:kernelSOS}. However, efficient algorithms exist, and once $\rho$ has been evaluated over $\mathcal{T}$, this complexity is independent of the complexity of evaluating $\rho$ itself \cite{rudi2024finding}.

\section{Numerical Experiments}
We now consider applications in three dynamical systems, specifically the Logistic and H\'enon Map and the Lorentz System. The kernel function we use is given by 
\begin{align*}
    K(x, y) &= \gamma_1^2 \sum_{i = 1}^d \max(0, 1 - \frac{\vert x_i - y_i \vert}{\sigma_1^2}) + \gamma_2^2 e^{- \frac{\Vert x - y \Vert^2}{\sigma_2^2}} + \gamma^2_3 e^{- \frac{\sum_{i = 1}^d \vert x_i - y_i \vert}{\sigma_3^2}} + \gamma_4^2 e^{- \sigma_4^2 \sum_{i = 1}^d \sin^2(\pi \sigma_5^2 \vert x_i - y_i \vert)} e^{- \frac{\Vert x - y \Vert^2}{\sigma_6^2}}.
\end{align*}
This kernel is made up of the triangular, Gaussian, Laplace, and locally periodic kernels. These kernels have been found to capture many physical properties. The goal is then to find parameters that appropriately adapt the kernel to the dynamical system being considered. The $\gamma$ parameters weigh each kernel appropriately in the linear combination, and the $\sigma$ parameters capture the appropriate length scale. The domain for the parameters is given by $\Theta = [0.001, 10]^{\otimes 10}$.

\subsection{Algorithmic Setup}
To find a good candidate parameter $\theta$, the function $\rho$ needs to be initialized first. For this, we fix a starting position $x_0^{(train)}$ and specify the number of steps $N^{(train)}$ for the training trajectory $x^{(train)} = \{x_t^{(train)}\}_{0 \leq i \leq N^{(train)}}$. This will make up the full data set $(X^b, Y^b)$, where $X^b_{t} = x^{(train)}_{t - 1}$ and $Y^b_t = x^{(train)}_{t}$ since each dynamical system only requires the current state to compute the next. 

We evaluate each optimization method using ten random seeds. For each random seed, we randomly subset half of the data $(X^b, Y^b)$ to obtain $(X^f, Y^f)$ and hence $\rho$. For the given random seed, the same $\rho$ is used for the gradient descent and \emph{KSOS} to keep comparisons fair. Lastly, we allocate a budget of $200$ function evaluations of $\rho$ to each method to better reason about the method complexity. 

\subsubsection{Gradient Descent}
We then run gradient descent for $200$ steps with a step-size $\eta$ starting each parameter at one. This is the baseline parameter value for the kernel and keeps the method consistent. As we have found the performance very sensitive to the learning rate, we present the best learning rate for the problem over a hyperparameter search in $\{10^{-4}, 10^{-3}, 10^{-2}, 10^{-1}, 0.5, 1\}$. Note that for some problems, it appears that the method is making little progress, mostly due to scale imbalance. Specifically, the \emph{KSOS} algorithm begins with a significantly worse starting point and then quickly improves. A larger learning rate would often lead to erratic results, hence settling for the one presented. The source of randomness then comes from $\rho$, as we randomly subset the full data.

\subsubsection{Kernel Sum of Squares}

We now characterize the optimization problem in Equation \ref{eq:kernelSOS} and discuss how to solve it. To obtain the subset $\mathcal{T}$, we use the budget of $200$ function evaluations to sample uniformly at random from $\Theta$. This is also the sole source of randomness of this method. Then, we set the regularizer to $\lambda = 10^{-5}$. Lastly, the feature map is given by a Gaussian kernel with $\sigma = 0.1$. These values were chosen as they performed well across our experiments.

While the optimization problem in Equation \ref{eq:kernelSOS} can be solved efficiently by standard solvers, we choose to use the interior point algorithm presented by Rudi et al. as it allows for parallelization and provides a clearer understanding of how the optimization is achieved \cite{rudi2024finding}. For the treatment, we let $n = N^{(SOS)}$. To enforce that $B$ is positive semi-definite, a log barrier \footnote{The log-barrier $\log \det (B)$ enforces that if an optimization algorithm starts with $B \succeq 0$ then $B$ will remain positive definite. Recall, for a symmetric matrix the determinant is the product of its eigenvalues and hence $\det(B) > 0$. To obtain a negative eigenvalue, an optimization algorithm has to cross a point where an eigenvalue is zero and hence $\det(B)$ is zero. This leads $\log \det(B)$ to approach negative infinity and causes a maximization algorithm to avoid such regions \cite{wright2001convergence, nemirovski2004interior}.} is added to the objective and weighted by a precision term $\epsilon$
\begin{align*}
    \max_{B \succeq 0, c \in \mathbb{R}} c - \lambda \text{Tr}(B) + \frac{\epsilon}{n} \log \det(B) \quad \text{such that} \quad f(x_i) - c - \Phi^T_i B \Phi_i = 0, \text{ for } 1 \leq i \leq n. 
\end{align*}
Under this formulation, we know that the optimal solution is at most $\epsilon$ away from the optimal value given in Equation \ref{eq:kernelSOS} \cite{nemirovski2004interior}. Recalling that $K^{(SOS)} = \Phi \Phi^T$ we have by standard Lagrange duality that 
\begin{align*}
    &\sup_{B \succeq 0, c} \inf_{\alpha \in \mathbb{R}^{n}} c + \sum_{i = 1}^{n} \alpha_i (f(x_i) - c - \Phi_i^T B \Phi_i ) - \lambda \text{Tr}(B) + \frac{\epsilon}{n} \log \det(B) \\
    &= \inf_{\alpha \in \mathbb{R}^{n}} \sum_{i = 1}^{n} \alpha_i f(x_i) - \frac{\epsilon}{n} \log \det (\Phi^T \text{Diag}(\alpha) \Phi + \lambda I) + \frac{\epsilon}{n} \log(\frac{\epsilon}{n}) - \epsilon \text{ s.t } \alpha^T \mathbf{1} = 1,
\end{align*}
where the last step follows from differentiating the first equation and setting to zero to remove the dependence on $B$ and $c$. To solve this minimization over the dual-variables $\alpha$ the following Newton Algorithm is deployed. Let $H(\alpha)$ be the objective function given in the above minimization problem. The Damped Newton algorithm is then given by the update step $\alpha_{t + 1} = \alpha_t - \frac{1}{1 + \sqrt{\frac{n}{\epsilon}} \lambda(\alpha)} \Delta$ where $\Delta = H''(\alpha)^{-1} H'(\alpha) - \frac{\mathbf{1}^T H''(\alpha)^{-1} H'(\alpha)}{\mathbf{1}^T H''(\alpha)^{-1} \mathbf{1}} H''(\alpha)^{-1} \mathbf{1}$ and $\lambda(\alpha)^2 = \Delta^T H''(\alpha) \Delta$ is the Newton decrement. Note, the update direction $\Delta$ is both the Newton update step $H''(\alpha)^{-1} H'(\alpha)$ and the correction term to ensure that the constraint $\alpha^T \mathbf{1} = 1$ is satisfied at each step. 

For this algorithm, we have found that a precision of $\epsilon = 10^{-6}$ and $100$ iterations of the algorithm are sufficient to provide a strong convergent solution. To obtain a candidate for $\theta^*$ we use $\sum_{i = 1}^{n} \alpha_i \theta_i$. We note that even with increased precision $\epsilon$, this only provides a candidate for Equation \ref{eq:kernelSOS}. A more principled way is provided in Section 7 of Rudi et al. and relies on replacing $c$ in Equation \ref{eq:kernelSOS} by a quadratic function \cite{rudi2024finding}. Nevertheless, we have found it to work well numerically and retain it as it keeps the method more simple.


\subsection{Evaluation}

Even when a good candidate solution for $\theta^*$ is found, it remains to confirm that this truly leads to an improved kernel for the estimation of the dynamical system. Hence, we compare the optimization methods in both how well they minimize $\rho$ and how well kernels induced by the candidate solutions perform as estimators for the forward map in the dynamical system.

Assessing the performance in minimizing $\rho$ is simple, as we simply consider the value achieved by the candidate solution given by the optimization algorithm. Notice here that the loss of \emph{KSOS} may at times increase as it optimizes over surrogates of $\rho$ rather than directly on the function itself and hence does not directly translate.


To assess the induced kernel performance, we compare predicted trajectories to true trajectories using three measures. The true trajectories are based on both the training trajectory $x^{(train)}$ and a test trajectory $x^{(test)}$, which has a different starting point to $x^{(train)}$.

To introduce the measures, let $x^{(pred)}$ be the predicted trajectory and $x^{(true)}$ the true trajectory. The first is the \textbf{Mean Error} (ME), specifically we write $\frac{1}{N} \sum_{i = 1}^{N^{(true)}} \Vert f_{\theta}(x^{(true)}_{t - 1}) - x^{(true)}_t \Vert$. This measures the one-step error and is reminiscent of the error assessment of supervised learning algorithms. Note, if $x^{(true)}$ is given by the training trajectory, the error should be zero as the function $f_{\theta}$ interpolates all the training points. Then we introduce the \textbf{Hausdorff Distance} (HD), which measures the largest distance from one trajectory to another. To ensure that this is a symmetric function, first introduce the one-sided metric given by $HD_1(X, X') = \max_{1 \leq i \leq N} \min_{1 \leq j \leq N} \Vert X'_i - X_j \Vert$. We then let $HD = \max\left(HD_1(x^{(pred)}, x^{(true)}), HD_1(x^{(true)}, x^{(pred)}) \right)$. Notice that in the continuous case, the definition of $HD_1$ would have sufficed, but due to the trajectories being discrete, symmetry needs to be enforced. The final metric assesses the number of iterations until the trajectories diverge more than some predefined amount. Specifically, we let $\textbf{Deviation($\gamma$)} = \min \{t :  \frac{\Vert x^{(pred)}_t - x^{(true)}_t \Vert}{\Vert x^{(true)}_t \Vert} \geq \gamma \}$. We chose to normalize by the size of the current point on the trajectory as it provided the most consistent results for the specific dynamical systems considered here.

The results are presented as summary statistics over the range of random runs. Specifically, in the summary statistic tables, we provide the median value and the $25$th and $75$th percentile values in brackets. For intuition, we also visualize the results across three figures by considering the first random run of each optimization method. The first figure presents loss values over the optimization period. Note that gradient descent and \emph{KSOS} have different lengths. This is because gradient descent executes $200$ iterations on $\rho$ directly, while the sum of squares method obtains $200$ samples and then performs $100$ IPM steps. As noted earlier, the seeming lack of progress in gradient descent is largely due to the different scales on which both methods operate. The second figure focuses on the distance to the true trajectory. The last figure visualizes the predicted trajectories and provides the true trajectory as a reference.

\subsection{Logistic Map}
For the first dynamical system, we consider the logistic map given by the following form exhibiting chaotic behavior
\begin{align*}
    x_{t + 1} &= 4 x_{t} (1 - x_{t}).
\end{align*}
For both the training and test trajectories, we let the number of steps be $200$. To construct the training set, we let $x_0^{(train)} = 0.1$, and for the test set, we let $x_0^{(test)} = 0.3$. We set the learning rate of gradient descent to $\eta = 10^{-3}$.

We make the following observations. As seen in Table \ref{tab:rhoLog}, the median performance is nearly a magnitude better for \emph{KSOS}. Given the percentile values, the \emph{KSOS} is then mostly biased to lower values, even though, on some occasions, it can perform as poorly as the gradient descent. As seen in Table \ref{tab:testLog}, the mean error for \emph{KSOS} is also a magnitude better on the test trajectory. The \emph{KSOS} also stays accurate for more steps compared to gradient descent. Notably, the Hausdorff distance is both small and comparable for both methods. While the trajectories quickly diverge, this happens as all trajectories, due to their chaotic behavior, cover the interval $[0, 1]$. Hence, for every point on a trajectory, there exists a point on another trajectory that, while not temporally, is spatially close. In Table \ref{tab:trainLog}, we see that the methods perform mostly similarly on the training trajectory. They have close to zero loss on the mean error, which is expected as they interpolate the points.

\begin{table}[H]
    \centering
    \begin{tabular}{l r r} 
        \toprule
        \textbf{Measure} & \textbf{KSOS} & \textbf{Gradient Descent}  \\
        \midrule
         Relative $\rho$ & $1.89 \ [0.49, 12.20] \times 10^{-3}$ & $9.16 \ [1.69, 15.24] \times 10^{-3}$ \\ 
        \bottomrule
    \end{tabular}
    \caption{Relative $\rho$ Statistics for the Logistic Map.}
    \label{tab:rhoLog}
\end{table}

\begin{table}[H]
    \centering
    \begin{tabular}{l r r} 
        \toprule
        \textbf{Measure} & \textbf{KSOS} & \textbf{Gradient Descent}  \\
        \midrule
        Mean Error & $2.08 \ [1.73, 2.46] \times 10^{-5}$ & $3.88 \ [1.78, 4.41] \times 10^{-4}$ \\ HD & $2.03 \ [1.73, 2.54] \times 10^{-2}$ & $1.92 \ [1.88, 2.29] \times 10^{-2}$ \\ Deviation(0.1) & $10.00$ & $4.00$ \\ Deviation(0.25) & $10.00$ & $8.00 \ [4.00, 8.00]$ \\   
        \bottomrule
    \end{tabular}
    \caption{Test Set Statistics for the Logistic Map.}
    \label{tab:testLog}
\end{table}

\begin{table}[H]
    \centering
    \begin{tabular}{l r r} 
        \toprule
        \textbf{Measure} & \textbf{KSOS} & \textbf{Gradient Descent}  \\
        \midrule
        Mean Error & $5.29 \ [2.63, 14.96] \times 10^{-14}$ & $1.99 \ [0.60, 4.26] \times 10^{-14}$ \\ HD & $1.60 \ [1.45, 1.64] \times 10^{-2}$ & $1.73 \ [1.56, 2.48] \times 10^{-2}$ \\ Deviation(0.1) & $3.80 \ [3.80, 4.30] \times 10^{1}$ & $3.80 \ [3.80, 4.33] \times 10^{1}$ \\ Deviation(0.25) & $4.05 \ [3.80, 4.45] \times 10^{1}$ & $4.05 \ [3.80, 4.45] \times 10^{1}$ \\ 
        \bottomrule
    \end{tabular}
    \caption{Training Set Statistics for the Logistic Map.}
    \label{tab:trainLog}
\end{table}

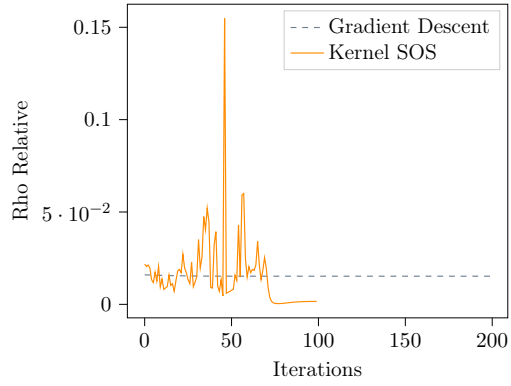
\begin{figure}
\begin{minipage}[c]{0.4\textwidth}
    \centering
    \resizebox{\textwidth}{!}{
\begin{tikzpicture}

\definecolor{darkgray176}{RGB}{176,176,176}
\definecolor{darkorange}{RGB}{255,140,0}
\definecolor{lightgray204}{RGB}{204,204,204}
\definecolor{slategray}{RGB}{112,128,144}

\begin{axis}[
legend cell align={left},
legend style={fill opacity=0.8, draw opacity=1, text opacity=1, draw=lightgray204},
tick align=outside,
tick pos=left,
x grid style={darkgray176},
xlabel={Iterations},
xmin=-9.95, xmax=208.95,
xtick style={color=black},
y grid style={darkgray176},
ylabel={Rho Relative},
ymin=-0.0073107379033316, ymax=0.162487189260518,
ytick style={color=black}
]
\addplot [semithick, slategray, dashed]
table {%
0 0.015982116368717
1 0.0159425930384705
2 0.0159046673780283
3 0.0158685637989715
4 0.0158343749946958
5 0.0158020926200646
6 0.0157716385774341
7 0.0157428922693011
8 0.0157157119107169
9 0.0156899498425424
10 0.0156654627311648
11 0.0156421178410733
12 0.015619796507403
13 0.0155983957243372
14 0.0155778285267837
15 0.0155580236304149
16 0.0155389246315863
17 0.0155204889480225
18 0.0155026866009766
19 0.0154854988874371
20 0.0154689169623683
21 0.0154529403361576
22 0.0154375752924569
23 0.0154228332357488
24 0.0154087289924817
25 0.0153952790999174
26 0.0153825001328883
27 0.0153704071266352
28 0.0153590121603315
29 0.0153483231625362
30 0.0153383429899662
31 0.0153290688169795
32 0.0153204918503568
33 0.015312597361977
34 0.0153053650113059
35 0.0152987694101175
36 0.0152927808696941
37 0.0152873662675314
38 0.0152824899666257
39 0.0152781147319003
40 0.0152742025944828
41 0.0152707156288588
42 0.0152676166197749
43 0.015264869607986
44 0.0152624403124761
45 0.0152602964360384
46 0.0152584078649435
47 0.0152567467773173
48 0.0152552876750374
49 0.0152540073554768
50 0.0152528848370707
51 0.0152519012512089
52 0.0152510397121313
53 0.015250285173358
54 0.0152496242779558
55 0.0152490452083479
56 0.015248537539347
57 0.0152480920977627
58 0.0152477008305444
59 0.0152473566817181
60 0.0152470534799016
61 0.01524678583581
62 0.015246549049108
63 0.0152463390255677
64 0.01524615220232
65 0.0152459854820299
66 0.0152458361737771
67 0.0152457019418132
68 0.0152455807596269
69 0.0152454708701407
70 0.0152453707505787
71 0.0152452790816011
72 0.0152451947208202
73 0.0152451166791989
74 0.015245044100659
75 0.0152449762447141
76 0.0152449124706741
77 0.0152448522242585
78 0.0152447950264191
79 0.0152447404628213
80 0.015244688175307
81 0.0152446378543392
82 0.0152445892322357
83 0.0152445420778885
84 0.0152444961913581
85 0.0152444513999682
86 0.0152444075546062
87 0.0152443645262913
88 0.015244322203715
89 0.0152442804906621
90 0.0152442393039512
91 0.0152441985719206
92 0.0152441582324401
93 0.0152441182320334
94 0.0152440785243384
95 0.0152440390694536
96 0.0152439998328512
97 0.0152439607847294
98 0.0152439218993173
99 0.0152438831543852
100 0.0152438445307385
101 0.015243806011718
102 0.0152437675830445
103 0.0152437292324373
104 0.015243690949024
105 0.0152436527237407
106 0.0152436145484935
107 0.0152435764165029
108 0.0152435383217628
109 0.0152435002591685
110 0.0152434622242236
111 0.0152434242132018
112 0.0152433862227223
113 0.0152433482499416
114 0.0152433102924155
115 0.0152432723479697
116 0.0152432344148102
117 0.0152431964913811
118 0.0152431585761982
119 0.0152431206681762
120 0.0152430827662707
121 0.0152430448695633
122 0.0152430069773257
123 0.0152429690888355
124 0.0152429312036259
125 0.0152428933211513
126 0.0152428554409862
127 0.015242817562757
128 0.015242779686175
129 0.0152427418109342
130 0.0152427039368328
131 0.0152426660636865
132 0.0152426281911997
133 0.0152425903193224
134 0.0152425524479068
135 0.0152425145768543
136 0.0152424767060534
137 0.0152424388354125
138 0.0152424009648684
139 0.0152423630942831
140 0.0152423252237548
141 0.0152422873530703
142 0.015242249482345
143 0.015242211611422
144 0.0152421737403746
145 0.0152421358690624
146 0.0152420979975229
147 0.0152420601257895
148 0.0152420222536911
149 0.0152419843813835
150 0.0152419465087819
151 0.0152419086358451
152 0.0152418707626333
153 0.0152418328890518
154 0.0152417950151679
155 0.0152417571409234
156 0.0152417192663407
157 0.0152416813914753
158 0.0152416435161995
159 0.0152416056406148
160 0.0152415677646464
161 0.0152415298883438
162 0.0152414920116134
163 0.0152414541345706
164 0.0152414162571795
165 0.0152413783793963
166 0.0152413405012685
167 0.0152413026227957
168 0.0152412647438922
169 0.015241226864668
170 0.0152411889850451
171 0.015241151105039
172 0.0152411132247113
173 0.0152410753439728
174 0.0152410374628809
175 0.0152409995814222
176 0.0152409616996043
177 0.0152409238173778
178 0.0152408859348181
179 0.015240848051841
180 0.0152408101685212
181 0.0152407722848289
182 0.0152407344007428
183 0.0152406965162926
184 0.0152406586315003
185 0.0152406207462887
186 0.0152405828607283
187 0.0152405449747812
188 0.0152405070884749
189 0.0152404692017651
190 0.0152404313147176
191 0.0152403934272964
192 0.0152403555394893
193 0.0152403176513264
194 0.015240279762771
195 0.0152402418738239
196 0.0152402039845363
197 0.0152401660948984
198 0.0152401282048179
199 0.0152400903143834
};
\addlegendentry{Gradient Descent}
\addplot [semithick, darkorange]
table {%
0 0.0217975642929923
1 0.0205019254353611
2 0.0212776337517492
3 0.0197365610113931
4 0.0132721389676083
5 0.011803886293961
6 0.0173861553971829
7 0.0125944422259529
8 0.0205475922775171
9 0.0094657930800911
10 0.0136193002987909
11 0.00811885081542973
12 0.00886851936438771
13 0.00953684044777126
14 0.0158457866879315
15 0.0102645356740453
16 0.0112275996017235
17 0.00711587800566882
18 0.0128000842373189
19 0.0180396175406353
20 0.0189234561476453
21 0.0171226628104371
22 0.0270585603922066
23 0.0196024023987517
24 0.0170287195538585
25 0.0132682937162567
26 0.011223688797836
27 0.0228373643102097
28 0.00957459973120167
29 0.0120145082406181
30 0.015061237203807
31 0.0350099125567985
32 0.0196654510290984
33 0.025573479482627
34 0.0475194721418852
35 0.0404234703462255
36 0.0520327514635368
37 0.0445390935318469
38 0.00915389537690359
39 0.00879301167925373
40 0.0324599094902552
41 0.0393506333694857
42 0.00990806493875396
43 0.00708100459734018
44 0.0133697384773847
45 0.00450865278142898
46 0.154769101662161
47 0.00600359662333505
48 0.00656973650536541
49 0.00699205353797427
50 0.00765027109458138
51 0.00808772531574176
52 0.015732559124007
53 0.0130111842205654
54 0.0429901188274568
55 0.014778393854791
56 0.0591954091794105
57 0.0600152002064103
58 0.0246089631778336
59 0.0147587095286009
60 0.0203781793135837
61 0.0171907736753167
62 0.0189051859412703
63 0.0184014451460198
64 0.0213053643868287
65 0.034160547752188
66 0.0209335551570881
67 0.0134525512540871
68 0.0184545904633414
69 0.0253179188151645
70 0.0195811797054984
71 0.00932081999355161
72 0.0037988984773587
73 0.00164050476369637
74 0.000841022520651769
75 0.000538104891746327
76 0.000428459341971221
77 0.000407349695025205
78 0.000436391597628316
79 0.000499097099845192
80 0.000585573440398224
81 0.000687469534583429
82 0.0007969189103747
83 0.000906935233845441
84 0.00101202368893116
85 0.00110848278043296
86 0.00119431888206878
87 0.00126890851487693
88 0.00133257797785624
89 0.0013862186132847
90 0.00143099088794729
91 0.00146812431841203
92 0.00149879748980541
93 0.00152407601886806
94 0.00154488825743981
95 0.00156202334557598
96 0.00157614110568283
97 0.00158778717078467
98 0.00159740949457854
99 0.00160537418680473
};
\addlegendentry{Kernel SOS}
\end{axis}

\end{tikzpicture}}
    \label{fig:Logistic_rho_0_idx}
\end{minipage}
\caption{Rho Relative on Logistic Map Training Set.}
\label{fig:Logistic_rho}
\end{figure}

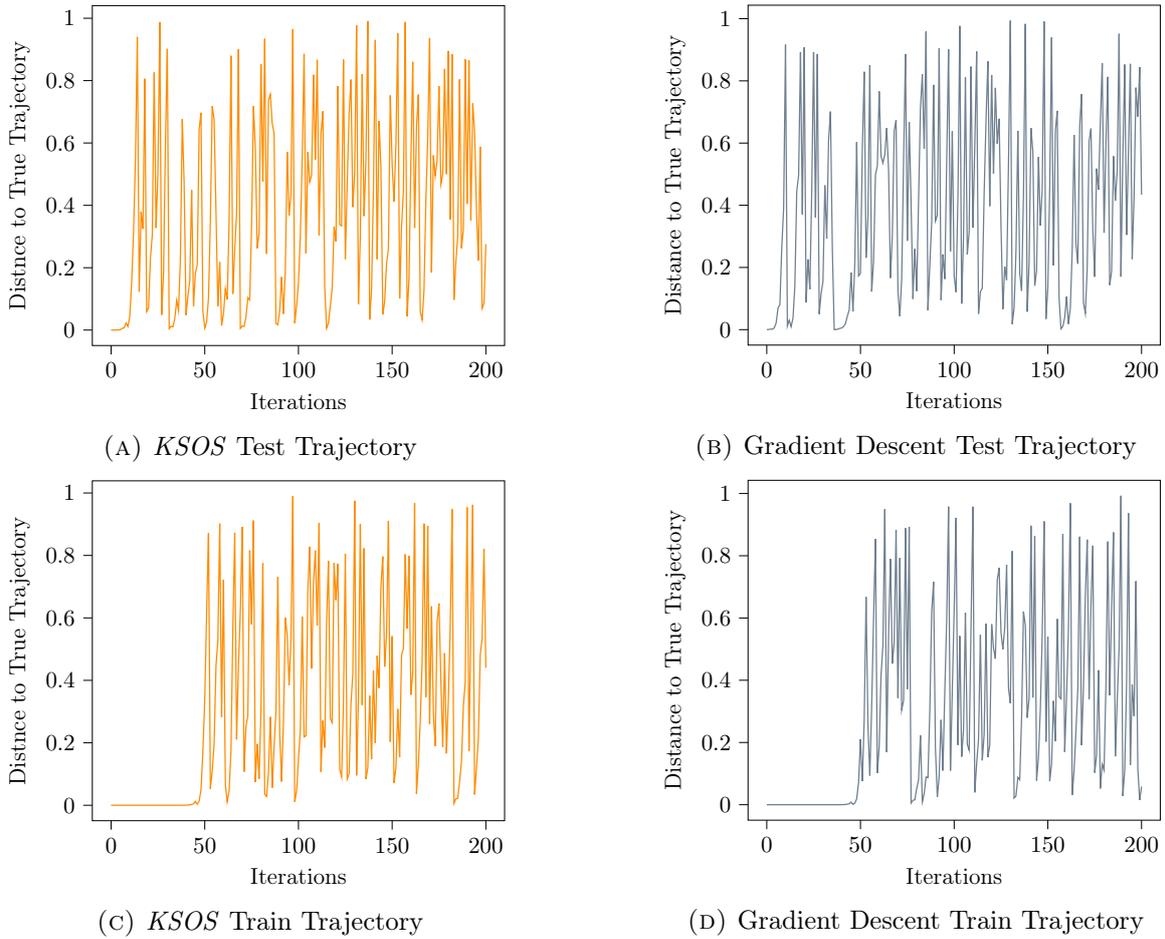
\begin{figure}
\begin{minipage}[c]{0.40\textwidth}
    \centering
    \resizebox{\textwidth}{!}{
\begin{tikzpicture}

\definecolor{darkgray176}{RGB}{176,176,176}
\definecolor{darkorange}{RGB}{255,140,0}

\begin{axis}[
tick align=outside,
tick pos=left,
x grid style={darkgray176},
xlabel={Iterations},
xmin=-10, xmax=210,
xtick style={color=black},
y grid style={darkgray176},
ylabel={Distnce to True Trajectory},
ymin=-0.0495457455599837, ymax=1.04046065675966,
ytick style={color=black}
]
\addplot [semithick, darkorange]
table {%
0 0
1 2.0391550246357e-05
2 5.3514845992364e-05
3 0.000134023953693552
4 0.000432589468092232
5 0.00163465245844845
6 0.00538056949030064
7 0.00758813343943787
8 0.0227890726334114
9 0.0116276329267541
10 0.0449488286590097
11 0.156100121095785
12 0.319448223169796
13 0.546612734506901
14 0.939991456651826
15 0.123223650264554
16 0.379191637910294
17 0.324682817973525
18 0.806228043857852
19 0.0594073726357769
20 0.0714525186300943
21 0.232110538005362
22 0.305712550970759
23 0.825931217341033
24 0.327855768212802
25 0.503545982379317
26 0.987262016804771
27 0.0487759617409511
28 0.185288802439835
29 0.599282997495621
30 0.901779091040521
31 0.00476543776006466
32 0.0119535796563637
33 0.0101665385573035
34 0.0368381460041471
35 0.0951070593757357
36 0.0623365123620723
37 0.231494513025071
38 0.67607622050124
39 0.46852741355878
40 0.0482921987881312
41 0.108104543068104
42 0.159539685272483
43 0.449477227646149
44 0.077261725398553
45 0.183023099250746
46 0.209777352665302
47 0.645096232227732
48 0.696921087451043
49 0.0607441144057401
50 0.00669898465912533
51 0.0265584270965314
52 0.10248798173921
53 0.353458994021417
54 0.717530854275418
55 0.674498300294954
56 0.378174331883465
57 0.0768373023821914
58 0.217845149656485
59 0.0143686903047994
60 0.0519407106851879
61 0.132205069105924
62 0.0977143416960091
63 0.350503273660414
64 0.879670051507971
65 0.116271423232996
66 0.255712146686491
67 0.376727054018796
68 0.900824528149139
69 0.0051892783150857
70 0.0129025812059284
71 0.0116629153915787
72 0.0419575661215734
73 0.103748235149351
74 0.0963590256380911
75 0.335581390287765
76 0.717621508751842
77 0.583089808381754
78 0.262315144849121
79 0.309231697307253
80 0.851832519693623
81 0.476273542900464
82 0.93412981482789
83 0.244636259940373
84 0.737608815634854
85 0.755532036536907
86 0.66276702657376
87 0.630596339114823
88 0.020984733514659
89 0.0171643541564964
90 0.0628444174151572
91 0.170091040621027
92 0.0520666714389423
93 0.19381073957352
94 0.571064704820227
95 0.367246980529543
96 0.434860323366739
97 0.965409343824455
98 0.0221613184214938
99 0.0765487034447514
100 0.151413204252369
101 0.302348066302394
102 0.551148807085855
103 0.885805135900779
104 0.246559472480338
105 0.571002561951249
106 0.475010265065879
107 0.496684966460203
108 0.818661279705825
109 0.546921018740725
110 0.866763243350732
111 0.304413786345168
112 0.630713799927726
113 0.701528276927332
114 0.139618280385531
115 0.00600832746077351
116 0.023101976053362
117 0.0782161657355524
118 0.135609297284917
119 0.331975845713448
120 0.284262988673991
121 0.782189182553093
122 0.338307215285576
123 0.334148544035603
124 0.867385420842814
125 0.227297795052722
126 0.466573651024629
127 0.690492633801901
128 0.80331532975432
129 0.394361404795671
130 0.505882562769254
131 0.97785938146042
132 0.0824025334740526
133 0.301084096982105
134 0.821325658827404
135 0.36573962486672
136 0.546887753671956
137 0.990914911199674
138 0.0339066029080531
139 0.130882797652748
140 0.450600311829148
141 0.930349115320742
142 0.227085543110771
143 0.670968430161485
144 0.521617440009514
145 0.0501758043305888
146 0.0912797114162981
147 0.212282998794165
148 0.260716861463407
149 0.752244799049285
150 0.531303377966662
151 0.412505837637155
152 0.594140537687045
153 0.951484813946931
154 0.101214741888663
155 0.32873440619781
156 0.446211960117424
157 0.987996826456249
158 0.0436255027753626
159 0.16626242555703
160 0.544371785251724
161 0.860182023765489
162 0.327685752526631
163 0.652685465048944
164 0.755284799951803
165 0.058558066593964
166 0.0331869403792289
167 0.126543410993304
168 0.41434835452479
169 0.616886851720894
170 0.936572752498916
171 0.184443689114322
172 0.560127299970429
173 0.494807157862723
174 0.544222570035949
175 0.781706231333681
176 0.468249357336075
177 0.499967232836288
178 0.837501380485065
179 0.500229101394373
180 0.895230810197478
181 0.355178339450085
182 0.884468305334328
183 0.0971000292684513
184 0.21986349257013
185 0.29924414749273
186 0.804086862916806
187 0.262545734335443
188 0.321821474760686
189 0.868045299115071
190 0.404756991038949
191 0.864855115479804
192 0.351613646883577
193 0.726614297085737
194 0.636313180674501
195 0.3863555097511
196 0.222703548932676
197 0.587860088707562
198 0.0701248197402354
199 0.0860155041308267
200 0.275763204674251
};
\end{axis}

\end{tikzpicture}}
    \subcaption{\emph{KSOS} Test Trajectory}
    \label{fig:Logistic_errTraj_sos}
\end{minipage}\hspace{0.1\textwidth}
\begin{minipage}[c]{0.40\textwidth}
    \centering
    \resizebox{\textwidth}{!}{
\begin{tikzpicture}

\definecolor{darkgray176}{RGB}{176,176,176}
\definecolor{slategray}{RGB}{112,128,144}

\begin{axis}[
tick align=outside,
tick pos=left,
x grid style={darkgray176},
xlabel={Iterations},
xmin=-10, xmax=210,
xtick style={color=black},
y grid style={darkgray176},
ylabel={Distance to True Trajectory},
ymin=-0.0496829078106538, ymax=1.04334106402373,
ytick style={color=black}
]
\addplot [semithick, slategray]
table {%
0 0
1 0.000567752574187619
2 0.00148237111130867
3 0.00144093997701111
4 0.00569655992400052
5 0.0216298172834406
6 0.0695037781050197
7 0.080098554121024
8 0.263764077046183
9 0.390148902707946
10 0.916380087429021
11 0.011169386021202
12 0.0303386050027608
13 0.009477498106643
14 0.0373651390776447
15 0.141209845446035
16 0.444538421810411
17 0.49595097157714
18 0.891748029432856
19 0.370745058502299
20 0.907663798041562
21 0.08732771390572
22 0.225874066274931
23 0.129941621401163
24 0.445190621697433
25 0.892660239692161
26 0.360896037646841
27 0.886491088206013
28 0.0502119041472241
29 0.115072989928848
30 0.15566785326164
31 0.463837277707922
32 0.292804808047626
33 0.608959869130349
34 0.701159167103653
35 0.25926373750811
36 6.0341262572261e-05
37 0.00038072420140095
38 0.00136323672243832
39 0.00425986528809552
40 0.00514122046386356
41 0.0102627938326656
42 0.0191398247990704
43 0.0431743411499979
44 0.0627549342017689
45 0.183631351284324
46 0.0584676049673279
47 0.215748515121559
48 0.603633794033066
49 0.173197832527038
50 0.181199390625786
51 0.578621870934186
52 0.829212996732786
53 0.231027430390483
54 0.361021797750882
55 0.850657908335103
56 0.122458832232966
57 0.220363583901035
58 0.498273301818331
59 0.525734451602146
60 0.766215097775103
61 0.557250405280694
62 0.535683557275823
63 0.564161798587165
64 0.648205568927313
65 0.514331835566652
66 0.166071018679882
67 0.303949780136211
68 0.638310879176327
69 0.673876469422304
70 0.123239258586506
71 0.04289202926158
72 0.163671845641817
73 0.539329449371345
74 0.886415732856136
75 0.285739722142134
76 0.668007902976126
77 0.410219240792261
78 0.0990111843337027
79 0.259582051071392
80 0.123994849983731
81 0.414645774701106
82 0.710657288561367
83 0.821381618316568
84 0.58166905986982
85 0.958638287539841
86 0.062140583934213
87 0.208518728915399
88 0.345153740031459
89 0.787092700014002
90 0.349993416858507
91 0.36799163879235
92 0.904668178550869
93 0.0950721501251981
94 0.24191867130331
95 0.162988573952655
96 0.538531548673159
97 0.901508610254042
98 0.251220717348276
99 0.638150056478267
100 0.171251219462791
101 0.120991577409458
102 0.425113384397596
103 0.976782546121153
104 0.0835843813789466
105 0.303953568972455
106 0.810928803186608
107 0.241654118455707
108 0.315383778425993
109 0.845999629748182
110 0.327789912614833
111 0.591614982009183
112 0.894646704495604
113 0.0509403341768072
114 0.122416924901966
115 0.133281322326304
116 0.437913462621871
117 0.674843888371297
118 0.862776945486917
119 0.397367299890375
120 0.818911750725218
121 0.501861655955073
122 0.777670156360674
123 0.598304167940852
124 0.678247215652996
125 0.335239210897996
126 0.0656853787948171
127 0.202561975600058
128 0.159739645801713
129 0.536631458340861
130 0.993658156213076
131 0.0179603570484079
132 0.0700271738295814
133 0.252359947346723
134 0.639120731600046
135 0.181092552258679
136 0.12497859139648
137 0.436671270825291
138 0.9832256888273
139 0.0577646122735778
140 0.215771079389709
141 0.64835447390353
142 0.573138347553283
143 0.141574132964446
144 0.188932863553315
145 0.555217526728564
146 0.334629613419138
147 0.452775356691071
148 0.990822022432434
149 0.0346724823490484
150 0.133626656959412
151 0.459156891566212
152 0.939583758818971
153 0.206490862818835
154 0.637274483255419
155 0.703582592616285
156 0.0997620544574863
157 0.00270944759971159
158 0.0105881527202544
159 0.0387678955724874
160 0.106473663269809
161 0.0181893666900775
162 0.0706454992929553
163 0.253270951762637
164 0.62588402909097
165 0.275422265033517
166 0.211880565837731
167 0.656229488819404
168 0.757841155802069
169 0.0871138190861964
170 0.052299507296134
171 0.196384950843968
172 0.605769784967155
173 0.646587513414755
174 0.318545731898264
175 0.170013398366404
176 0.517929317148034
177 0.450219199806382
178 0.664570092769607
179 0.85675618708607
180 0.311642201825346
181 0.603896225406097
182 0.812922710030278
183 0.143459901625245
184 0.186792033779208
185 0.558070273405323
186 0.414565693886988
187 0.510136688308623
188 0.951393937954036
189 0.170301718842924
190 0.554725085106064
191 0.852536766487931
192 0.304586741785495
193 0.572066839297823
194 0.854676191754069
195 0.227188661891681
196 0.426612536877347
197 0.778156451258776
198 0.685137688207613
199 0.843971067109538
200 0.433807773178172
};
\end{axis}

\end{tikzpicture}}
    \subcaption{Gradient Descent Test Trajectory}
    \label{fig:figs/Logistic_errTraj_flow}
\end{minipage}

\vspace{0.5cm}

\begin{minipage}[c]{0.40\textwidth}
    \centering
    \resizebox{\textwidth}{!}{
\begin{tikzpicture}

\definecolor{darkgray176}{RGB}{176,176,176}
\definecolor{darkorange}{RGB}{255,140,0}

\begin{axis}[
tick align=outside,
tick pos=left,
x grid style={darkgray176},
xlabel={Iterations},
xmin=-10, xmax=210,
xtick style={color=black},
y grid style={darkgray176},
ylabel={Distnce to True Trajectory},
ymin=-0.0494734048303188, ymax=1.03894150143669,
ytick style={color=black}
]
\addplot [semithick, darkorange]
table {%
0 0
1 6.10622663543836e-16
2 1.22124532708767e-15
3 2.94209101525666e-15
4 4.88498130835069e-15
5 1.15463194561016e-14
6 7.32747196252603e-15
7 2.62706523201928e-14
8 7.97140131680862e-14
9 6.26165785888588e-14
10 2.3589463715723e-13
11 6.64690524843081e-13
12 2.33146835171283e-14
13 9.2465206498471e-14
14 3.68809583567442e-13
15 1.44840910099031e-12
16 5.74831848787483e-12
17 2.226231005098e-11
18 7.80135400724191e-11
19 1.67374336612625e-10
20 2.84473999911938e-10
21 7.2787831317811e-10
22 5.26458210359237e-10
23 1.96996724488585e-09
24 5.88614629259965e-09
25 2.73956957119736e-09
26 1.0709289073596e-08
27 3.81697498608169e-08
28 9.08056758675002e-08
29 1.07000737492591e-07
30 3.52875744458192e-07
31 5.078075535625e-07
32 1.506064597534e-06
33 6.05176447843192e-07
34 2.37306225991246e-06
35 8.72799786799661e-06
36 2.41950559345394e-05
37 3.80757408435795e-06
38 1.52538713451096e-05
39 6.1108899882905e-05
40 0.000232469945117053
41 0.000747921289284914
42 0.000880560222824123
43 0.00291242412110421
44 0.00427915480706198
45 0.012494954180129
46 0.00328333843965378
47 0.0129328949437451
48 0.0499042047828482
49 0.172083336038083
50 0.338118439823811
51 0.619690604925273
52 0.871217484272798
53 0.052616263678791
54 0.109120731895439
55 0.199358237349036
56 0.44573666521888
57 0.527069805687458
58 0.902044073111199
59 0.282411433764823
60 0.722525156004248
61 0.0643669794715684
62 0.0115963869219723
63 0.0458175657706584
64 0.174348443770078
65 0.567386212365617
66 0.872810430072302
67 0.210692794228113
68 0.447404748770438
69 0.621996377610418
70 0.890857045861575
71 0.107534859754869
72 0.253385524126515
73 0.286603850714235
74 0.81586890710082
75 0.578398313124605
76 0.911796460853417
77 0.0738636087618272
78 0.196070236858519
79 0.0850585226424458
80 0.306091660475762
81 0.774927073755488
82 0.0351828563622704
83 0.0283144157645505
84 0.103787253789112
85 0.28291164362541
86 0.0561283698190277
87 0.187458607058947
88 0.300479199570352
89 0.731430969845368
90 0.230379165171044
91 0.0759182015833542
92 0.267320953288889
93 0.600176637175035
94 0.544934049916978
95 0.384782814816955
96 0.529107698774459
97 0.989468096606375
98 0.0115685952350735
99 0.0443428887675149
100 0.148382180088608
101 0.239499156211669
102 0.603864818441958
103 0.218869587768073
104 0.222608577192255
105 0.689983745811059
106 0.827998463031721
107 0.437827985841823
108 0.711347724138067
109 0.815610717671042
110 0.575393771466841
111 0.903723603376142
112 0.106599705454637
113 0.270836558243712
114 0.184587361399042
115 0.587159216449666
116 0.781939303322979
117 0.277741786794139
118 0.265108202242908
119 0.77605865993445
120 0.657237099729344
121 0.773428330764794
122 0.114549298475708
123 0.0912398812755764
124 0.326444356909974
125 0.805331755162218
126 0.0841366366371675
127 0.10045766323164
128 0.324536789744952
129 0.421624687826365
130 0.975382071229051
131 0.0955503050960675
132 0.345448965911399
133 0.901092479666571
134 0.321425812793729
135 0.8227258550812
136 0.0840867134174654
137 0.119461685824134
138 0.350562489983015
139 0.147164270580013
140 0.430978449128623
141 0.198735552197402
142 0.478473183270411
143 0.375809296675104
144 0.698892277290878
145 0.79656760941464
146 0.443727417003922
147 0.546184794440524
148 0.910644273191594
149 0.203558274614453
150 0.541317009562348
151 0.0718674761851321
152 0.118387786421703
153 0.308069108827501
154 0.1545875916773
155 0.481506974768113
156 0.501726285379495
157 0.803434596534833
158 0.565632956256833
159 0.798508710834355
160 0.353781025868865
161 0.424206986360772
162 0.967119002456686
163 0.0371684782881307
164 0.129440765349086
165 0.268944067346458
166 0.459122204226279
167 0.901811430952485
168 0.346833918617455
169 0.894859333301691
170 0.260182717008607
171 0.637060789571378
172 0.29360362369519
173 0.189975140954397
174 0.589117242754528
175 0.646208223287314
176 0.401886781057946
177 0.187252188514799
178 0.486737286534717
179 0.166047909547477
180 0.339803717397909
181 0.572614671428052
182 0.947859537650131
183 0.00646742671320873
184 0.0206193924968037
185 0.0222716980244301
186 0.0760308751215135
187 0.138977820266733
188 0.317299670727276
189 0.393175074528794
190 0.954026714903224
191 0.173711265247482
192 0.572884553033731
193 0.96216800058131
194 0.0346074689004151
195 0.117905400068039
196 0.213707861309342
197 0.47996520797413
198 0.533998673668521
199 0.821307282477005
200 0.440252076397601
};
\end{axis}

\end{tikzpicture}}
    \subcaption{\emph{KSOS} Train Trajectory}
    \label{fig:Logistic_errTrajTrain_sos}
\end{minipage}\hspace{0.1\textwidth}
\begin{minipage}[c]{0.40\textwidth}
    \centering
    \resizebox{\textwidth}{!}{
\begin{tikzpicture}

\definecolor{darkgray176}{RGB}{176,176,176}
\definecolor{slategray}{RGB}{112,128,144}

\begin{axis}[
tick align=outside,
tick pos=left,
x grid style={darkgray176},
xlabel={Iterations},
xmin=-10, xmax=210,
xtick style={color=black},
y grid style={darkgray176},
ylabel={Distance to True Trajectory},
ymin=-0.0496422639526937, ymax=1.04248754300657,
ytick style={color=black}
]
\addplot [semithick, slategray]
table {%
0 0
1 1.14908083048704e-14
2 1.82076576038526e-14
3 6.32827124036339e-14
4 1.06692432666478e-13
5 2.60347299274599e-13
6 1.77191594730175e-13
7 6.53616050172445e-13
8 2.0923818233598e-12
9 1.54809498553732e-12
10 6.15035800066721e-12
11 1.65186753164903e-11
12 3.73034936274053e-14
13 1.50164602974456e-13
14 6.17948183942452e-13
15 2.50701195914083e-12
16 9.88682746783098e-12
17 3.83351683730382e-11
18 1.224528811683e-10
19 2.86823342854348e-10
20 4.81546580388681e-10
21 1.236201363497e-09
22 8.80617800902428e-10
23 3.29490329464832e-09
24 9.14701925314176e-09
25 4.69402816616338e-09
26 1.83040837209436e-08
27 6.56675225307879e-08
28 1.57600318018147e-07
29 1.79107872311235e-07
30 5.89797660710545e-07
31 8.14183468866858e-07
32 2.51182761762392e-06
33 9.01001439235216e-07
34 3.53057544617147e-06
35 1.28644780023335e-05
36 3.55422358724322e-05
37 2.49514005090656e-06
38 9.94021031228316e-06
39 3.92603630251752e-05
40 0.00014942013121505
41 0.000481157559319823
42 0.000574682620736788
43 0.00190332004080651
44 0.00288340927430697
45 0.00834973632440289
46 0.00122646903504231
47 0.00488527055109613
48 0.0191932900664243
49 0.071492595714727
50 0.210130326398163
51 0.0756934301092769
52 0.271166291897857
53 0.667262192821687
54 0.256437974700652
55 0.0936728486142694
56 0.319190957365799
57 0.598657042466671
58 0.853574440354422
59 0.101937132872159
60 0.18726811738778
61 0.41773318655658
62 0.515112719371712
63 0.949720256879295
64 0.168660718562798
65 0.545020768281913
66 0.789648903750432
67 0.453370859400644
68 0.522469432716976
69 0.883192545004402
70 0.342822473985021
71 0.792659480539845
72 0.304399412771789
73 0.3348420647024
74 0.888566716432556
75 0.371483911565304
76 0.893045634487153
77 0.00526882802820844
78 0.0140561328511046
79 0.0157496930276395
80 0.0523014588093708
81 0.0794321673694617
82 0.223022786314443
83 0.0101011590380401
84 0.0362932570929957
85 0.0891418333404705
86 0.0870178819965841
87 0.301398336580533
88 0.620506461324088
89 0.716220994794841
90 0.182004005960726
91 0.0248200661040907
92 0.092565448958052
93 0.272514679024015
94 0.10973454588196
95 0.364859593313976
96 0.592342375350908
97 0.957916393341728
98 0.109595885531057
99 0.367677671829491
100 0.624321385713705
101 0.921961268241978
102 0.1921817311249
103 0.542392451963118
104 0.154190061711659
105 0.242981925892713
106 0.617203393503305
107 0.193904544265623
108 0.175141276698829
109 0.576661942341807
110 0.957729356496421
111 0.0395708724684113
112 0.13191063191243
113 0.209679119755013
114 0.545407112100695
115 0.142503647530582
116 0.225707785596683
117 0.58217099461103
118 0.153005689659674
119 0.191761667532118
120 0.579930335769648
121 0.503290335532965
122 0.469287511309916
123 0.722084296435499
124 0.761400808986582
125 0.553669305477015
126 0.499495328951061
127 0.569777820724914
128 0.77082014357662
129 0.374403391711236
130 0.325965970112578
131 0.814802332022863
132 0.0212163540783963
133 0.0279508886298371
134 0.0878675548534313
135 0.0801174881197152
136 0.281139255412476
137 0.620818053555521
138 0.576761104677736
139 0.27963761851542
140 0.341237061371286
141 0.896692737513072
142 0.344444827875109
143 0.863260968286729
144 0.0774555442635592
145 0.152248963142414
146 0.30780702174222
147 0.546644549092892
148 0.910461854380094
149 0.202846440867074
150 0.539507012390494
151 0.0759158452316543
152 0.13141953382113
153 0.333734633834329
154 0.204586266274509
155 0.596806502604449
156 0.346810528230303
157 0.340366830860406
158 0.870104868847978
159 0.168884468038072
160 0.350413670501037
161 0.566631845697871
162 0.968669827224788
163 0.0312385722469668
164 0.109951148409276
165 0.236952857190755
166 0.374957788113663
167 0.861449649234289
168 0.191374423472878
169 0.374786861797406
170 0.669900319263702
171 0.851903897863235
172 0.339504955904314
173 0.64011967896699
174 0.832466457959653
175 0.102843817577689
176 0.159413872295608
177 0.432204754321122
178 0.0521712221745283
179 0.130395826524427
180 0.111008015467041
181 0.387249142324348
182 0.845711800621145
183 0.351316738992219
184 0.635227588137846
185 0.875325798881077
186 0.155388638474594
187 0.333365581973904
188 0.501984647570974
189 0.992845279053875
190 0.0279404031154148
191 0.108586616386439
192 0.386481425918998
193 0.937066304179186
194 0.127781448017612
195 0.386011575454639
196 0.284403769999273
197 0.718315995185183
198 0.11432327987713
199 0.0150554909816142
200 0.0584316896388439
};
\end{axis}

\end{tikzpicture}}
    \subcaption{Gradient Descent Train Trajectory}
    \label{fig:Logistic_errTrajTrain_flow}
\end{minipage}
\caption{Distances of Predicted Trajectory to True Trajectory for Logistic Map.}
\label{fig:Logistic_ErrTraj}
\end{figure}

\begin{figure}
\begin{minipage}[c]{0.32\textwidth}
    \centering
    \resizebox{\textwidth}{!}{
\begin{tikzpicture}

\definecolor{darkgray176}{RGB}{176,176,176}
\definecolor{steelblue31119180}{RGB}{31,119,180}

\begin{axis}[
tick align=outside,
tick pos=left,
x grid style={darkgray176},
xlabel={Iterations},
xmin=-10, xmax=210,
xtick style={color=black},
y grid style={darkgray176},
ylabel={\(\displaystyle x_1\)},
ymin=-0.0497227924067796, ymax=1.04991846394987,
ytick style={color=black}
]
\addplot [semithick, steelblue31119180]
table {%
0 0.3
1 0.84
2 0.5376
3 0.99434496
4 0.0224922420903938
5 0.0879453645445638
6 0.32084390959875
7 0.871612381088557
8 0.447616952886773
9 0.989024065500534
10 0.043421853445319
11 0.166145584354769
12 0.554164916616725
13 0.988264647231613
14 0.0463905370551545
15 0.176953820507555
16 0.582564663661341
17 0.972732305257959
18 0.106096670261984
19 0.379360667285216
20 0.941784605608526
21 0.219305448989276
22 0.684842276131552
23 0.863333331818028
24 0.471955559960043
25 0.996854037531381
26 0.0125442615550601
27 0.0495476122283933
28 0.188370585403432
29 0.611548431832802
30 0.950227789422571
31 0.18917975052666
32 0.613563090069323
33 0.948413698295627
34 0.195700620723353
35 0.629607551087389
36 0.932807530804519
37 0.250710565115582
38 0.751419110620029
39 0.747153723260134
40 0.755660148314612
41 0.738551554255002
42 0.772372623850092
43 0.703252615108066
44 0.83475349780693
45 0.551760382824104
46 0.989283451079609
47 0.0424068179985118
48 0.162433919143012
49 0.544196564219413
50 0.992186654844797
51 0.0310091871707535
52 0.120190469927051
53 0.422978883463062
54 0.976270990429614
55 0.0926637747007795
56 0.336308798235931
57 0.892820761860139
58 0.382767396206479
59 0.945026066431165
60 0.207807200787216
61 0.658493472352791
62 0.899519276886221
63 0.361537389585245
64 0.923312422068527
65 0.283226373289908
66 0.812036779051814
67 0.610532194075877
68 0.951130536291091
69 0.185924956902852
70 0.605427469214098
71 0.955540194940442
72 0.169932523174497
73 0.564221842968184
74 0.98350221954308
75 0.0649024147876622
76 0.24276036536957
77 0.735311081500811
78 0.778514779691674
79 0.689718069973193
80 0.856028215702587
81 0.492975638494528
82 0.999802633381762
83 0.000789310658624437
84 0.00315475058923448
85 0.0125791925518168
86 0.0496838258662445
87 0.188861373254149
88 0.612771019786823
89 0.94913078838496
90 0.193126139698818
91 0.623313735455403
92 0.939174890592139
93 0.22850166189353
94 0.705154609621699
95 0.831646344603873
96 0.560042808443557
97 0.985579444616841
98 0.0568504118623997
99 0.214473770133901
100 0.673899088233806
101 0.879036428445804
102 0.425325543644197
103 0.977694902271861
104 0.087230321373908
105 0.318484769627651
106 0.868208884571492
107 0.45768886929047
108 0.992839072872324
109 0.0284385930013912
110 0.11051935771797
111 0.393219317150309
112 0.954391543080615
113 0.174113302307272
114 0.575191441067713
115 0.977384988760643
116 0.0884142900240045
117 0.322388813374223
118 0.873817065541533
119 0.441043206039667
120 0.986096385783676
121 0.0548412149121923
122 0.207334624236588
123 0.657387911317044
124 0.900916181485033
125 0.35706486169384
126 0.918278184949596
127 0.300173439981087
128 0.84027738364403
129 0.536845208721493
130 0.994569722377078
131 0.0216031588314388
132 0.0845458494397696
133 0.30959139512911
134 0.854978252764485
135 0.495961760257094
136 0.999934770479115
137 0.000260901063977317
138 0.00104333197844853
139 0.00416897374732512
140 0.0166063736208769
141 0.065322407904163
142 0.244221563719061
143 0.73830956613471
144 0.772834202754745
145 0.70224599122873
146 0.836386236127634
147 0.547377200575535
148 0.991021603462502
149 0.0355911397324533
150 0.137297642019993
151 0.473787998062972
152 0.997251723817813
153 0.0109628926408536
154 0.043370830503195
155 0.165959206258632
156 0.553666992466548
157 0.988479415678382
158 0.0455514418340271
159 0.173906031923473
160 0.574650895936421
161 0.977708974943559
162 0.0871765410334977
163 0.31830716690773
164 0.867950857611618
165 0.458448665531498
166 0.993093946415547
167 0.0274334400333682
168 0.106723385605215
169 0.381334018280703
170 0.943673539130382
171 0.212615162702086
172 0.669639821165005
173 0.88488932430042
174 0.407440832158263
175 0.965731201793781
176 0.132377790702881
177 0.459415644926022
178 0.993411640492917
179 0.0261798121043534
180 0.101977718170137
181 0.366313052667795
182 0.928511200451985
183 0.265512604348794
184 0.780062645122859
185 0.686259659227149
186 0.861229357378345
187 0.478053405472111
188 0.998073387954514
189 0.00769160084604985
190 0.0305297604898996
191 0.118390776857316
192 0.417497603249748
193 0.972773418121856
194 0.105941180469506
195 0.378870587000934
196 0.941310661226007
197 0.220979601153059
198 0.688590468109175
199 0.857734541353449
200 0.488103991690549
};
\end{axis}

\end{tikzpicture}}
    \subcaption{True Test Trajectory}
    \label{fig:Logistic_Traj_true}
\end{minipage}\hfill
\begin{minipage}[c]{0.32\textwidth}
    \centering
    \resizebox{\textwidth}{!}{
\begin{tikzpicture}

\definecolor{darkgray176}{RGB}{176,176,176}
\definecolor{darkorange}{RGB}{255,140,0}

\begin{axis}[
tick align=outside,
tick pos=left,
x grid style={darkgray176},
xlabel={Iterations},
xmin=-10, xmax=210,
xtick style={color=black},
y grid style={darkgray176},
ylabel={\(\displaystyle x_1\)},
ymin=-0.0494872289459596, ymax=1.04984877075206,
ytick style={color=black}
]
\addplot [semithick, darkorange]
table {%
0 0.3
1 0.839979608449754
2 0.537653514845992
3 0.994210936046306
4 0.0229248315584861
5 0.0895800170030122
6 0.326224479089051
7 0.879200514527995
8 0.424827880253361
9 0.97739643257378
10 0.0883706821043287
11 0.322245705450554
12 0.873613139786521
13 0.441651912724712
14 0.98638199370698
15 0.0537301702430015
16 0.203373025751047
17 0.648049487284434
18 0.912324714119836
19 0.319953294649439
20 0.870332086978432
21 0.451415986994638
22 0.990554827102311
23 0.0374021144769951
24 0.144099791747241
25 0.493308055152064
26 0.999806278359831
27 0.000771650487442162
28 0.00308178296359698
29 0.0122654343371804
30 0.0484486983820496
31 0.184414312766596
32 0.60160951041296
33 0.958580236852931
34 0.158862474719206
35 0.534500491711654
36 0.995144043166591
37 0.019216052090511
38 0.0753428901187887
39 0.278626309701354
40 0.803952347102743
41 0.630447011186898
42 0.931912309122575
43 0.253775387461917
44 0.757491772408377
45 0.73478348207485
46 0.779506098414307
47 0.687503050226244
48 0.859355006594055
49 0.483452449813673
50 0.998885639503923
51 0.0044507600742221
52 0.0177024881878405
53 0.069519889441645
54 0.258740136154196
55 0.767162074995733
56 0.714483130119396
57 0.815983459477948
58 0.600612545862964
59 0.959394756735965
60 0.155866490102028
61 0.526288403246867
62 0.99723361858223
63 0.0110341159248315
64 0.0436423705605566
65 0.166954950056912
66 0.556324632365323
67 0.987259248094673
68 0.0503060081419521
69 0.191114235217937
70 0.618330050420026
71 0.943877279548863
72 0.21189008929607
73 0.667970078117536
74 0.887143193904989
75 0.400483805075427
76 0.960381874121412
77 0.152221273119057
78 0.516199634842554
79 0.998949767280446
80 0.00419569600896414
81 0.0167020955940642
82 0.0656728185538722
83 0.245425570598998
84 0.740763566224088
85 0.768111229088724
86 0.712450852440005
87 0.819457712368972
88 0.591786286272164
89 0.966295142541456
90 0.130281722283661
91 0.453222694834376
92 0.991241562031082
93 0.0346909223200099
94 0.134089904801472
95 0.46439936407433
96 0.994903131810296
97 0.0201701007923864
98 0.0790117302838935
99 0.291022473578652
100 0.825312292486175
101 0.576688362143409
102 0.976474350730052
103 0.0918897663710822
104 0.333789793854246
105 0.889487331578899
106 0.393198619505614
107 0.954373835750674
108 0.1741777931665
109 0.575359611742116
110 0.977282601068702
111 0.0888055308051407
112 0.323677743152889
113 0.875641579234603
114 0.435573160682182
115 0.983393316221416
116 0.0653123139706425
117 0.24417264763867
118 0.738207768256616
119 0.773019051753116
120 0.701833397109685
121 0.837030397465285
122 0.545641839522164
123 0.991536455352647
124 0.033530760642219
125 0.129767066641118
126 0.451704533924967
127 0.990666073782987
128 0.0369620538897103
129 0.142483803925823
130 0.488687159607825
131 0.999462540291859
132 0.00214331596571698
133 0.0085072981470051
134 0.0336525939370805
135 0.130222135390375
136 0.453047016807159
137 0.991175812263651
138 0.0349499348865016
139 0.135051771400073
140 0.467206685450025
141 0.995671523224905
142 0.01713602060829
143 0.0673411359732252
144 0.251216762745232
145 0.752421795559319
146 0.745106524711336
147 0.7596601993697
148 0.730304741999095
149 0.787835938781739
150 0.668601019986655
151 0.886293835700127
152 0.403111186130768
153 0.962447706587784
154 0.144585572391858
155 0.494693612456442
156 0.999878952583972
157 0.00048258922213239
158 0.0019259390586645
159 0.00764360636644323
160 0.0302791106846969
161 0.11752695117807
162 0.414862293560129
163 0.970992631956674
164 0.112666057659816
165 0.399890598937534
166 0.959907006036318
167 0.153976851026673
168 0.521071740130005
169 0.998220870001597
170 0.00710078663146562
171 0.0281714735877638
172 0.109512521194576
173 0.390082166437698
174 0.951663402194213
175 0.184024970460101
176 0.600627148038956
177 0.959382877762309
178 0.155910260007852
179 0.526408913498726
180 0.997208528367615
181 0.0111347132177101
182 0.0440428951176568
183 0.168412575080343
184 0.560199152552729
185 0.985503806719879
186 0.0571424944615395
187 0.215507671136668
188 0.676251913193827
189 0.875736899961121
190 0.435286751528849
191 0.98324589233712
192 0.0658839563661712
193 0.246159121036119
194 0.742254361144007
195 0.765226096752034
196 0.718607112293331
197 0.808839689860621
198 0.618465648368939
199 0.943750045484276
200 0.212340787016299
};
\end{axis}

\end{tikzpicture}}
    \subcaption{\emph{KSOS} Test Trajectory}
    \label{fig:Logistic_Traj_sos}
\end{minipage}\hfill
\begin{minipage}[c]{0.32\textwidth}
    \centering
    \resizebox{\textwidth}{!}{
\begin{tikzpicture}

\definecolor{darkgray176}{RGB}{176,176,176}
\definecolor{slategray}{RGB}{112,128,144}

\begin{axis}[
tick align=outside,
tick pos=left,
x grid style={darkgray176},
xlabel={Iterations},
xmin=-10, xmax=210,
xtick style={color=black},
y grid style={darkgray176},
ylabel={\(\displaystyle x_1\)},
ymin=-0.0497980677817583, ymax=1.05015020607843,
ytick style={color=black}
]
\addplot [semithick, slategray]
table {%
0 0.3
1 0.839432247425812
2 0.539082371111309
3 0.992904020022989
4 0.0281888020143943
5 0.109575181828004
6 0.39034768770377
7 0.951710935209581
8 0.18385287584059
9 0.598875162792588
10 0.95980194087434
11 0.154976198333567
12 0.523826311613964
13 0.997742145338256
14 0.0090253979775098
15 0.0357439750615205
16 0.138026241850929
17 0.476781333680819
18 0.99784469969484
19 0.00861560878291622
20 0.0341208075669643
21 0.131977735083556
22 0.458968209856621
23 0.993274953219192
24 0.0267649382626092
25 0.10419379783922
26 0.373440299201901
27 0.936038700434406
28 0.238582489550656
29 0.726621421761649
30 0.794559936160931
31 0.653017028234582
32 0.906367898116949
33 0.339453829165278
34 0.896859787827006
35 0.370343813579279
36 0.932867872067091
37 0.251091289316982
38 0.752782347342468
39 0.742893857972039
40 0.760801368778476
41 0.728288760422336
42 0.791512448649162
43 0.660078273958068
44 0.897508432008699
45 0.36812903153978
46 0.930815846112281
47 0.258155333120071
48 0.766067713176078
49 0.717394396746451
50 0.810987264219011
51 0.60963105810494
52 0.949403466659837
53 0.191951453072579
54 0.615249192678732
55 0.943321683035882
56 0.213849966002965
57 0.672457177959104
58 0.88104069802481
59 0.419291614829019
60 0.974022298562319
61 0.101243067072097
62 0.363835719610397
63 0.92569918817241
64 0.275106853141215
65 0.79755820885656
66 0.645965760371933
67 0.914481974212088
68 0.312819657114764
69 0.859801426325155
70 0.482188210627592
71 0.998432224202022
72 0.00626067753267969
73 0.0248923935968387
74 0.097086486686944
75 0.350642136929796
76 0.910768268345696
77 0.325091840708551
78 0.877525964025377
79 0.430136018901801
80 0.980023065686318
81 0.0783298637934224
82 0.289145344820395
83 0.822170928975192
84 0.584823810459054
85 0.971217480091658
86 0.111824409800457
87 0.397380102169548
88 0.957924759818283
89 0.162038088370957
90 0.543119556557325
91 0.991305374247753
92 0.0345067120412703
93 0.133429511768332
94 0.463235938318389
95 0.994634918556528
96 0.0215112597703976
97 0.0840708343627993
98 0.308071129210675
99 0.852623826612168
100 0.502647868771015
101 1.00002800585526
102 0.000212159246601211
103 0.000912356150708205
104 0.00364593999496143
105 0.0145312006551961
106 0.0572800813848845
107 0.216034750834763
108 0.677455294446331
109 0.874438222749574
110 0.438309270332803
111 0.984834299159492
112 0.0597448385850106
113 0.225053636484079
114 0.697608365969679
115 0.844103666434339
116 0.526327752645875
117 0.99723270174552
118 0.0110401200546166
119 0.043675906149292
120 0.167184635058458
121 0.556702870867265
122 0.985004780597262
123 0.0590837433761918
124 0.222668965832038
125 0.692304072591836
126 0.852592806154779
127 0.502735415581144
128 1.00001702944574
129 0.000213750380632782
130 0.000911566164002319
131 0.00364280178303084
132 0.0145186756101882
133 0.0572314477823868
134 0.215857521164439
135 0.677054312515774
136 0.874956179082635
137 0.436932171889269
138 0.984269020805749
139 0.0619335860209029
140 0.232377453010586
141 0.713676881807693
142 0.817359911272343
143 0.596735433170264
144 0.96176706630806
145 0.147028464500166
146 0.501756622708496
147 1.00015255726661
148 0.000199581030068589
149 0.000918657383404888
150 0.00367098506058097
151 0.0146311064967596
152 0.0576679649988416
153 0.217453755459688
154 0.680645313758614
155 0.869541798874918
156 0.453904938009062
157 0.991188863278093
158 0.0349632891137727
159 0.135138136350986
160 0.468177232666612
161 0.995898341633636
162 0.0165310417405424
163 0.065036215145093
164 0.242066828520648
165 0.733870930565015
166 0.781213380577815
167 0.683662928852772
168 0.864564541407284
169 0.4684478373669
170 0.995973046426516
171 0.0162302118581179
172 0.0638700361978504
173 0.238301810885665
174 0.725986564056527
175 0.795717803427377
176 0.650307107850915
177 0.909634844732403
178 0.32884154772331
179 0.882935999190423
180 0.413619919995483
181 0.970209278073892
182 0.115588490421707
183 0.408972505974039
184 0.966854678902068
185 0.128189385821826
186 0.446663663491357
187 0.988190093780734
188 0.0466794500004774
189 0.177993319688974
190 0.585254845595964
191 0.970927543345247
192 0.112910861464253
193 0.400706578824033
194 0.960617372223575
195 0.151681925109253
196 0.51469812434866
197 0.999136052411835
198 0.00345277990156196
199 0.0137634742439109
200 0.0542962185123772
};
\end{axis}

\end{tikzpicture}}
    \subcaption{Gradient Descent Test Trajectory}
    \label{fig:Logistic_Traj_flow}
\end{minipage}\hfill

\vspace{0.5cm}

\begin{minipage}[c]{0.32\textwidth}
    \centering
    \resizebox{\textwidth}{!}{
\begin{tikzpicture}

\definecolor{darkgray176}{RGB}{176,176,176}
\definecolor{steelblue31119180}{RGB}{31,119,180}

\begin{axis}[
tick align=outside,
tick pos=left,
x grid style={darkgray176},
xlabel={Iterations},
xmin=-10, xmax=210,
xtick style={color=black},
y grid style={darkgray176},
ylabel={\(\displaystyle x_1\)},
ymin=-0.0497594622394164, ymax=1.04992925028317,
ytick style={color=black}
]
\addplot [semithick, steelblue31119180]
table {%
0 0.1
1 0.36
2 0.9216
3 0.28901376
4 0.82193922612265
5 0.585420538734197
6 0.970813326249438
7 0.113339247303761
8 0.401973849297512
9 0.961563495113813
10 0.147836559913285
11 0.503923645865164
12 0.999938420012499
13 0.000246304781624128
14 0.000984976462314709
15 0.00393602513473358
16 0.0156821313634893
17 0.0617448084775503
18 0.231729548414484
19 0.712123859224412
20 0.820013873390967
21 0.590364483349242
22 0.967337040596098
23 0.126384361947522
24 0.44164542001056
25 0.986378971977024
26 0.0537419824742921
27 0.2034151271761
28 0.648149652848124
29 0.912206721443921
30 0.320342475185814
31 0.870892695110561
32 0.449754434854499
33 0.989901532732837
34 0.0399859529040691
35 0.153548305897691
36 0.519884894614559
37 0.998418363864672
38 0.00631653824985557
39 0.0251065583775747
40 0.0979048764160326
41 0.353278046359976
42 0.913890673280218
43 0.31477804228659
44 0.862771305523247
45 0.473587919555836
46 0.997209608026444
47 0.0111304227447597
48 0.0440261457371306
49 0.168351376914654
50 0.560036763222377
51 0.985582348247121
52 0.0568391322832473
53 0.214433781298139
54 0.673807738945284
55 0.879163479530912
56 0.424940223160047
57 0.977464119602946
58 0.0881120579671347
59 0.321393292831724
60 0.872398576618023
61 0.445277200531482
62 0.988021660873313
63 0.047339434073811
64 0.180393648221529
65 0.591407119611426
66 0.96657895393737
67 0.129216318970838
68 0.450077847529859
69 0.990031114770992
70 0.0394780262251962
71 0.151678046682236
72 0.514687267347589
73 0.999137136711442
74 0.00344847502201393
75 0.0137463321681459
76 0.0542294820802756
77 0.205154581414323
78 0.652264716556147
79 0.907261824368305
80 0.3365512256488
81 0.893137992652362
82 0.381770074933085
83 0.944086739274687
84 0.211147872001506
85 0.666257792602967
86 0.889433385595155
87 0.393366552735581
88 0.954517231698025
89 0.173656344358254
90 0.573999273689525
91 0.978096429973691
92 0.0856952145856461
93 0.313406179131065
94 0.860730984054127
95 0.479492628573366
96 0.99831779086868
97 0.00671751721503357
98 0.0266895687099972
99 0.103908942528286
100 0.372447496763758
101 0.934921435672674
102 0.243373379169681
103 0.736571109924846
104 0.776136439795706
105 0.69499466646781
106 0.847908320196431
107 0.515839202952392
108 0.998996478599332
109 0.00401005738186629
110 0.0159759072866417
111 0.0628827106920414
112 0.23571390155225
113 0.720611432669064
114 0.805322383102812
115 0.62711296950568
116 0.935369171933792
117 0.241814736518736
118 0.73336147888444
119 0.782169680691468
120 0.6815210851939
121 0.868200382520116
122 0.457713913248162
123 0.992847547468864
124 0.0284051798157027
125 0.110393302301361
126 0.392826484433446
127 0.954055350244423
128 0.175334955657659
129 0.578370435928742
130 0.975432299089356
131 0.0958565159304373
132 0.346672177136461
133 0.905962314943708
134 0.340778395382181
135 0.898593922491708
136 0.364491539810697
137 0.926549828868496
138 0.272220973969027
139 0.792466861201525
140 0.657852540395713
141 0.90033030196248
142 0.358942597322519
143 0.920411236599532
144 0.293017568563409
145 0.828633092306388
146 0.568001162564565
147 0.981503367559471
148 0.0726180281155576
149 0.269378600432663
150 0.787255080246411
151 0.669938075490513
152 0.884484201994323
153 0.408687593667154
154 0.966648177798821
155 0.128957912628161
156 0.449311077594995
157 0.989722532581678
158 0.0406873643271507
159 0.156127610845042
160 0.527007119907444
161 0.99708246189722
162 0.011636104296796
163 0.0460028214943603
164 0.175546247635673
165 0.578919050306833
166 0.975087133994671
167 0.09716886045092
168 0.350908292038358
169 0.911086650468322
170 0.324031063226941
171 0.876139733163837
172 0.43407560454175
173 0.982615896333857
174 0.0683275864234714
175 0.254635709428058
176 0.75918545964851
177 0.731291590027162
178 0.78601680153083
179 0.672777556968296
180 0.880591663232269
181 0.420599943512381
182 0.974782524119052
183 0.0983262191645686
184 0.354632695157479
185 0.915473386731287
186 0.309527459672139
187 0.854880845524205
188 0.496238341920101
189 0.99994339971396
190 0.000226388329791984
191 0.000905348312464471
192 0.00361811462759036
193 0.0144200954965279
194 0.0568486253695956
195 0.214467436652732
196 0.673884621073354
197 0.879056554216705
198 0.425264514821433
199 0.977658429020496
200 0.0873697007426853
};
\end{axis}

\end{tikzpicture}}
    \subcaption{True Train Trajectory}
    \label{fig:Logistic_TrajTrain_true}
\end{minipage}\hfill
\begin{minipage}[c]{0.32\textwidth}
    \centering
    \resizebox{\textwidth}{!}{
\begin{tikzpicture}

\definecolor{darkgray176}{RGB}{176,176,176}
\definecolor{darkorange}{RGB}{255,140,0}

\begin{axis}[
tick align=outside,
tick pos=left,
x grid style={darkgray176},
xlabel={Iterations},
xmin=-10, xmax=210,
xtick style={color=black},
y grid style={darkgray176},
ylabel={\(\displaystyle x_1\)},
ymin=-0.0499470169344352, ymax=1.04999236854985,
ytick style={color=black}
]
\addplot [semithick, darkorange]
table {%
0 0.1
1 0.359999999999999
2 0.921599999999999
3 0.289013760000003
4 0.821939226122655
5 0.585420538734186
6 0.970813326249445
7 0.113339247303735
8 0.401973849297433
9 0.96156349511375
10 0.147836559913521
11 0.503923645865828
12 0.999938420012476
13 0.000246304781716593
14 0.000984976462683519
15 0.00393602513618199
16 0.0156821313692376
17 0.0617448084998126
18 0.231729548492497
19 0.712123859391787
20 0.820013873106493
21 0.59036448407712
22 0.96733704006964
23 0.12638436391749
24 0.441645425896706
25 0.986378974716594
26 0.053741971765003
27 0.20341508900635
28 0.648149562042448
29 0.912206828444659
30 0.32034212231007
31 0.870892187303007
32 0.449755940919096
33 0.989902137909285
34 0.0399835798418092
35 0.153539577899823
36 0.519860699558625
37 0.998422171438756
38 0.00630128437851046
39 0.0250454494776918
40 0.0976724064709155
41 0.352530125070691
42 0.913010113057394
43 0.317690466407694
44 0.867050460330309
45 0.461092965375707
46 0.99392626958679
47 0.0240633176885048
48 0.0939303505199789
49 0.340434712952738
50 0.898155203046188
51 0.365891743321848
52 0.928056616556045
53 0.26705004497693
54 0.782928470840723
55 0.679805242181876
56 0.870676888378927
57 0.450394313915488
58 0.990156131078334
59 0.0389818590669015
60 0.149873420613775
61 0.50964418000305
62 0.999618047795286
63 0.00152186830315266
64 0.00604520445145168
65 0.0240209072458087
66 0.0937685238650679
67 0.339909113198951
68 0.897482596300297
69 0.368034737160574
70 0.930335072086771
71 0.259212906437105
72 0.768072791474104
73 0.712533285997206
74 0.819317382122834
75 0.59214464529275
76 0.966025942933692
77 0.131290972652496
78 0.456194479697628
79 0.992320347010751
80 0.0304595651730386
81 0.118210918896875
82 0.416952931295355
83 0.972401155039237
84 0.107360618212394
85 0.383346148977557
86 0.945561755414183
87 0.205907945676634
88 0.654038032127674
89 0.905087314203622
90 0.343620108518481
91 0.902178228390337
92 0.353016167874535
93 0.9135828163061
94 0.315796934137149
95 0.864275443390321
96 0.469210092094221
97 0.996185613821409
98 0.0151209734749237
99 0.0595660537607713
100 0.22406531667515
101 0.695422279461005
102 0.847238197611639
103 0.517701522156773
104 0.998745016987961
105 0.0050109206567503
106 0.0199098571647091
107 0.0780112171105692
108 0.287648754461265
109 0.819620775052908
110 0.591369678753483
111 0.966606314068183
112 0.129114196097612
113 0.449774874425353
114 0.989909744501853
115 0.0399537530560143
116 0.153429868610813
117 0.519556523312874
118 0.998469681127348
119 0.00611102075701832
120 0.0242839854645562
121 0.0947720517553212
122 0.343164614772454
123 0.901607666193288
124 0.354849536725676
125 0.915725057463579
126 0.308689847796279
127 0.853597687012783
128 0.499871745402611
129 0.999995123755107
130 5.02278603048715e-05
131 0.000306210834369763
132 0.00122321122506142
133 0.00486983527713783
134 0.0193525825884519
135 0.0758680674105078
136 0.280404826393231
137 0.807088143044362
138 0.622783463952042
139 0.939631131781538
140 0.226874091267089
141 0.701594749765078
142 0.83741578059293
143 0.544601939924428
144 0.991909845854287
145 0.0320654828917478
146 0.124273745560643
147 0.435318573118947
148 0.983262301307152
149 0.0658203258182098
150 0.245938070684063
151 0.741805551675645
152 0.76609641557262
153 0.716756702494655
154 0.81206058612152
155 0.610464887396274
156 0.95103736297449
157 0.186287936046845
158 0.606320320583984
159 0.954636321679397
160 0.173226094038579
161 0.572875475536447
162 0.978755106753482
163 0.083171299782491
164 0.304987012984759
165 0.847863117653291
166 0.515964929768392
167 0.998980291403405
168 0.00407437342090336
169 0.0162273171666309
170 0.0638483462183344
171 0.239078943592459
172 0.727679228236939
173 0.79264075537946
174 0.657444829178
175 0.900843932715372
176 0.357298678590565
177 0.918543778541961
178 0.299279514996112
179 0.838825466515773
180 0.54078794583436
181 0.993214614940433
182 0.0269229864689207
183 0.104793645877777
184 0.375252087654283
185 0.937745084755717
186 0.233496584550625
187 0.715903025257473
188 0.813538012647377
189 0.606768325185166
190 0.954253103233016
191 0.174616613559947
192 0.576502667661321
193 0.976588096077838
194 0.0914560942700107
195 0.332372836720771
196 0.887592482382696
197 0.399091346242575
198 0.959263188489954
199 0.156351146543492
200 0.527621777140286
};
\end{axis}

\end{tikzpicture}}
    \subcaption{\emph{KSOS} Train Trajectory}
    \label{fig:Logistic_TrajTrain_sos}
\end{minipage}\hfill
\begin{minipage}[c]{0.32\textwidth}
    \centering
    \resizebox{\textwidth}{!}{
\begin{tikzpicture}

\definecolor{darkgray176}{RGB}{176,176,176}
\definecolor{slategray}{RGB}{112,128,144}

\begin{axis}[
tick align=outside,
tick pos=left,
x grid style={darkgray176},
xlabel={Iterations},
xmin=-10, xmax=210,
xtick style={color=black},
y grid style={darkgray176},
ylabel={\(\displaystyle x_1\)},
ymin=-0.0497383009800792, ymax=1.04992302577409,
ytick style={color=black}
]
\addplot [semithick, slategray]
table {%
0 0.1
1 0.360000000000012
2 0.921600000000018
3 0.289013759999937
4 0.821939226122543
5 0.585420538734458
6 0.970813326249261
7 0.113339247304415
8 0.401973849299605
9 0.961563495115361
10 0.147836559907135
11 0.503923645848645
12 0.999938420012536
13 0.000246304781473964
14 0.000984976461696761
15 0.00393602513222657
16 0.0156821313536025
17 0.0617448084392151
18 0.231729548292031
19 0.712123858937589
20 0.820013873872513
21 0.59036448211304
22 0.967337041476716
23 0.126384358652619
24 0.441645410863541
25 0.986378967282996
26 0.0537420007783758
27 0.203415192843622
28 0.648149810448442
29 0.912206542336049
30 0.320343064983475
31 0.87089350929403
32 0.449751923026881
33 0.989900631731398
34 0.0399894834795152
35 0.153561170375693
36 0.519920436850432
37 0.998415868724621
38 0.00632647846016785
39 0.0251458187405999
40 0.0980542965472476
41 0.353759203919296
42 0.914465355900955
43 0.312874722245783
44 0.85988789624894
45 0.481937655880239
46 0.998436077061486
47 0.00624515219366359
48 0.0248328556707063
49 0.0968587811999273
50 0.349906436824215
51 0.909888918137844
52 0.328005424181105
53 0.881695974119826
54 0.417369764244632
55 0.972836328145181
56 0.105749265794248
57 0.378807077136275
58 0.941686498321556
59 0.219456159959565
60 0.685130459230243
61 0.863010387088062
62 0.472908941501601
63 0.997059690953106
64 0.0117329296587314
65 0.0463863513295126
66 0.176930050186938
67 0.582587178371482
68 0.972547280246835
69 0.10683856976659
70 0.382300500210217
71 0.94433752722208
72 0.2102878545758
73 0.664295072009042
74 0.89201519145457
75 0.38523024373345
76 0.947275116567429
77 0.199885753386114
78 0.638208583705042
79 0.923011517395944
80 0.284249766839429
81 0.8137058252829
82 0.604792861247528
83 0.954187898312727
84 0.174854614908511
85 0.577115959262497
86 0.976451267591739
87 0.0919682161550479
88 0.334010770373937
89 0.889877339153094
90 0.391995267728799
91 0.9532763638696
92 0.178260663543698
93 0.58592085815508
94 0.970465529936086
95 0.11463303525939
96 0.405975415517772
97 0.964633910556762
98 0.136285454241054
99 0.471586614357777
100 0.996768882477463
101 0.0129601674306963
102 0.051191648044781
103 0.194178657961728
104 0.621946378084047
105 0.937976592360523
106 0.230704926693126
107 0.709743747218015
108 0.823855201900502
109 0.580671999723674
110 0.973705263783063
111 0.102453583160453
112 0.36762453346468
113 0.930290552424077
114 0.259915271002117
115 0.769616617036261
116 0.70966138633711
117 0.823985731129766
118 0.580355789224766
119 0.973931348223586
120 0.101590749424252
121 0.364910046987151
122 0.927001424558078
123 0.270763251033365
124 0.789805988802285
125 0.664062607778376
126 0.892321813384507
127 0.384277529519509
128 0.946155099234279
129 0.203967044217505
130 0.649466328976777
131 0.910658847953301
132 0.325455823058064
133 0.878011426313871
134 0.428645950235613
135 0.978711410611423
136 0.0833522843982207
137 0.305731775312975
138 0.848982078646762
139 0.512829242686105
140 0.999089601766998
141 0.00363756444940819
142 0.0144977694474097
143 0.0571502683128034
144 0.21556202429985
145 0.676384129163974
146 0.875808184306784
147 0.434858818466579
148 0.983079882495652
149 0.0665321595655886
150 0.247748067855917
151 0.745853920722167
152 0.753064668173193
153 0.742422227501483
154 0.762061911524311
155 0.725764415232611
156 0.796121605825297
157 0.649355701721271
158 0.910792233175129
159 0.325012078883113
160 0.877420790408481
161 0.430450616199349
162 0.980305931521584
163 0.077241393741327
164 0.285497396044949
165 0.815871907497588
166 0.600129345881007
167 0.958618509685209
168 0.159533868565481
169 0.536299788670917
170 0.993931382490643
171 0.0242358353006017
172 0.0945706486374363
173 0.342496217366867
174 0.900794044383125
175 0.357479527005747
176 0.918599331944119
177 0.29908683570604
178 0.838188023705358
179 0.542381730443869
180 0.99159967869931
181 0.0333508011880332
182 0.129070723497907
183 0.449642958156788
184 0.989860283295326
185 0.0401475878502097
186 0.154138821197544
187 0.521515263550301
188 0.998222989491075
189 0.00709812066008479
190 0.0281667914452068
191 0.109491964698903
192 0.390099540546588
193 0.951486399675714
194 0.184630073387208
195 0.600479012107371
196 0.958288391072626
197 0.160740559031522
198 0.539587794698563
199 0.992713920002111
200 0.0289380111038414
};
\end{axis}

\end{tikzpicture}}
    \subcaption{Gradient Descent Train Trajectory}
    \label{fig:Logistic_TrajTrain_flow}
\end{minipage}
\caption{Predicted Trajectories for Logistic Dynamics.}
\label{fig:Logistic_Traj}
\end{figure}
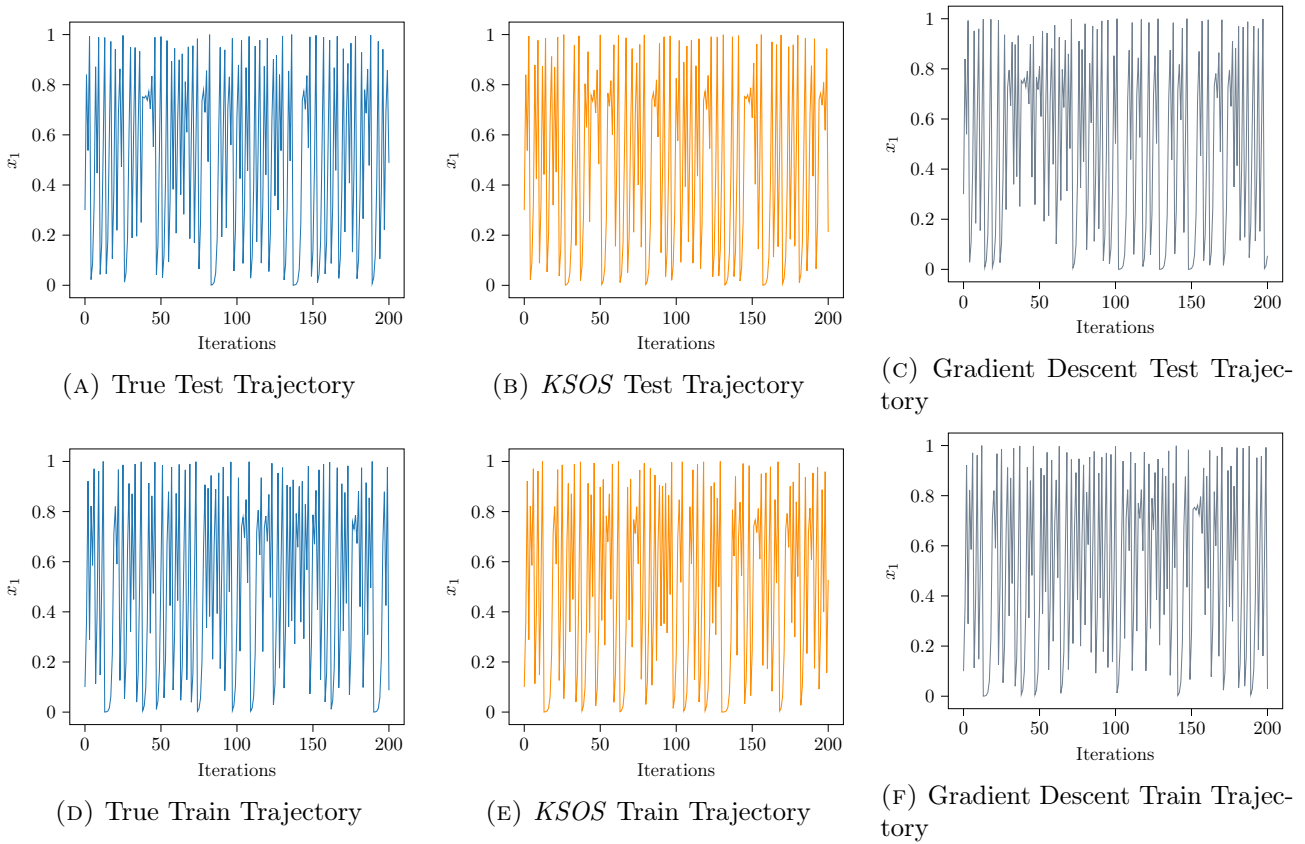

\subsection{The H\'enon Map}

The Henon map is given by
\begin{align*}
    x_{t + 1} &= 1 - 1.4 x_{t}^2 + y_t \\
    y_{t + 1} &= 0.3 x_{t}
\end{align*}
For both the training and test trajectories, we let the number of steps be $1000$. To construct the training set we let $X_0^{(train)} = (-0.75, -0.3)$ and for the test set we let $X_0^{(test)} = (0.5, 0)$. We set the learning rate of gradient descent to $\eta = 10^{-1}$.

\begin{figure}
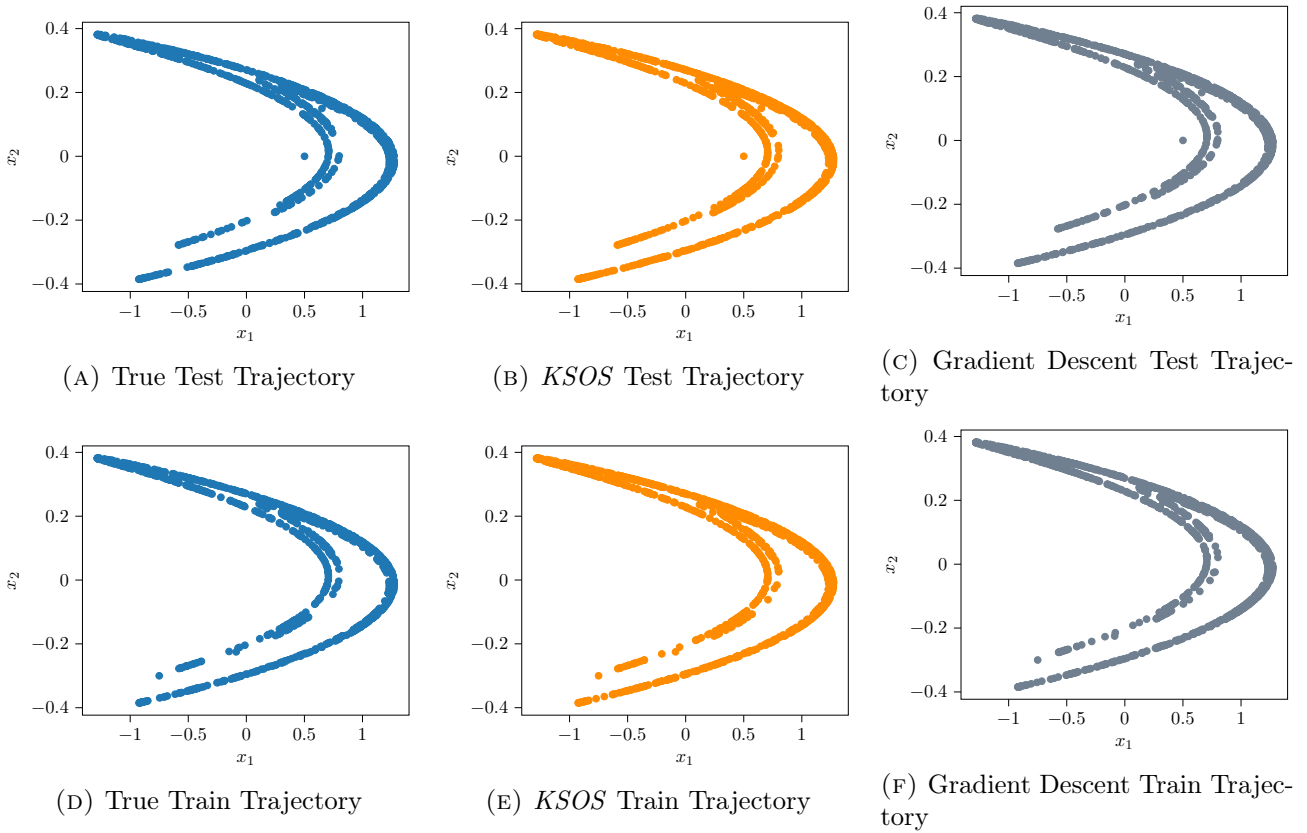

\begin{minipage}[c]{0.32\textwidth}
    \centering
    \resizebox{\textwidth}{!}{\input{figs/Henon_Traj_true}}
    \subcaption{True Test Trajectory}
    \label{fig:Henon_Traj_true.1}
\end{minipage}\hfill
\begin{minipage}[c]{0.32\textwidth}
    \centering
    \resizebox{\textwidth}{!}{\input{figs/Henon_Traj_sos}}
    \subcaption{\emph{KSOS} Test Trajectory}
    \label{fig:Henon_Traj_sos}
\end{minipage}\hfill
\begin{minipage}[c]{0.32\textwidth}
    \centering
    \resizebox{\textwidth}{!}{\input{figs/Henon_Traj_flow}}
    \subcaption{Gradient Descent Test Trajectory}
    \label{fig:Henon_Traj_flow}
\end{minipage}\hfill

\vspace{0.5cm}

\begin{minipage}[c]{0.32\textwidth}
    \centering
    \resizebox{\textwidth}{!}{\input{figs/Henon_TrajTrain_true}}
    \subcaption{True Train Trajectory}
    \label{fig:Henon_TrajTrain_true}
\end{minipage}\hfill
\begin{minipage}[c]{0.32\textwidth}
    \centering
    \resizebox{\textwidth}{!}{\input{figs/Henon_TrajTrain_sos}}
    \subcaption{\emph{KSOS} Train Trajectory}
    \label{fig:Henon_TrajTrain_sos}
\end{minipage}\hfill
\begin{minipage}[c]{0.32\textwidth}
    \centering
    \resizebox{\textwidth}{!}{\input{figs/Henon_TrajTrain_flow}}
    \subcaption{Gradient Descent Train Trajectory}
    \label{fig:Henon_TrajTrain_flow}
\end{minipage}
\caption{Predicted Trajectories for Henon Dynamics.}
\label{fig:Henon_Traj}
\end{figure}

We see in Table \ref{tab:rhoHenon} that while \emph{KSOS} improves on gradient descent in both the median and 25th percentile values, it is more prone to outliers. However, this small improvement does lead to more pronounced improvements in the behavior of the learned kernel. We see in Table \ref{tab:testHenon} that the median mean error is significantly improved for \emph{KSOS}, and the 25th percentile of the flow is larger than the 75th percentile of \emph{KSOS}. Also, the predicted trajectory using \emph{KSOS} results remains close to the true trajectory for longer than the trajectory obtained via the gradient descent parameters. While the Hausdorff distance is also slightly improved, the difference is not large, which we suspect to be due to the predicted trajectories remaining in a similar region as seen in Figure \ref{fig:Henon_Traj}. This is then similar to the Logistic dynamics, where the deviation from the true trajectory can not become too large. For the results on the training trajectory given in Table \ref{tab:trainHenon}, the statistics are very similar. The most significant difference is that the trajectory predicted by gradient descent remains closer to the true trajectory for slightly longer.

\begin{table}[H]
    \centering
    \begin{tabular}{l r r} 
        \toprule
        \textbf{Measure} & \textbf{KSOS} & \textbf{Gradient Descent}  \\
        \midrule
        Relative $\rho$ & $6.08 \ [2.35, 21.53] \times 10^{-3}$ & $7.94 \ [6.70, 10.96] \times 10^{-3}$ \\ 
        \bottomrule
    \end{tabular}
    \caption{Relative $\rho$ Statistics for the Henon Map.}
    \label{tab:rhoHenon}
\end{table}

\begin{table}[H]
    \centering
    \begin{tabular}{l r r} 
        \toprule
        \textbf{Measure} & \textbf{KSOS} & \textbf{Gradient Descent}  \\
        \midrule
        Mean Error & $8.68 \ [6.40, 11.98] \times 10^{-5}$ & $3.01 \ [2.90, 3.21] \times 10^{-4}$ \\ HD & $9.64 \ [9.21, 9.70] \times 10^{-2}$ & $9.69 \ [9.63, 9.70] \times 10^{-2}$ \\ Deviation(0.1) & $9.00 \ [8.25, 9.00]$ & $4.00$ \\ Deviation(0.25) & $1.00 \ [0.93, 1.00] \times 10^{1}$ & $4.00 \ [4.00, 6.00]$ \\  
        \bottomrule
    \end{tabular}
    \caption{Test Set Statistics for the Henon Map.}
    \label{tab:testHenon}
\end{table}

\begin{table}[H]
    \centering
    \begin{tabular}{l r r} 
        \toprule
        \textbf{Measure} & \textbf{KSOS} & \textbf{Gradient Descent}  \\
        \midrule
        Mean Error & $2.43 \ [1.55, 4.14] \times 10^{-13}$ & $3.04 \ [2.73, 3.25] \times 10^{-15}$ \\ HD & $8.51 \ [8.36, 8.75] \times 10^{-2}$ & $8.48 \ [8.40, 8.52] \times 10^{-2}$ \\ Deviation(0.1) & $7.20 \ [6.50, 7.20] \times 10^{1}$ & $8.00 \ [7.60, 8.40] \times 10^{1}$ \\ Deviation(0.25) & $7.20 \ [6.60, 7.40] \times 10^{1}$ & $8.00 \ [7.60, 8.40] \times 10^{1}$ \\  
        \bottomrule
    \end{tabular}
    \caption{Training Set Statistics for the Henon Map.}
    \label{tab:trainHenon}
\end{table}

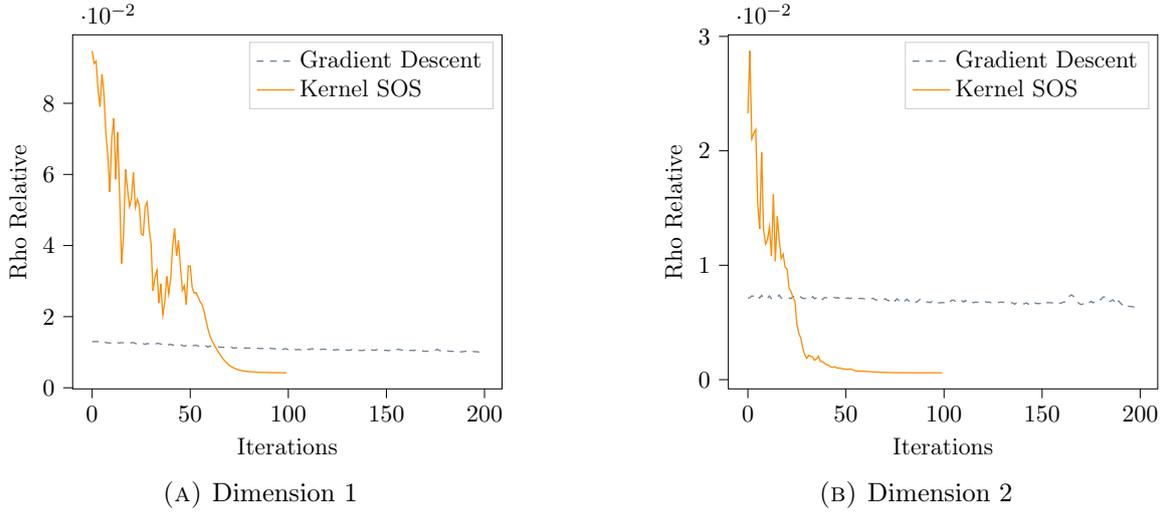
\begin{figure}
\begin{minipage}[c]{0.40\textwidth}
    \centering
    \resizebox{\textwidth}{!}{
\begin{tikzpicture}

\definecolor{darkgray176}{RGB}{176,176,176}
\definecolor{darkorange}{RGB}{255,140,0}
\definecolor{lightgray204}{RGB}{204,204,204}
\definecolor{slategray}{RGB}{112,128,144}

\begin{axis}[
legend cell align={left},
legend style={fill opacity=0.8, draw opacity=1, text opacity=1, draw=lightgray204},
tick align=outside,
tick pos=left,
x grid style={darkgray176},
xlabel={Iterations},
xmin=-9.95, xmax=208.95,
xtick style={color=black},
y grid style={darkgray176},
ylabel={Rho Relative},
ymin=-0.000304915706560854, ymax=0.0992538623680307,
ytick style={color=black}
]
\addplot [semithick, slategray, dashed]
table {%
0 0.0129401849432189
1 0.0130109180870414
2 0.0129903300052129
3 0.0129723542814751
4 0.0132179122727107
5 0.0130895862291245
6 0.013141243655467
7 0.0126992446976065
8 0.0126430054815633
9 0.0125931441936842
10 0.0124636808819929
11 0.0127115521496146
12 0.0125337014400668
13 0.012677885102191
14 0.012603767275053
15 0.0127116368260866
16 0.0126375973166688
17 0.0125540209106467
18 0.0125376700816444
19 0.0126665261487223
20 0.0126250278315764
21 0.0127474650245158
22 0.0124583249644328
23 0.0123161250812048
24 0.0125911702122099
25 0.0123779123239264
26 0.0124117995446601
27 0.0122077486846274
28 0.0124622425125107
29 0.0123434886541501
30 0.0122983256725575
31 0.0121101577372277
32 0.0122713794416364
33 0.0124638855381292
34 0.0124661320731349
35 0.0122669473329308
36 0.0121680240473915
37 0.0123967056678065
38 0.012047432207458
39 0.0120659151859391
40 0.0122523858117504
41 0.0119461006203028
42 0.0121863508229079
43 0.0121398459080679
44 0.0120802691812667
45 0.0120976191110885
46 0.0119089196991066
47 0.0117063193552067
48 0.0119241731304467
49 0.0120568281247979
50 0.0116523370678996
51 0.0117631412891737
52 0.0118886632947669
53 0.0119557681126825
54 0.011777637110398
55 0.0114968255343375
56 0.0116742541480774
57 0.0116149952565022
58 0.011777697465044
59 0.011442931012363
60 0.0117234188877769
61 0.011505365178618
62 0.0114811496105094
63 0.0114762872724791
64 0.0115208732001936
65 0.0114228247036573
66 0.0113992868038129
67 0.0113656775872316
68 0.0113504720410837
69 0.0113257763770672
70 0.0112615923455011
71 0.0114092223758766
72 0.0111459373482922
73 0.0112192953548808
74 0.0112418061131675
75 0.0112361464428848
76 0.0110528667218488
77 0.0110769434749669
78 0.0111911927984142
79 0.0111560891934763
80 0.0111377279955535
81 0.0111306401033009
82 0.0111089321090828
83 0.0110904397082942
84 0.0111299327285315
85 0.0110904015162512
86 0.0110426265084301
87 0.0110366283884111
88 0.011178312082007
89 0.0114517560171512
90 0.0113326608073739
91 0.0110321754634484
92 0.0110713851028609
93 0.0108955029222534
94 0.0111426174134389
95 0.0109432575907185
96 0.0109069255115746
97 0.0107335520685136
98 0.0109973258253533
99 0.0109571118448465
100 0.0106928283850118
101 0.0106707124822528
102 0.0107273247876321
103 0.0109017056473639
104 0.0106784602033496
105 0.0108033750538453
106 0.0107031196222349
107 0.0105834992189316
108 0.0109568342552584
109 0.0107619483825705
110 0.0107913128419131
111 0.0108090488511544
112 0.0109877099676878
113 0.0108830646071633
114 0.0106190955598802
115 0.0108540028005766
116 0.0107165360869188
117 0.010842628117222
118 0.010846440649572
119 0.0108042020593336
120 0.0106164994115563
121 0.0107684964436985
122 0.0105836798626691
123 0.010653537312028
124 0.0106298972961896
125 0.0107578446837311
126 0.0105923503766557
127 0.0107328902630313
128 0.0106735393458596
129 0.0106956899055524
130 0.0105549616544847
131 0.0105458736071713
132 0.0106972761333525
133 0.010933680763854
134 0.0108403115406096
135 0.0106661210509065
136 0.0105183973946273
137 0.0104926430612927
138 0.0105138428080347
139 0.0106086382850311
140 0.0108266069536525
141 0.0105646033395613
142 0.0106944206861742
143 0.0106128867116572
144 0.0105447178204892
145 0.0107103927985492
146 0.0108378078971638
147 0.0106909826520373
148 0.0108098221176535
149 0.010524057126217
150 0.0105409167933047
151 0.0104247887104756
152 0.0105550243397765
153 0.0105164121916487
154 0.0108806849326201
155 0.0106939545507739
156 0.0108390435556051
157 0.0106959059920311
158 0.0105631059859395
159 0.0103619650732651
160 0.0103288710655582
161 0.0103879042051476
162 0.0104762557822495
163 0.0105500242668842
164 0.010459704848786
165 0.010579788627663
166 0.0107706614015668
167 0.010639292124029
168 0.010512174776947
169 0.0102444017364847
170 0.010302899195221
171 0.0102611723934877
172 0.010373146222182
173 0.010500708329749
174 0.0103539244267818
175 0.0105724948383585
176 0.0107997226761489
177 0.010468973187017
178 0.010400938073018
179 0.0103230983206395
180 0.0100803939427583
181 0.0103116652634305
182 0.0103052518398531
183 0.0102419910672447
184 0.0102530747589024
185 0.0100804311635377
186 0.0101417347494631
187 0.0103688324354601
188 0.0100769089645706
189 0.0101561571104101
190 0.0102194547650049
191 0.0103249807020498
192 0.0100615764901562
193 0.0102302962412142
194 0.0102625725031228
195 0.0101218043357806
196 0.0101492649498348
197 0.0100472035712619
198 0.0106202854220618
199 0.0100771210798047
};
\addlegendentry{Gradient Descent}
\addplot [semithick, darkorange]
table {%
0 0.0947284633646401
1 0.0911726681251005
2 0.0919067720782301
3 0.0836642274246425
4 0.0790858705114948
5 0.0881601064708624
6 0.08267408814902
7 0.071340506682638
8 0.065219888915994
9 0.0550267208039109
10 0.0703214049952694
11 0.0757628099300959
12 0.0586244319089726
13 0.0718568200471563
14 0.0540449276749669
15 0.0349675936773767
16 0.0429580175594724
17 0.0613749329521319
18 0.0559210703801519
19 0.0509671064225894
20 0.0531824840238518
21 0.0605086254730977
22 0.0507728348310015
23 0.0530126188930318
24 0.0512631748362776
25 0.0432589683187744
26 0.0429182451318896
27 0.0509394356051862
28 0.0521966738886863
29 0.0446230805439326
30 0.0405537301128595
31 0.0272958331224195
32 0.0310542895145012
33 0.0330082930556459
34 0.0238555246704082
35 0.0292294841561798
36 0.0205201019884121
37 0.0241844129004252
38 0.0313147927743436
39 0.0264528832131726
40 0.0305556600984637
41 0.039817182393441
42 0.0447778738587519
43 0.0370770126173489
44 0.0413995381399138
45 0.0339731494364799
46 0.0273679159330517
47 0.0286476202282182
48 0.023368175302634
49 0.0342870048665248
50 0.0342134851758014
51 0.0281623509524803
52 0.0266277578549343
53 0.0267220012346047
54 0.0255206163814984
55 0.0242433965459743
56 0.023364302875584
57 0.0216596426014379
58 0.019091147412198
59 0.0165998369667347
60 0.0146749126515591
61 0.0132991145781965
62 0.0122484335986378
63 0.0113093083252322
64 0.0103917644149554
65 0.00950419049807694
66 0.00868152116973875
67 0.00794708477102191
68 0.00730857328114243
69 0.00676409382094745
70 0.00630689271697149
71 0.00592778900633528
72 0.00561655344178336
73 0.0053628801572928
74 0.00515707922624742
75 0.0049904957709056
76 0.00485569014900789
77 0.0047464453614815
78 0.00465767087146063
79 0.00458525770495499
80 0.00452592099604276
81 0.00447705057795078
82 0.00443657927701657
83 0.0044028719356346
84 0.00437463459214027
85 0.00435084159343246
86 0.00433067782498653
87 0.00431349332699726
88 0.00429876781280869
89 0.00428608299372302
90 0.00427510104244133
91 0.00426554778446997
92 0.00425719962190929
93 0.00424987326610549
94 0.00424341772167447
95 0.00423770796613032
96 0.00423263995598844
97 0.00422812668853512
98 0.00422409504771881
99 0.00422048329682967
};
\addlegendentry{Kernel SOS}
\end{axis}

\end{tikzpicture}}
    \subcaption{Dimension 1}
    \label{fig:Henon_rho_0_idx}
\end{minipage}\hspace{0.1\textwidth}
\begin{minipage}[c]{0.40\textwidth}
    \centering
    \resizebox{\textwidth}{!}{
\begin{tikzpicture}

\definecolor{darkgray176}{RGB}{176,176,176}
\definecolor{darkorange}{RGB}{255,140,0}
\definecolor{lightgray204}{RGB}{204,204,204}
\definecolor{slategray}{RGB}{112,128,144}

\begin{axis}[
legend cell align={left},
legend style={fill opacity=0.8, draw opacity=1, text opacity=1, draw=lightgray204},
tick align=outside,
tick pos=left,
x grid style={darkgray176},
xlabel={Iterations},
xmin=-9.95, xmax=208.95,
xtick style={color=black},
y grid style={darkgray176},
ylabel={Rho Relative},
ymin=-0.000804997255364953, ymax=0.0301363018785793,
ytick style={color=black}
]
\addplot [semithick, slategray, dashed]
table {%
0 0.00713224854642247
1 0.00716133276153463
2 0.00731871243703075
3 0.00728003695422286
4 0.00713516753702204
5 0.00729442448468631
6 0.00713240082693778
7 0.0073668439284087
8 0.00709667357160715
9 0.00719282673120014
10 0.00711476946146039
11 0.00731474768715212
12 0.00707999095202605
13 0.00716469073849724
14 0.00723497618284219
15 0.00721320009116577
16 0.00739195361617107
17 0.0071303967050832
18 0.00708801095191403
19 0.00708251107422497
20 0.00713472127368486
21 0.00714092031100955
22 0.00708297267266667
23 0.00725943675484375
24 0.00727990903059317
25 0.00736667987278883
26 0.00739023209089074
27 0.00721584148653243
28 0.00710604200811749
29 0.00708440033670554
30 0.00713426132461548
31 0.0070517701063828
32 0.00705304987166389
33 0.00725311591537481
34 0.00706644617172825
35 0.00716916731321204
36 0.00727683758126241
37 0.00720765665763323
38 0.00699615066882919
39 0.00716451515019345
40 0.00715974835794508
41 0.00713246514407673
42 0.00721864252302729
43 0.00715593126608105
44 0.00719036948709473
45 0.00713335236631552
46 0.00718001336576279
47 0.0070464157167468
48 0.00699022358565771
49 0.00702825193961776
50 0.00708749688283472
51 0.00714821844761737
52 0.00708487211845632
53 0.00713032332833308
54 0.00710370595374998
55 0.00707406949505651
56 0.0070492598108276
57 0.00710555357875353
58 0.00709882277787055
59 0.00707162103167003
60 0.00703711358155612
61 0.00710796989553053
62 0.0070971979590444
63 0.0070981261902433
64 0.00710133286791936
65 0.00696573849926008
66 0.00697019445987468
67 0.00696740145604946
68 0.00694773526104731
69 0.0069570452939911
70 0.00707162833347563
71 0.00694482119408057
72 0.00690308821338115
73 0.00711616750884914
74 0.0069758299490621
75 0.00686885761781297
76 0.0070607071199752
77 0.00691427581659387
78 0.00706298760699964
79 0.00700393839583979
80 0.00686795635607451
81 0.00703750911020706
82 0.00688249711420996
83 0.00682243287917994
84 0.00694920196532778
85 0.00683123428435273
86 0.00675926383120595
87 0.00692074162010836
88 0.0070379541936948
89 0.00701520462970795
90 0.00684582509124221
91 0.00678921460062987
92 0.00677921217383459
93 0.00673245474636308
94 0.00684902960744505
95 0.00674784573089104
96 0.00683140392463699
97 0.00671883301622
98 0.0067172778103981
99 0.00673374806384153
100 0.00672682923913381
101 0.00675496424809108
102 0.00693760845845692
103 0.0068269222311409
104 0.00694040488847192
105 0.00695263395210688
106 0.00678163045303204
107 0.00672589026965342
108 0.00684634167466736
109 0.00689382668774519
110 0.00678888462635474
111 0.00692282698893787
112 0.0067396597397027
113 0.00673986662157644
114 0.0068194272401032
115 0.0067221620986595
116 0.00678287135561129
117 0.00675264108583695
118 0.00678120928460635
119 0.00682777064617024
120 0.00681682840939968
121 0.00682171715423219
122 0.00675415884916075
123 0.00677619224103487
124 0.00672522677776377
125 0.00673061705059341
126 0.00677221511031456
127 0.00671706340078404
128 0.00674079890266532
129 0.00678241581956218
130 0.00676371019574795
131 0.00674331502755354
132 0.00675990156401418
133 0.00675573013624653
134 0.00672134050178019
135 0.00671508335112181
136 0.00659357369977664
137 0.00671343747156383
138 0.00669400428616129
139 0.00664721545391112
140 0.00663031858033825
141 0.00662918361315568
142 0.00670831187810417
143 0.0065931686254076
144 0.0067910160271506
145 0.00676508250954699
146 0.00676099677529474
147 0.00665696492006063
148 0.00665078661403296
149 0.00665199843248554
150 0.0068160012616445
151 0.00682418837588816
152 0.00671284955599649
153 0.00672845779758047
154 0.00672313226326404
155 0.00672184497006345
156 0.00667786695233796
157 0.00678564370759738
158 0.00675718571709227
159 0.00669591904365108
160 0.00670070832684411
161 0.00678033395642286
162 0.0067932626352557
163 0.00691681333057503
164 0.00729984092297931
165 0.0074119549329601
166 0.0072664057279771
167 0.00713117441175404
168 0.00678621398670642
169 0.00663800364902722
170 0.00656145791819562
171 0.00662766268558079
172 0.00657633875506025
173 0.00653299976319555
174 0.00672781005361878
175 0.0068398681804609
176 0.00671558223157298
177 0.00684183924080617
178 0.00704471592315481
179 0.00698827553240311
180 0.0070057930105818
181 0.00721967675421498
182 0.00722089282220117
183 0.00697088939367296
184 0.00679118351519192
185 0.00687886128490978
186 0.00697582142917386
187 0.0067906638296189
188 0.00658303622639245
189 0.00687490768251764
190 0.0066731587280553
191 0.00653196570702208
192 0.00644691259690944
193 0.00658203079155983
194 0.00641318078044384
195 0.00639787770631073
196 0.00638258580751949
197 0.00635009332399905
198 0.00644831235130239
199 0.00657683802625531
};
\addlegendentry{Gradient Descent}
\addplot [semithick, darkorange]
table {%
0 0.0232746859799455
1 0.0287298791906727
2 0.0210604488069857
3 0.0215407680917583
4 0.0218576058227334
5 0.0153534736089127
6 0.0131793145497626
7 0.0198586791624055
8 0.0130874624335808
9 0.0118598298672649
10 0.0122439477521257
11 0.0133396045955476
12 0.0108017403845231
13 0.0162008047292709
14 0.0103573397257094
15 0.0142611301136369
16 0.0122039704990661
17 0.0106102628926914
18 0.0109566440014278
19 0.00981581953344357
20 0.00966098389530123
21 0.00798555567783665
22 0.00769033403083152
23 0.00719044310436656
24 0.00686867894031373
25 0.00482523530698376
26 0.0040342344317279
27 0.00361858033624252
28 0.00276898405351911
29 0.00216665956548323
30 0.00186794982169602
31 0.00212920759422708
32 0.00204411970713836
33 0.0019945323233399
34 0.00169151211449903
35 0.0018129548936977
36 0.00204052587250725
37 0.00158934534196664
38 0.0015661867861434
39 0.00144830817249808
40 0.00132157073175654
41 0.00128207113002587
42 0.00115029343780582
43 0.00106301882413173
44 0.00113128653275396
45 0.00105393975945223
46 0.001010176757818
47 0.00102182815399421
48 0.000954901071548941
49 0.000936954119684752
50 0.000910131575143969
51 0.000898325270238698
52 0.000930001875301123
53 0.000892817972044102
54 0.000811867936146404
55 0.000760972818935945
56 0.000748264050597691
57 0.000747746039034047
58 0.000743218270775392
59 0.000734048852894298
60 0.000723464619797443
61 0.000712618946287713
62 0.000701058588434056
63 0.000688790696330543
64 0.000676719511590318
65 0.000665795314020556
66 0.000656422009625679
67 0.000648498468219572
68 0.000641705886396116
69 0.000635738179948997
70 0.000630402960650533
71 0.000625624191321839
72 0.000621396555597675
73 0.000617732832704831
74 0.000614628400520401
75 0.000612049036655016
76 0.000609935916287441
77 0.000608217841466696
78 0.000606823031048087
79 0.000605687019687218
80 0.000604756391321515
81 0.000603989474813105
82 0.00060335530698219
83 0.000602831821698624
84 0.000602403838805388
85 0.000602061145789312
86 0.00060179679620187
87 0.000601605682779449
88 0.000601483403566472
89 0.000601425432541602
90 0.000601426592487408
91 0.000601480808749777
92 0.00060158110922004
93 0.000601719815182422
94 0.000601888855488175
95 0.000602080137651617
96 0.000602285911388223
97 0.000602499081039931
98 0.000602713434078361
99 0.000602923777636488
};
\addlegendentry{Kernel SOS}
\end{axis}

\end{tikzpicture}}
    \subcaption{Dimension 2}
    \label{fig:Henon_rho_1_idx}
\end{minipage}
\caption{Rho Relative on Henon Map Training Set.}
\label{fig:Henon_rho}
\end{figure}

\begin{figure}
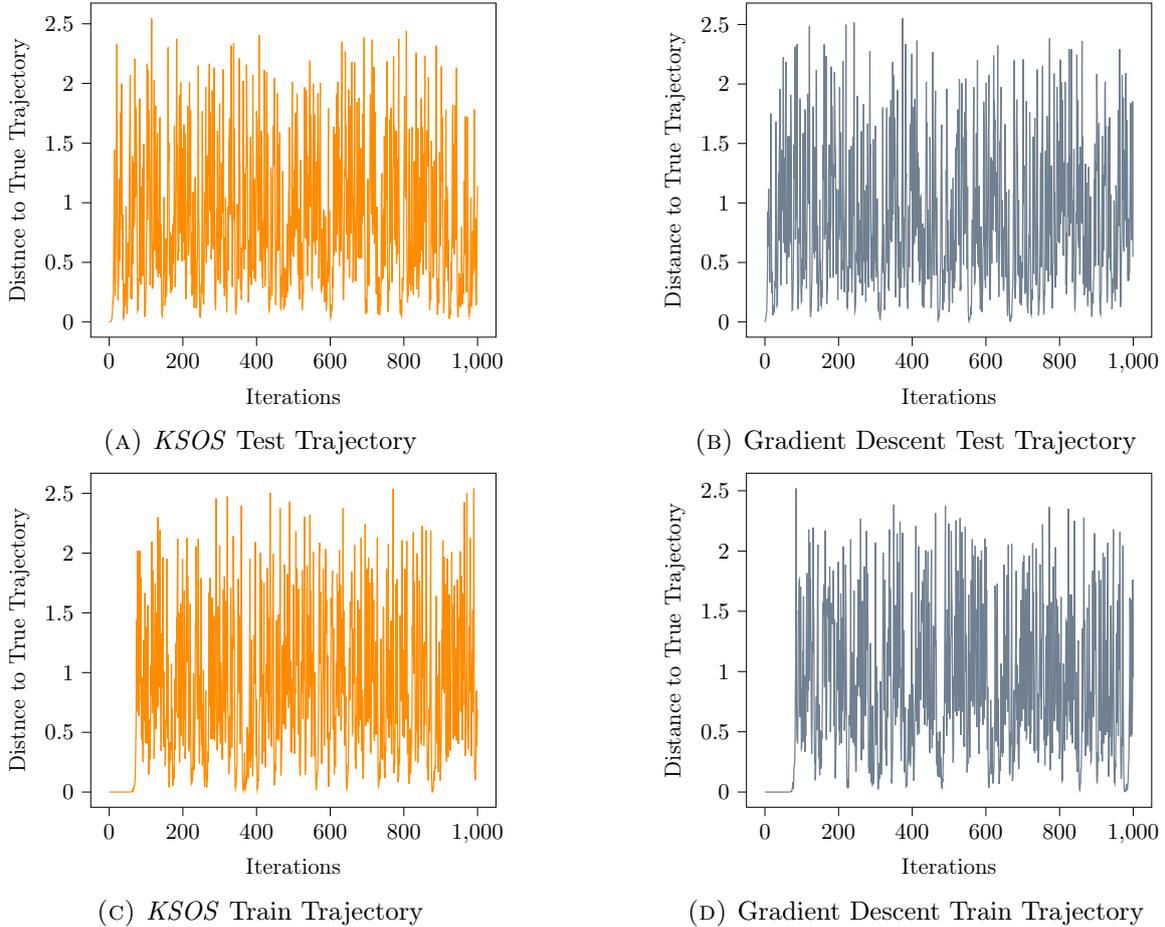

\begin{minipage}[c]{0.40\textwidth}
    \centering
    \resizebox{\textwidth}{!}{\input{figs/Henon_errTraj_sos}}
    \subcaption{\emph{KSOS} Test Trajectory}
    \label{fig:Henon_errTraj_sos}
\end{minipage}\hspace{0.1\textwidth}
\begin{minipage}[c]{0.40\textwidth}
    \centering
    \resizebox{\textwidth}{!}{\input{figs/Henon_errTraj_flow}}
    \subcaption{Gradient Descent Test Trajectory}
    \label{fig:figs/Henon_errTraj_flow}
\end{minipage}

\vspace{0.5cm}

\begin{minipage}[c]{0.40\textwidth}
    \centering
    \resizebox{\textwidth}{!}{\input{figs/Henon_errTrajTrain_sos}}
    \subcaption{\emph{KSOS} Train Trajectory}
    \label{fig:Henon_errTrajTrain_sos}
\end{minipage}\hspace{0.1\textwidth}
\begin{minipage}[c]{0.40\textwidth}
    \centering
    \resizebox{\textwidth}{!}{\input{figs/Henon_errTrajTrain_flow}}
    \subcaption{Gradient Descent Train Trajectory}
    \label{fig:Henon_errTrajTrain_flow}
\end{minipage}
\caption{Distances of Predicted Trajectory to True Trajectory for Henon Map.}
\label{fig:Henon_ErrTraj}
\end{figure}

\subsection{Lorentz System}

The continuous version of the Lorentz map is given by
\begin{align*}
    \dot{x} &= 10 (y - x) \\
    \dot{y} &= 28 x - y - x z \\
    \dot{z} &= x y - \frac{10}{3} z.
\end{align*}
To obtain the discrete system, we use the standard forward Euler method with a step size of $10^{-2}$.
For both the training and test trajectories, we let the number of steps be $1000$. To construct the training set we let $X_0^{(train)} = (0.5, 1.5, 2.5)$ and for the test set we let $X_0^{(test)} = (0.7, 1.1, 2)$. We set the learning rate of gradient descent to $\eta = 0.5$.

\begin{figure}
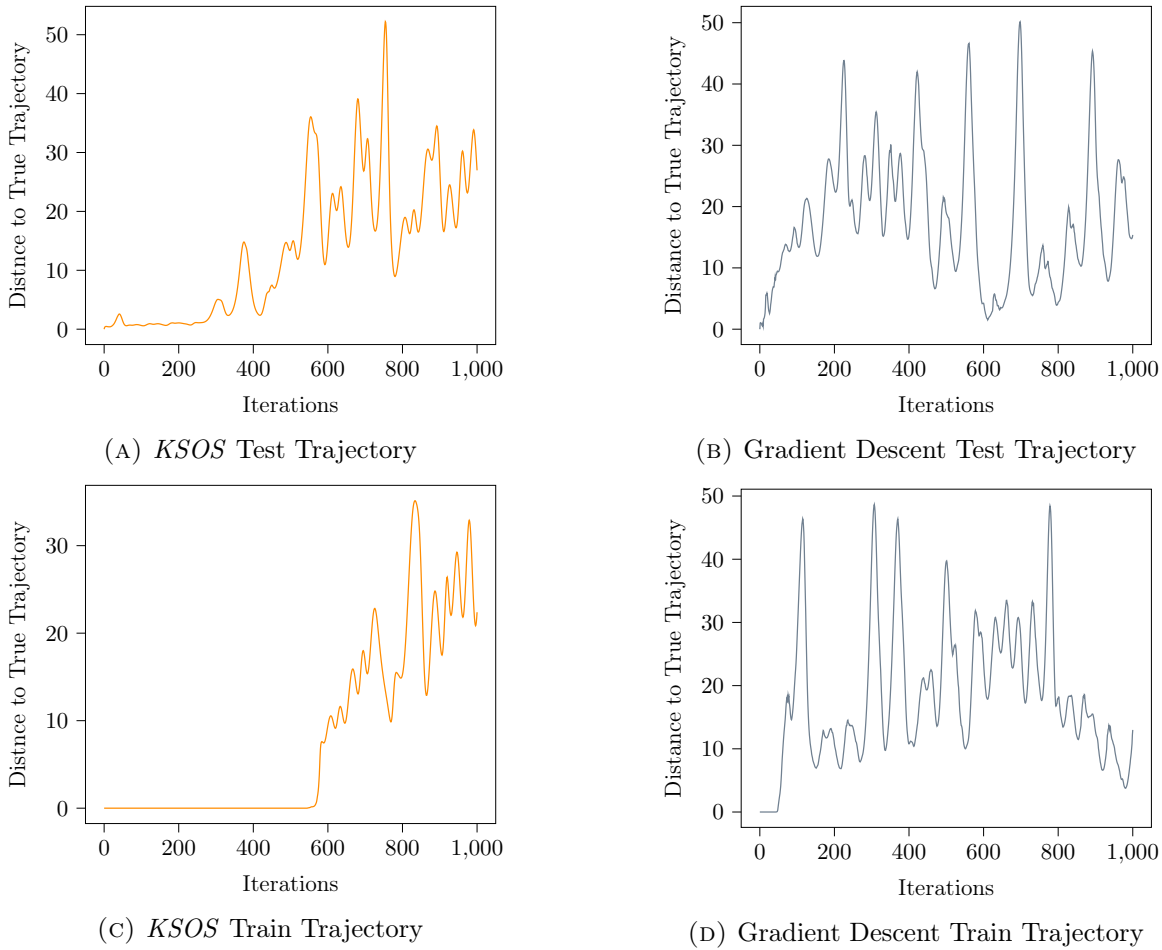

\begin{minipage}[c]{0.40\textwidth}
    \centering
    \resizebox{\textwidth}{!}{\input{figs/Lorentz_errTraj_sos}}
    \subcaption{\emph{KSOS} Test Trajectory}
    \label{fig:Lorentz_errTraj_sos}
\end{minipage}\hspace{0.1\textwidth}
\begin{minipage}[c]{0.40\textwidth}
    \centering
    \resizebox{\textwidth}{!}{\input{figs/Lorentz_errTraj_flow}}
    \subcaption{Gradient Descent Test Trajectory}
    \label{fig:figs/Lorentz_errTraj_flow}
\end{minipage}

\vspace{0.5cm}

\begin{minipage}[c]{0.40\textwidth}
    \centering
    \resizebox{\textwidth}{!}{\input{figs/Lorentz_errTrajTrain_sos}}
    \subcaption{\emph{KSOS} Train Trajectory}
    \label{fig:Lorentz_errTrajTrain_sos}
\end{minipage}\hspace{0.1\textwidth}
\begin{minipage}[c]{0.40\textwidth}
    \centering
    \resizebox{\textwidth}{!}{\input{figs/Lorentz_errTrajTrain_flow}}
    \subcaption{Gradient Descent Train Trajectory}
    \label{fig:Lorentz_errTrajTrain_flow}
\end{minipage}
\caption{Distances of Predicted Trajectory to True Trajectory for Lorentz Map.}
\label{fig:Lorentz_ErrTraj}
\end{figure}

\begin{figure}
\begin{minipage}[c]{0.32\textwidth}
    \centering
    \includegraphics[width=1\textwidth]{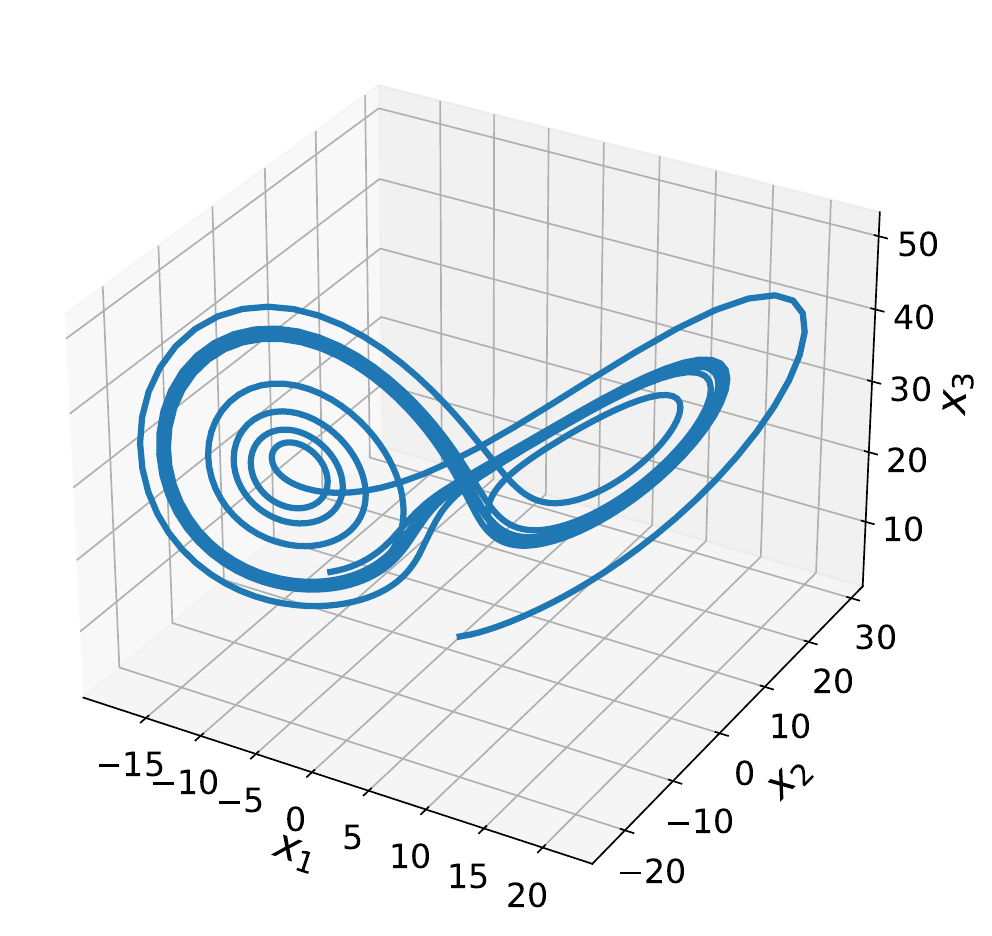}
    \subcaption{True Test Trajectory}
    \label{fig:Lorentz_Traj_true.1}
\end{minipage}\hfill
\begin{minipage}[c]{0.32\textwidth}
    \centering
    \includegraphics[width=1\textwidth]{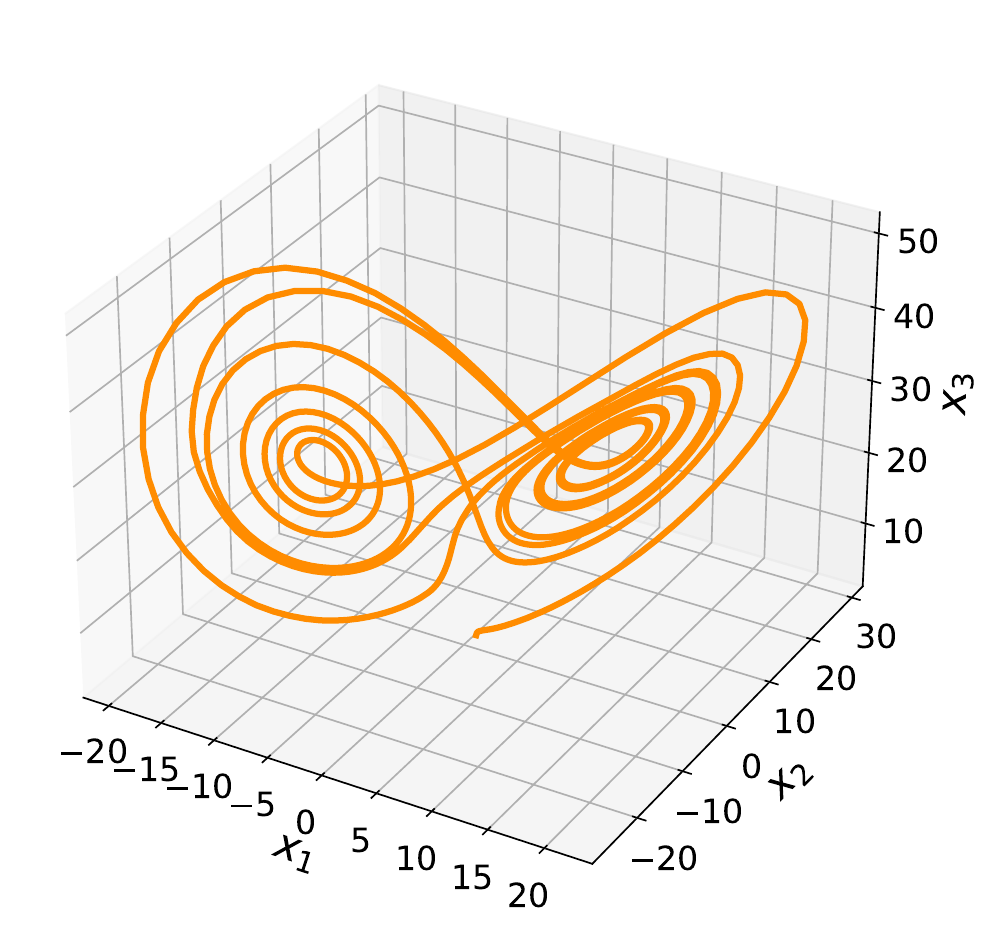}
    \subcaption{\emph{KSOS} Test Trajectory}
    \label{fig:Lorentz_Traj_sos}
\end{minipage}\hfill
\begin{minipage}[c]{0.32\textwidth}
    \centering
    \includegraphics[width=1\textwidth]{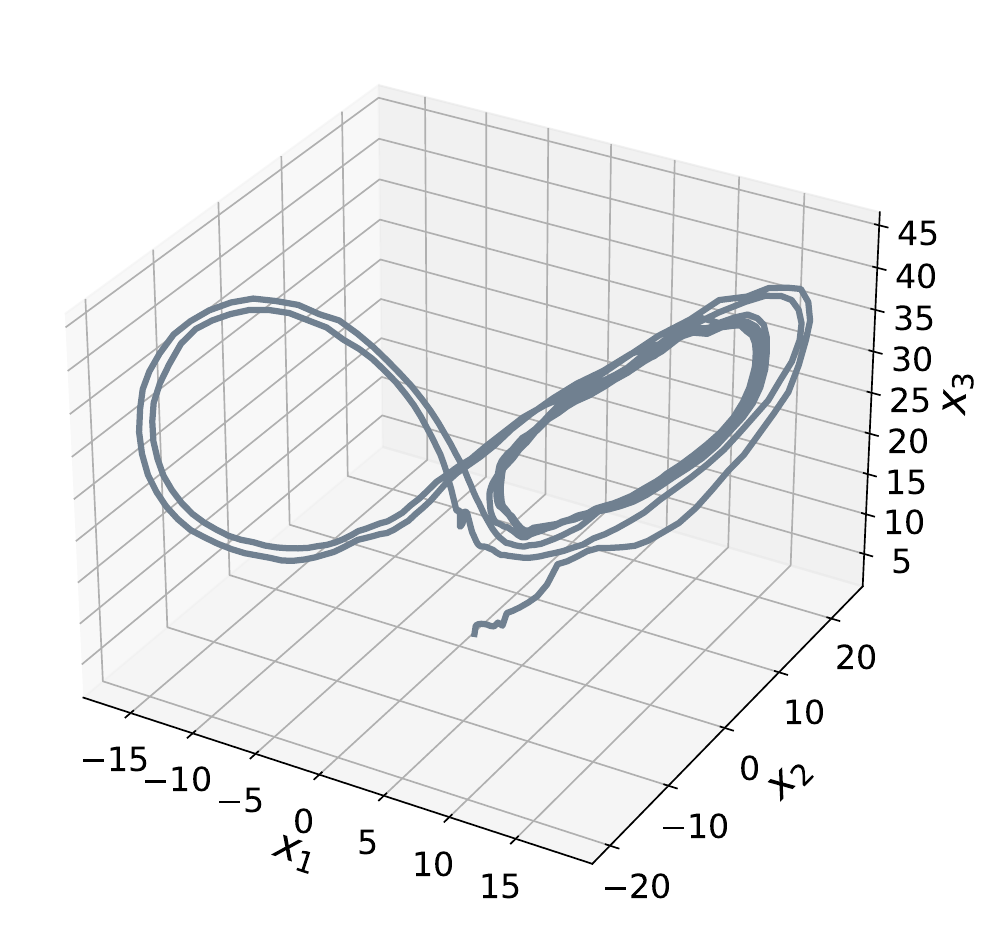}
    \subcaption{Gradient Descent Test Trajectory}
    \label{fig:Lorentz_Traj_flow}
\end{minipage}\hfill

\vspace{0.5cm}

\begin{minipage}[c]{0.32\textwidth}
    \centering
    \includegraphics[width=1\textwidth]{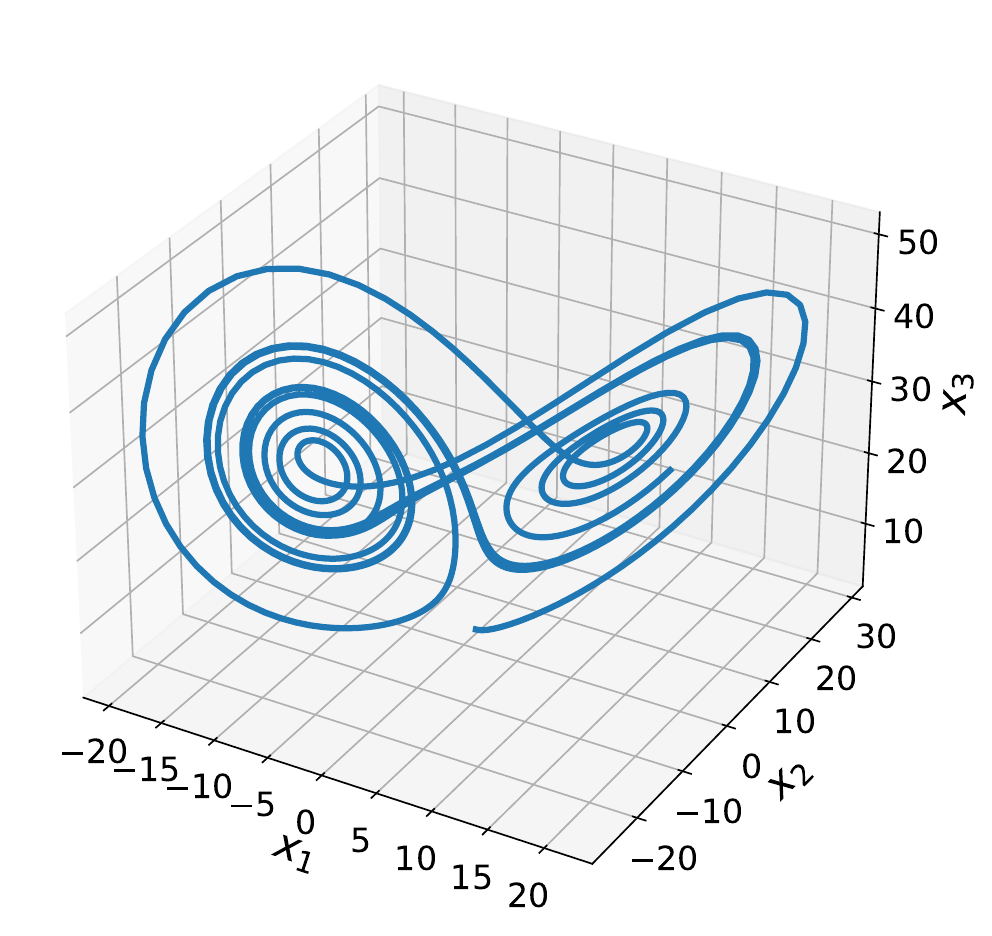}
    \subcaption{True Train Trajectory}
    \label{fig:Lorentz_TrajTrain_true}
\end{minipage}\hfill
\begin{minipage}[c]{0.32\textwidth}
    \centering
    \includegraphics[width=1\textwidth]{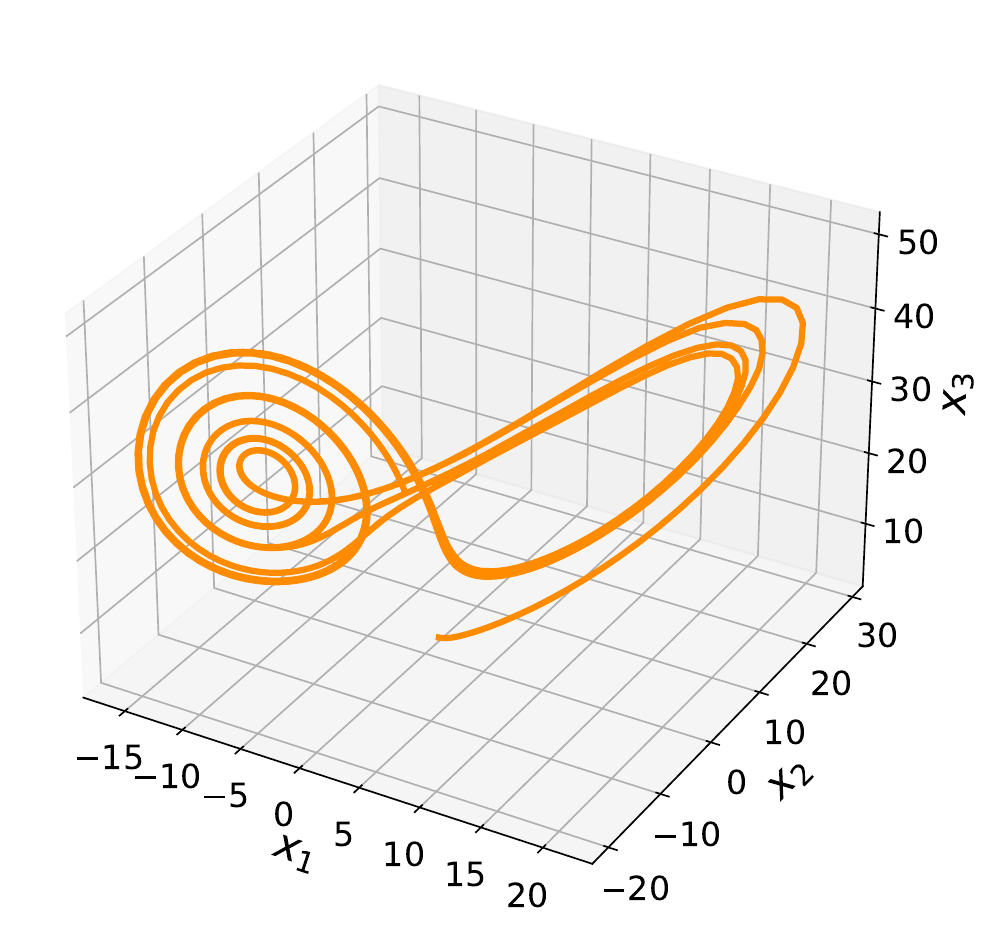}
    \subcaption{\emph{KSOS} Train Trajectory}
    \label{fig:Lorentz_TrajTrain_sos}
\end{minipage}\hfill
\begin{minipage}[c]{0.32\textwidth}
    \centering
    \includegraphics[width=1\textwidth]{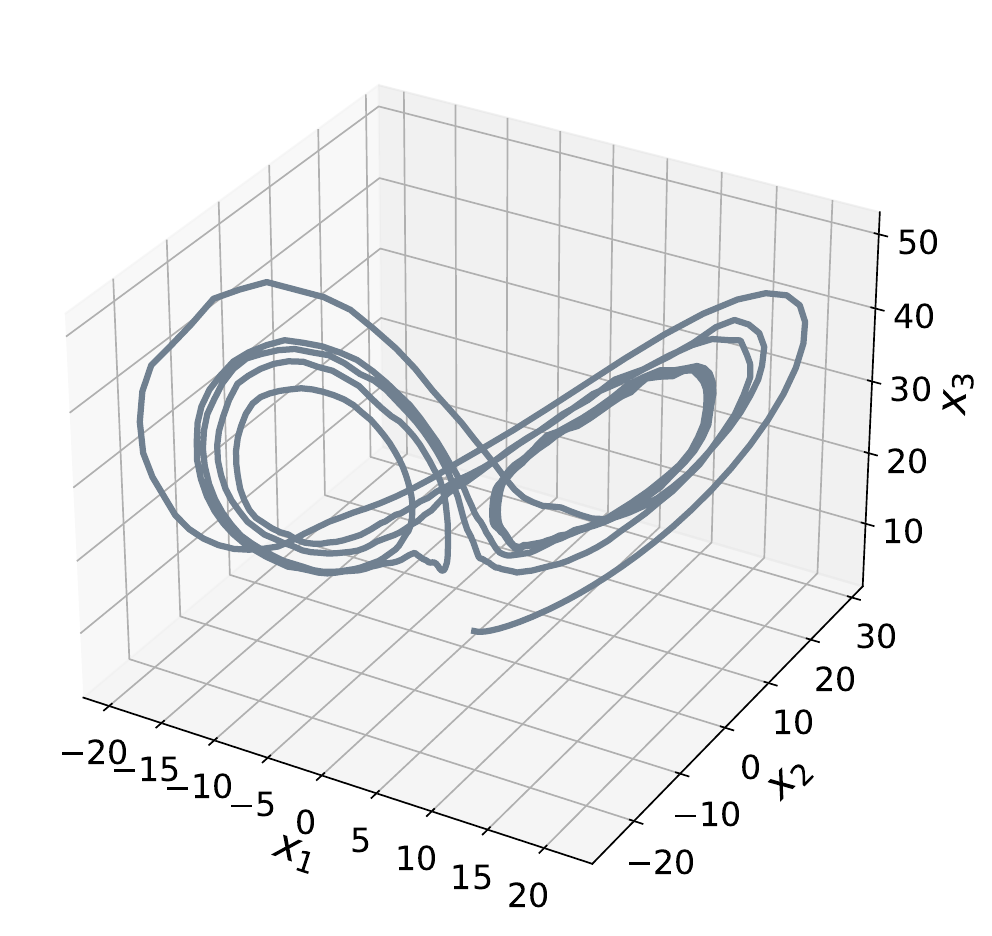}
    \subcaption{Gradient Descent Train Trajectory}
    \label{fig:Lorentz_TrajTrain_flow}
\end{minipage}
\caption{Predicted Trajectories for Lorentz Dynamics.}
\label{fig:Lorentz_Traj}
\end{figure}

In Table \ref{tab:rhoLorentz}, we see that the performance of \emph{KSOS} as a minimizer of $\rho$ is almost three times better than of gradient descent and that the 75th percentile of \emph{KSOS} is lower than the 25th percentile of the gradient descent. To reason whether this translates to a better kernel, we consider Table \ref{tab:testLorentz} on the test trajectory. While the Mean Error's 25th percentile of gradient descent is larger than the 75th percentile of \emph{KSOS}, the difference in the median value is not as pronounced. However, observing the number of iterations, the predicted trajectory stays true is much larger for \emph{KSOS}, specifically when $\gamma = 0.25$. To better understand this consider Figure \ref{fig:Lorentz_ErrTraj}, where for \emph{KSOS} only minor fluctuations in the error occur for the first few hundred iterations. We also observe more pronounced improvements for \emph{KSOS} in the Hausdorff Distance, which is explained by the behavior of the trajectories in Figure \ref{fig:Lorentz_Traj}. Similar results hold on the Training Trajectory, with the exception of the Mean Error being close to zero, as expected.

\begin{table}[H]
    \centering
    \begin{tabular}{l r r} 
        \toprule
        \textbf{Measure} & \textbf{KSOS} & \textbf{Gradient Descent}  \\
        \midrule
        Relative $\rho$ & $2.98 \ [2.07, 4.41] \times 10^{-2}$ & $6.59 \ [4.73, 10.63] \times 10^{-2}$ \\  
        \bottomrule
    \end{tabular}
    \caption{Relative $\rho$ Statistics for the Lorentz Map.}
    \label{tab:rhoLorentz}
\end{table}

\begin{table}[H]
    \centering
    \begin{tabular}{l r r} 
        \toprule
        \textbf{Measure} & \textbf{KSOS} & \textbf{Gradient Descent}  \\
        \midrule
        Mean Error & $1.10 \ [0.93, 1.32] \times 10^{-1}$ & $2.30 \ [2.14, 3.00] \times 10^{-1}$ \\ HD & $6.32 \ [5.52, 6.44]$ & $8.57 \ [6.72, 12.75]$ \\ Deviation(0.1) & $1.00$ & $1.00$ \\ Deviation(0.25) & $3.32 \ [3.14, 3.46] \times 10^{2}$ & $6.15 \ [4.42, 16.57] \times 10^{1}$ \\ 
        \bottomrule
    \end{tabular}
    \caption{Test Set Statistics for the Lorentz Map.}
    \label{tab:testLorentz}
\end{table}

\begin{table}[H]
    \centering
    \begin{tabular}{l r r} 
        \toprule
        \textbf{Measure} & \textbf{KSOS} & \textbf{Gradient Descent}  \\
        \midrule
        Mean Error & $1.17 \ [1.07, 1.21] \times 10^{-13}$ & $4.29 \ [3.66, 4.59] \times 10^{-14}$ \\ HD & $7.38 \ [3.17, 9.25]$ & $9.46 \ [7.96, 11.15]$ \\ Deviation(0.1) & $5.57 \ [3.44, 7.41] \times 10^{2}$ & $2.37 \ [0.64, 3.16] \times 10^{2}$ \\ Deviation(0.25) & $3.71 \ [0.81, 5.79] \times 10^{2}$ & $2.74 \ [1.32, 3.27] \times 10^{2}$ \\ 
        \bottomrule
    \end{tabular}
    \caption{Training Set Statistics for the Lorentz Map.}
    \label{tab:trainLorentz}
\end{table}

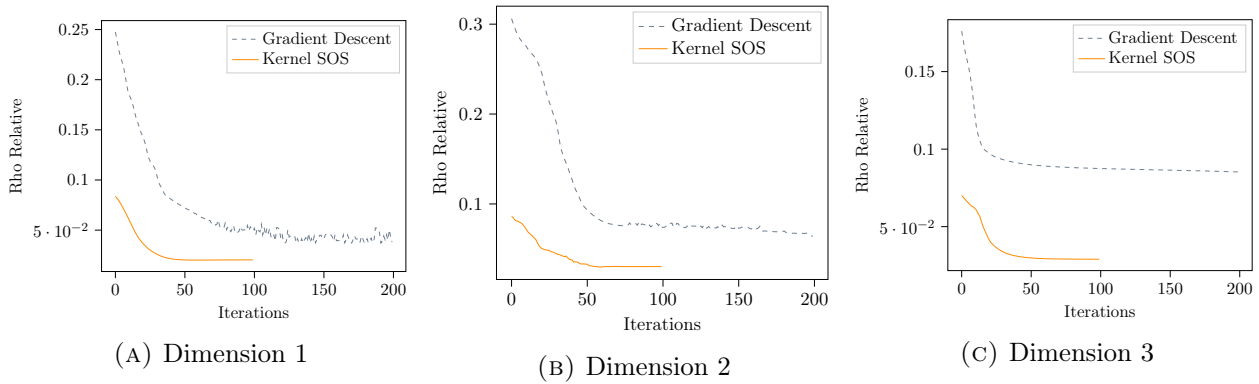
\begin{figure}
\begin{minipage}[c]{0.32\textwidth}
    \centering
    \resizebox{\textwidth}{!}{
\begin{tikzpicture}

\definecolor{darkgray176}{RGB}{176,176,176}
\definecolor{darkorange}{RGB}{255,140,0}
\definecolor{lightgray204}{RGB}{204,204,204}
\definecolor{slategray}{RGB}{112,128,144}

\begin{axis}[
legend cell align={left},
legend style={fill opacity=0.8, draw opacity=1, text opacity=1, draw=lightgray204},
tick align=outside,
tick pos=left,
x grid style={darkgray176},
xlabel={Iterations},
xmin=-9.95, xmax=208.95,
xtick style={color=black},
y grid style={darkgray176},
ylabel={Rho Relative},
ymin=0.00872237596087575, ymax=0.258920740605668,
ytick style={color=black}
]
\addplot [semithick, slategray, dashed]
table {%
0 0.247548087667268
1 0.240650625021187
2 0.233343365066718
3 0.226340399534875
4 0.221684428715886
5 0.21690732252074
6 0.21195596510864
7 0.20445841099618
8 0.19629462145945
9 0.188609975008819
10 0.184597489483497
11 0.181569269405824
12 0.177399913837026
13 0.172693500684044
14 0.166661753766668
15 0.161341476219228
16 0.157693573685143
17 0.153550793591995
18 0.148889359643356
19 0.145494211440879
20 0.141571127969102
21 0.138534094487431
22 0.133645910072836
23 0.126790376943298
24 0.121879101644351
25 0.118652208807019
26 0.116456428940309
27 0.114696597725008
28 0.112514010472098
29 0.109390039928339
30 0.103529147294717
31 0.0981371480520009
32 0.0950149824524062
33 0.0927513313895887
34 0.0897067912066748
35 0.0868829143895881
36 0.0847267981096538
37 0.0836903993212909
38 0.0827573586280013
39 0.0816964001923651
40 0.0808332920844379
41 0.0798592065319773
42 0.0789902681665726
43 0.0780425189571021
44 0.0771691383244698
45 0.0763451099637262
46 0.0754802753066599
47 0.0746183112096913
48 0.0738225302661438
49 0.0727988807315836
50 0.0722083463848187
51 0.0713482914233707
52 0.0706931112127287
53 0.0701442581369526
54 0.0687121760710649
55 0.0682658049053493
56 0.0675889551526585
57 0.0665918527508137
58 0.0663526365659539
59 0.0647817950196028
60 0.0644103360192331
61 0.0639410821114567
62 0.0625056344346736
63 0.0622484255062801
64 0.0614339212473624
65 0.061533756363231
66 0.059365689081623
67 0.0592957706952078
68 0.0596167083132845
69 0.0571404728557432
70 0.0573301152239919
71 0.0577293413073767
72 0.0550391431681029
73 0.0558568806029408
74 0.0601546369450615
75 0.0539788823208365
76 0.052921884052243
77 0.0530395160753586
78 0.0519991628121842
79 0.056896682679496
80 0.0505867520322172
81 0.0535745056466345
82 0.0512523805178816
83 0.0485298935666304
84 0.0512912335027812
85 0.0547561711668407
86 0.0482484896376962
87 0.0493589777105358
88 0.0478057242482534
89 0.0491203226378536
90 0.0536233092069627
91 0.0503403539965366
92 0.0499839874537421
93 0.0504464405781185
94 0.0492425847888691
95 0.0545272732843319
96 0.0489907840703006
97 0.0532783800583744
98 0.0512852538097833
99 0.0490042449378127
100 0.0519243336554021
101 0.0451588870257172
102 0.0521515664083978
103 0.0493822665878958
104 0.0482995666745416
105 0.055708405797662
106 0.0454306623816696
107 0.0525790051107602
108 0.0451756660621369
109 0.0519762030064733
110 0.0471890921193411
111 0.0514517504584732
112 0.0425187213197713
113 0.0457621128282867
114 0.0449358621014706
115 0.0391115678172282
116 0.0525696420055769
117 0.0521217865893935
118 0.0510765154273503
119 0.041266940016856
120 0.0451348226652591
121 0.0374156424114088
122 0.0373050200365109
123 0.0385394998254002
124 0.0503119423235912
125 0.0431127884590571
126 0.0443725292571039
127 0.044881325208462
128 0.0443342405009589
129 0.0440134901685253
130 0.0373805759862917
131 0.041400879808648
132 0.037174148312755
133 0.0374682611825444
134 0.0413309685278888
135 0.0371671163687018
136 0.0402755171564966
137 0.0424493171828684
138 0.038153507747496
139 0.0469930785183068
140 0.0374741553461306
141 0.0414916475881688
142 0.0375569277258713
143 0.0439590486558377
144 0.0374361165664978
145 0.0410343502257271
146 0.0413082105063881
147 0.0448439467103011
148 0.0447850704478022
149 0.0411670505376109
150 0.0410749916336438
151 0.0453382594376947
152 0.046547405956106
153 0.0453198364874785
154 0.045267068013642
155 0.0453164653333705
156 0.043379820193911
157 0.0390376807314653
158 0.0453941017462662
159 0.0394350869725866
160 0.0395930970505929
161 0.0395234437904594
162 0.0398090360224296
163 0.0402959670465941
164 0.0452100093430322
165 0.0374667720189336
166 0.0411404047880214
167 0.0452098734900241
168 0.0412578822230937
169 0.0375263674828531
170 0.0371578316136312
171 0.0405537955237651
172 0.0443050895354603
173 0.0411666175099891
174 0.0412452728586521
175 0.0372996965272496
176 0.0380655382599963
177 0.044153845293312
178 0.041303112854876
179 0.0375254439511409
180 0.0374400617032726
181 0.0413090026909466
182 0.0371358699551423
183 0.0401144465538598
184 0.0419597975179901
185 0.0374716985493962
186 0.0394807255797188
187 0.0521955685843282
188 0.0520345846363401
189 0.0519274173443399
190 0.0470181975980357
191 0.0371662703432213
192 0.0393122482218574
193 0.0441963656513543
194 0.040503455650083
195 0.0454267664790484
196 0.040075833691376
197 0.0502891027499462
198 0.0433149148456696
199 0.0380805935750693
};
\addlegendentry{Gradient Descent}
\addplot [semithick, darkorange]
table {%
0 0.0837730153636235
1 0.0821129885128434
2 0.0803079983503897
3 0.0782378470653293
4 0.0758693118313007
5 0.0734438905143241
6 0.0708384919108728
7 0.0681851207304252
8 0.0655659986143058
9 0.0627014603973728
10 0.059846057284993
11 0.0569809578352399
12 0.0542404515569058
13 0.0513085543771884
14 0.0485005990107272
15 0.046011342596731
16 0.0437521678952646
17 0.0416997392759817
18 0.0397600100033447
19 0.0380279409532911
20 0.0365002498868413
21 0.0350527853352615
22 0.0336640075398875
23 0.0323311043529636
24 0.0311529040993176
25 0.0300586820353196
26 0.0290694393305894
27 0.0281780442053428
28 0.0273562855541278
29 0.0265998318698747
30 0.025895342362189
31 0.0252140456664272
32 0.0245869100405361
33 0.0240121385875914
34 0.0235090536146815
35 0.023072170573967
36 0.0226770479795187
37 0.0223258336181933
38 0.0220065942376375
39 0.0217284922502067
40 0.021484330414853
41 0.0212702237822835
42 0.021080821671151
43 0.0209102499268934
44 0.0207587836971015
45 0.020635933466272
46 0.0205152323690564
47 0.0204293680438469
48 0.0203517609278052
49 0.020286288101788
50 0.0202334529969684
51 0.0201900867950099
52 0.0201604642962149
53 0.0201365463991372
54 0.0201192795867681
55 0.0201065415111444
56 0.0200991977293526
57 0.0200950288992754
58 0.0200966130230004
59 0.0200996947565928
60 0.0201020394449184
61 0.020109239388145
62 0.0201156945603991
63 0.0201233393114989
64 0.0201332933450437
65 0.020143310616174
66 0.0201563650194095
67 0.020163279497484
68 0.0201734422597699
69 0.0201833216841469
70 0.0201948348153553
71 0.0202016287993606
72 0.0202108925527991
73 0.0202196907495277
74 0.0202272867302085
75 0.0202352965739661
76 0.0202434169360923
77 0.020251516345398
78 0.0202581853641312
79 0.0202650025036417
80 0.0202706773999805
81 0.0202759558488466
82 0.0202816038522635
83 0.0202873741657703
84 0.0202923912931955
85 0.0202969231397012
86 0.0203011903945625
87 0.0203053993415323
88 0.0203108263653174
89 0.020312643274688
90 0.0203157758113991
91 0.0203186267642735
92 0.020320837825093
93 0.0203236712051814
94 0.0203263678320339
95 0.020329192065798
96 0.0203315241970978
97 0.0203330556736068
98 0.0203348371224534
99 0.0203363692492319
};
\addlegendentry{Kernel SOS}
\end{axis}

\end{tikzpicture}}
    \subcaption{Dimension 1}
    \label{fig:Lorentz_rho_0_idx}
\end{minipage}
\begin{minipage}[c]{0.32\textwidth}
    \centering
    \resizebox{\textwidth}{!}{
\begin{tikzpicture}

\definecolor{darkgray176}{RGB}{176,176,176}
\definecolor{darkorange}{RGB}{255,140,0}
\definecolor{lightgray204}{RGB}{204,204,204}
\definecolor{slategray}{RGB}{112,128,144}

\begin{axis}[
legend cell align={left},
legend style={fill opacity=0.8, draw opacity=1, text opacity=1, draw=lightgray204},
tick align=outside,
tick pos=left,
x grid style={darkgray176},
xlabel={Iterations},
xmin=-9.95, xmax=208.95,
xtick style={color=black},
y grid style={darkgray176},
ylabel={Rho Relative},
ymin=0.0159681627810637, ymax=0.320001720956459,
ytick style={color=black}
]
\addplot [semithick, slategray, dashed]
table {%
0 0.306182013766668
1 0.301248715972343
2 0.296617572856421
3 0.291838241179806
4 0.288640184610037
5 0.284950960927494
6 0.282559600474821
7 0.280353886509807
8 0.27848377472991
9 0.276887027939087
10 0.274463989078291
11 0.27184611136178
12 0.269579709597496
13 0.267871692640349
14 0.265952024046859
15 0.263981511857618
16 0.261181361724993
17 0.258480649137734
18 0.254730051605287
19 0.249616955191696
20 0.245319166971822
21 0.237635692027078
22 0.230195613750343
23 0.223782566725052
24 0.217938262139186
25 0.212788511234394
26 0.208151667343934
27 0.202469112673954
28 0.198033018107384
29 0.193801384107312
30 0.186963778521244
31 0.177400812101999
32 0.166562103733192
33 0.159359400025466
34 0.154504196021221
35 0.149743693066159
36 0.145561998531834
37 0.141598301378422
38 0.136352419984149
39 0.131242004437681
40 0.126953311288164
41 0.121826966160028
42 0.116912858040571
43 0.113440681536305
44 0.110559981447069
45 0.106133682115471
46 0.102154920265663
47 0.0987298500604445
48 0.0967936441038827
49 0.0949658254505495
50 0.0932327100220779
51 0.0915978641131762
52 0.0900386183945948
53 0.0885898993889976
54 0.0872002903728556
55 0.0859300156042087
56 0.0846916767669248
57 0.0835847052595654
58 0.082611289882804
59 0.0815642041006633
60 0.0807337085127612
61 0.0800708682104855
62 0.0793949615065418
63 0.0788071766458157
64 0.0781827085617822
65 0.0777181813826914
66 0.0775946425224217
67 0.0770338621103412
68 0.0767687845891755
69 0.0764816192625556
70 0.0763285977248986
71 0.0760776107780411
72 0.075925250104945
73 0.0758693060215873
74 0.0756817640089249
75 0.0756085482291071
76 0.0757056669918461
77 0.0767257156543587
78 0.0788528082174405
79 0.0766102270207428
80 0.0777874708378747
81 0.0782396888609059
82 0.0782656997120486
83 0.0763866208481916
84 0.0790878936951203
85 0.0755420698767393
86 0.0777766753576875
87 0.0772674337817468
88 0.0769086296027662
89 0.074511386454185
90 0.0747275705339794
91 0.0765504332222413
92 0.0745246769259655
93 0.0758531539712741
94 0.0788082514089801
95 0.0786473214589357
96 0.0781777303225456
97 0.076020484947553
98 0.0782689887541443
99 0.0741075697097884
100 0.0739021392267849
101 0.0741162398143955
102 0.0752171751233465
103 0.0783671261951115
104 0.0783339533060277
105 0.0784599014725289
106 0.0785373724057417
107 0.0739758773799477
108 0.0753431550550019
109 0.0781792973924457
110 0.078087969717497
111 0.0756077677960476
112 0.0762649291585947
113 0.0743068270692129
114 0.0737419362000575
115 0.0737165544477882
116 0.0754472634072942
117 0.0732998253146903
118 0.0739223834045627
119 0.0754360335927192
120 0.0753818690088889
121 0.0753398058062537
122 0.0752617887284367
123 0.0770993953971528
124 0.0755166354822627
125 0.0752497811807893
126 0.0735560290247989
127 0.0766362388065275
128 0.0731632329084826
129 0.0731404541669904
130 0.074759153062977
131 0.0735523773746463
132 0.0748657039015488
133 0.0748263614450982
134 0.0736066908447315
135 0.0731717940984611
136 0.0747052269659978
137 0.0727670201972397
138 0.0736716192439767
139 0.0765408062492525
140 0.0745854124964584
141 0.0747895131382807
142 0.0745033316707172
143 0.0744618363155463
144 0.0745603165353934
145 0.0731014409289754
146 0.0726552039213634
147 0.0726103319869462
148 0.0743596736751628
149 0.0759444329737202
150 0.0729776390756575
151 0.073197559244836
152 0.0742431300322756
153 0.0742096175704139
154 0.0741425420771298
155 0.0741440543086863
156 0.0740759760015377
157 0.0730034417952006
158 0.0721335083750334
159 0.0728715075380669
160 0.0727790746147747
161 0.0740240338289407
162 0.0722121080899845
163 0.0734359702290167
164 0.0752408128648115
165 0.0714527108001882
166 0.0696711112202277
167 0.0699569145237671
168 0.0698639403484256
169 0.0696276666850044
170 0.0695878232006155
171 0.0698836587172769
172 0.069926383383121
173 0.0690440771060025
174 0.0691645007380556
175 0.0689989839831207
176 0.0695399998760445
177 0.0693246023418638
178 0.0696845001568088
179 0.069501519864544
180 0.0699055809783581
181 0.0685510252903607
182 0.0675753380050386
183 0.0675513784323939
184 0.0674625751512195
185 0.0673626823479706
186 0.0674651646732879
187 0.0674235276625148
188 0.0672788529094469
189 0.067380971595147
190 0.0673661590143815
191 0.0673529201388706
192 0.0673233679241126
193 0.0671989168046198
194 0.0671736149464552
195 0.067577597829797
196 0.066577306666496
197 0.0648219898399673
198 0.0642491845010248
199 0.0641896031530035
};
\addlegendentry{Gradient Descent}
\addplot [semithick, darkorange]
table {%
0 0.0860538676724592
1 0.0849389221173911
2 0.0824424562678887
3 0.0811140281627907
4 0.0808822857209147
5 0.080059323356327
6 0.0791018855109302
7 0.077804490218555
8 0.0759236656397027
9 0.0735915390195838
10 0.0710444706737791
11 0.0685565779393393
12 0.0669512380365511
13 0.0655945237528521
14 0.0639215251310977
15 0.0622278786429434
16 0.0602706509392581
17 0.0572191413172293
18 0.0537642471973931
19 0.05173510546608
20 0.0503906307394744
21 0.0497679307580996
22 0.0489851510916143
23 0.048887799668328
24 0.0483843399483634
25 0.0481332519166319
26 0.0469065353349218
27 0.0466693642629384
28 0.0460200363111448
29 0.0450817746026985
30 0.0442499462607152
31 0.0438793342627151
32 0.0433779841047193
33 0.0425934721653055
34 0.0419070517922485
35 0.0413212186768942
36 0.0419930191843028
37 0.0403004221635102
38 0.0386011256007657
39 0.0377601300377255
40 0.0377679241945692
41 0.0352928217984817
42 0.0357388999091214
43 0.0354642426346856
44 0.0346813725823197
45 0.0336192462442992
46 0.0331981659297494
47 0.0332801165103702
48 0.0333354931411891
49 0.0329299291564994
50 0.0330214727061292
51 0.0315419675650296
52 0.0310897256890996
53 0.0310873795522029
54 0.0304877103270812
55 0.0304036949252119
56 0.0305371581075172
57 0.0301679996513229
58 0.0298508113288078
59 0.0297878699708544
60 0.0300299581463853
61 0.0302502510982888
62 0.0303155921207116
63 0.0303187311948911
64 0.0303074367380645
65 0.0303191684011056
66 0.0303698770089046
67 0.0304442233613199
68 0.0304921696681141
69 0.0304898782669989
70 0.0304475644966778
71 0.0303906631224483
72 0.030343574117277
73 0.0303162439208251
74 0.0303060456801959
75 0.0303054191224459
76 0.0303078006710477
77 0.0303097614277307
78 0.0303100129742352
79 0.0303085385947673
80 0.0303061837521781
81 0.030304052857687
82 0.0303029475536811
83 0.0303031525716683
84 0.0303045246780064
85 0.0303067007005015
86 0.0303092833255876
87 0.0303119486783699
88 0.0303144805758697
89 0.0303167601284506
90 0.0303187382588109
91 0.0303204082950445
92 0.0303217856996406
93 0.0303228956495328
94 0.0303237664960004
95 0.030324426750873
96 0.0303249038312848
97 0.0303252235351186
98 0.030325409770526
99 0.0303254843766183
};
\addlegendentry{Kernel SOS}
\end{axis}

\end{tikzpicture}}
    \subcaption{Dimension 2}
    \label{fig:Lorentz_rho_1_idx}
\end{minipage}
\begin{minipage}[c]{0.32\textwidth}
    \centering
    \resizebox{\textwidth}{!}{
\begin{tikzpicture}

\definecolor{darkgray176}{RGB}{176,176,176}
\definecolor{darkorange}{RGB}{255,140,0}
\definecolor{lightgray204}{RGB}{204,204,204}
\definecolor{slategray}{RGB}{112,128,144}

\begin{axis}[
legend cell align={left},
legend style={fill opacity=0.8, draw opacity=1, text opacity=1, draw=lightgray204},
tick align=outside,
tick pos=left,
x grid style={darkgray176},
xlabel={Iterations},
xmin=-9.95, xmax=208.95,
xtick style={color=black},
y grid style={darkgray176},
ylabel={Rho Relative},
ymin=0.0216346964681622, ymax=0.183487905145664,
ytick style={color=black}
]
\addplot [semithick, slategray, dashed]
table {%
0 0.176130941114868
1 0.170914884624595
2 0.165426823808362
3 0.160510299805639
4 0.157059426638431
5 0.153157576709162
6 0.147928119128802
7 0.142357205755535
8 0.135031529069235
9 0.126344012497788
10 0.117903437546419
11 0.112003722604725
12 0.108767816863989
13 0.105865864098559
14 0.103085703535059
15 0.100933617095116
16 0.099909000601184
17 0.0991781045883131
18 0.0984552995564703
19 0.097793625548695
20 0.0971695622487568
21 0.0965892627473588
22 0.0959856623593816
23 0.095457503390005
24 0.0950542882410528
25 0.0946840909995662
26 0.09433780691247
27 0.0940170929207224
28 0.0937132140341641
29 0.0934264176984176
30 0.0931563898556256
31 0.0929015022934847
32 0.0926596808799599
33 0.09242809103789
34 0.0922096582186185
35 0.0920037506928039
36 0.091808520461654
37 0.0916233375505532
38 0.0914476757672427
39 0.0912801780129557
40 0.0911210843178029
41 0.0909696019366106
42 0.0908252610176676
43 0.0906878336373435
44 0.0905569059434352
45 0.0904306178607838
46 0.0903100701964936
47 0.090197333756753
48 0.0900969318641075
49 0.0899890427355019
50 0.0898902730778741
51 0.0897848345257137
52 0.0897012027368831
53 0.0896041480485918
54 0.0895344326973573
55 0.0894322642135975
56 0.0893668619325265
57 0.0892928323775255
58 0.089226871502629
59 0.0891352530551468
60 0.0890886680923241
61 0.0889972531108654
62 0.0889456601598513
63 0.0888783880038491
64 0.0888213285220765
65 0.0887656569326969
66 0.0887304300534872
67 0.0886405840276892
68 0.0886055504483231
69 0.0885683827938764
70 0.0885152673685018
71 0.0884525741460251
72 0.0884326128826352
73 0.0883527820867122
74 0.0883210272198761
75 0.0882935398992285
76 0.0882400539456183
77 0.0881771082139107
78 0.0881733629226931
79 0.0881051194260173
80 0.0880758549764574
81 0.0880361944503347
82 0.0880253386422183
83 0.0879587854626974
84 0.0879415948123633
85 0.0879076693016962
86 0.0878721803285932
87 0.0878324333829462
88 0.0878021555005215
89 0.0877981938980452
90 0.0877436840157322
91 0.0877382628077982
92 0.0876913012509392
93 0.0876648807707705
94 0.0876422226117021
95 0.0875940239189452
96 0.0875647189925796
97 0.0875493127998342
98 0.087543882910728
99 0.0874723686349169
100 0.087472962379373
101 0.087459909966748
102 0.0874275944340834
103 0.0873955032876803
104 0.0873927198287159
105 0.0873242157786573
106 0.0873253574329192
107 0.0873188200694256
108 0.0872924791289883
109 0.0872610529473232
110 0.0872387694561342
111 0.0872084823400067
112 0.0872048698075653
113 0.0871450634146522
114 0.0871338819521665
115 0.0871219679848055
116 0.0871130970943669
117 0.0870897437191274
118 0.0870665967995432
119 0.0870366631716614
120 0.0870286301796198
121 0.0869999059686386
122 0.0869848106584298
123 0.0869594325778485
124 0.0869516894678617
125 0.0868989542233964
126 0.0869292597230451
127 0.0868845448972549
128 0.0868683309992123
129 0.0868139728574064
130 0.0868595844029065
131 0.0868163057504516
132 0.0867496561026563
133 0.0868009848125469
134 0.0867269728127468
135 0.0867484181304984
136 0.0866843988594858
137 0.0867012142105303
138 0.0866356890211036
139 0.0866219645368884
140 0.0866490204002865
141 0.0866122085510582
142 0.0865755731562851
143 0.0865672627071
144 0.0865415950441581
145 0.086541074649928
146 0.0864831076246628
147 0.0865042577380855
148 0.0864550435001603
149 0.086451702011512
150 0.0864184150710967
151 0.0864188081882206
152 0.0863655515389079
153 0.0863834802318579
154 0.0863273427960191
155 0.0863338630183558
156 0.0862702894905493
157 0.0862742827795212
158 0.0862721114153711
159 0.0862213666934117
160 0.0862353100472372
161 0.0861632167472715
162 0.0862221904428889
163 0.0860957557299193
164 0.0861233965839722
165 0.0861202868048294
166 0.0860602101003695
167 0.0860956488538168
168 0.0860470242792422
169 0.0860104433820593
170 0.086025747502688
171 0.0860010075132501
172 0.0859418946266725
173 0.0859368527675579
174 0.0859100108222165
175 0.0858722066534454
176 0.0858958470992432
177 0.0858705577728022
178 0.0858286505807039
179 0.0857751982989898
180 0.0857893677638655
181 0.0857659614959868
182 0.0857095360835265
183 0.085715896054146
184 0.0856624343740912
185 0.0856678407785885
186 0.0856164997147331
187 0.0855819354347903
188 0.0855842548865174
189 0.0855681824389392
190 0.0855205658910894
191 0.0854680195523797
192 0.0854875431354691
193 0.0854634460512391
194 0.0854161778843691
195 0.085366464527584
196 0.0853612961957175
197 0.0853107728827196
198 0.0853173664793311
199 0.0852916901324592
};
\addlegendentry{Gradient Descent}
\addplot [semithick, darkorange]
table {%
0 0.0700858152380625
1 0.0691505715725536
2 0.0680729749648448
3 0.0670381555777462
4 0.0660678418597633
5 0.0651122622215248
6 0.0641642111791533
7 0.0632824165791195
8 0.0628803590923581
9 0.0622061607904884
10 0.0612852320569094
11 0.0599739946384511
12 0.0581888861550138
13 0.0567033533063674
14 0.0538520909103684
15 0.0510402558145813
16 0.0487765718492073
17 0.0467103907952315
18 0.0448216508627853
19 0.0428137508512906
20 0.0410377310934305
21 0.0397529522332302
22 0.0387828938184697
23 0.0379085207299928
24 0.0370980284109298
25 0.036392867878879
26 0.0357448867713748
27 0.0351504842173141
28 0.0346040997592526
29 0.034099189704651
30 0.0336591440582321
31 0.0332495662000313
32 0.032839699944662
33 0.0325317112081166
34 0.0322432639739053
35 0.031942904662709
36 0.0317009487787784
37 0.0314900644255647
38 0.0312725515023462
39 0.031082531772091
40 0.0308879963658395
41 0.0307515833970639
42 0.03058742399946
43 0.0304625805219526
44 0.0303511922279331
45 0.0302304460309567
46 0.0301320327420057
47 0.0300381521715007
48 0.0299522118669417
49 0.0298705788609354
50 0.0298324136259177
51 0.0297193547607266
52 0.0296651025431249
53 0.0296104333963958
54 0.0295582225105714
55 0.0295112398520074
56 0.0294055311718633
57 0.0294367107834381
58 0.0294092874509726
59 0.0293566396095849
60 0.0293429662648602
61 0.0292928163372962
62 0.0292850429453421
63 0.0292502887387961
64 0.0292403517605837
65 0.0292070494029092
66 0.029204400076972
67 0.0291728203266436
68 0.0291748199965606
69 0.029169985506586
70 0.0291427201796255
71 0.02913575291357
72 0.0291069952795959
73 0.0291140119297256
74 0.0290888862807016
75 0.0290979921125127
76 0.0290639978896442
77 0.0290565900693028
78 0.0290688160067371
79 0.0290645620433664
80 0.0290585963461731
81 0.0290472780027551
82 0.0290000876706273
83 0.029028032107125
84 0.0290258705997434
85 0.0290285234312724
86 0.0290265096707282
87 0.0290251108484676
88 0.0290275838797878
89 0.0290222694996134
90 0.0290046987113723
91 0.0289916604989577
92 0.0289917344906706
93 0.0289952987165693
94 0.0289962488238512
95 0.0289961835165061
96 0.0289961268078824
97 0.0289957040265565
98 0.0289947186470874
99 0.0289933918512342
};
\addlegendentry{Kernel SOS}
\end{axis}

\end{tikzpicture}}
    \subcaption{Dimension 3}
    \label{fig:Lorentz_rho_2_idx}
\end{minipage}
\caption{Rho Relative on Lorentz Map Training Set.}
\label{fig:Lorentz_rho}
\end{figure}

\paragraph{\bf Discussion}
While many of the results for \emph{KSOS} are promising, we mean this paper to be an invitation to further explore global optimization methods in the context of kernel learning. We found that a lot of the benefit of \emph{KSOS} appears already in sampling the domain. That is, the sample points in $\mathcal{T}$ often perform comparable to gradient descent. \emph{KSOS} helps then extract additional performance of this random sampling by building a surrogate. However, even then, many solutions remain relatively close to the sample points, as the accuracy of such surrogate functions is strongest locally. 

We believe that further work aiming to combine local and global methods will yield the strongest results. An avenue of exploration may be a principled multi-start methodology, that is finding an appropriate distribution for parameter initialization, then using the \emph{KSOS} framework to find an appropriate starting point for gradient descent \cite{marti2003multi}. When it is difficult to formulate a distribution for the parameters, one may also focus on a Bayesian Optimization framework to better explore the parameter space \cite{frazier2018tutorial}. This is specifically useful when evaluating $\rho$ is prohibitive and hence exploring the parameter space is expensive. 

Lastly, we see value in appropriate metaheuristic algorithms, such as genetic algorithms and ant colony optimization \cite{katoch2021review, abdel2018metaheuristic, dorigo2003ant}. These methods have been shown to numerically perform well on global optimization tasks and combinatorial problems, which is crucial when limiting the number of kernels in the base kernel to reduce model complexity\cite{bh_sparse_kfs}. While some links to more principled optimization methods, such as stochastic gradient descent, exist, we note that it remains difficult to theoretically prove the observed performance of metaheuristic algorithms making numerical experiments the most accurate performance test \cite{bortz1998simplex, chada2020iterative, lengyel2023curvature, lengyel2024thesis}.


\section{Conclusion}
In this paper, we used the method of the Kernel Sum of Squares (KSOS) method as a novel global optimization approach for data-adapted kernel learning in dynamical systems. Traditional kernel-based methods, while theoretically robust and numerically efficient, often face challenges in selecting appropriate base kernels and optimizing their parameters, especially when relying on gradient-based methods prone to local minima.

\emph{KSOS} addresses these limitations by providing a global optimization framework that leverages kernel-based surrogate functions, ensuring more reliable and accurate learning of dynamical systems from data. Through extensive numerical experiments on the Logistic Map, Henon Map, and Lorentz System, we demonstrated that \emph{KSOS} consistently outperforms gradient descent in minimizing the relative-$\rho$ metric and improving the accuracy of the induced kernel.

Our results highlight the significant potential of \emph{KSOS} in forecasting the evolution of chaotic dynamical systems. By effectively adapting kernels to the underlying dynamics, \emph{KSOS} enhances the predictive capabilities and robustness of kernel-based methods. The approach also allows for tighter control over function evaluations, making it computationally efficient for large datasets.

The theoretical foundations laid out in this paper, combined with the promising empirical results, suggest that \emph{KSOS} can be a powerful tool for researchers and practitioners working with time series data across various scientific fields. Future work will focus on further refining the \emph{KSOS} algorithm, exploring its applications in other complex systems and particularly the database of 135 chaotic systems that were explored in other papers by Hamzi and Owhadi and their collaborators, and integrating it with other machine learning techniques to broaden its applicability and effectiveness.

In summary, the Kernel Sum of Squares method represents a significant step forward in the quest for more accurate and reliable kernel learning for dynamical systems, offering a robust alternative to traditional gradient-based optimization methods.

\section{Acknowledgment} 
HO acknowledges support from the Air Force Office of Scientific Research under MURI award number FA9550-20-1-0358 (Machine Learning and Physics-Based Modeling and Simulation) and the Department of Energy under the MMICCs SEA-CROGS award.  BH acknowledges support from the Air Force Office of Scientific Research (award number  FA9550-21-1-0317) and the Department of Energy (award number SA22-0052-S001). HO is grateful for support from a Department of Defense Vannevar Bush Faculty
Fellowship.
\appendix 
\section{Appendix}

\subsection{Reproducing Kernel Hilbert Spaces (RKHS)}

We give a brief overview of reproducing kernel Hilbert spaces as used in statistical learning
theory ~\cite{CuckerandSmale}. Early work developing
the theory of RKHS was undertaken by N. Aronszajn~\cite{aronszajn50reproducing}.

\begin{definition} Let  ${\mathcal H}$  be a Hilbert space of functions on a set ${\mathcal X}$.
	Denote by $\langle f, g \rangle$ the inner product on ${\mathcal H}$   and let $\|f\|= \langle f, f \rangle^{1/2}$
	be the norm in ${\mathcal H}$, for $f$ and $g \in {\mathcal H}$. We say that ${\mathcal H}$ is a reproducing kernel
	Hilbert space (RKHS) if there exists a function $k:{\mathcal X} \times {\mathcal X} \rightarrow \mathbf{R}$
	such that\\
	\begin{itemize}
		\item[i.] $k_x:=k(x,\cdot)\in{\mathcal{H}}$ for all $x\in{\mathcal{H}}$.
		\item[ii.] $k$ spans ${\mathcal H}$: ${\mathcal H}=\overline{\mbox{span}\{k_x~|~x \in {\mathcal X}\}}$.
		\item[iii.] $k$ has the {\em reproducing property}:
		$\forall f \in {\mathcal H}$, $f(x)=\langle f,k_x \rangle$.
	\end{itemize}
	$k$ will be called a reproducing kernel of ${\mathcal H}$. ${\mathcal H}_k$  will denote the RKHS ${\mathcal H}$
	with reproducing kernel $k$ where it is convenient to explicitly note this dependence.
\end{definition}

The important properties of reproducing kernels are summarized in the following proposition.
\begin{proposition}\label{prop1} If $k$ is a reproducing kernel of a Hilbert space ${\mathcal H}$, then\\
	\begin{itemize}
		\item[i.] $k(x,y)$ is unique.
		\item[ii.]  $\forall x,y \in {\mathcal X}$, $k(x,y)=k(y,x)$ (symmetry).
		\item[iii.] $\sum_{i,j=1}^q\alpha_i\alpha_j k(x_i,x_j) \ge 0$ for $\alpha_i \in \mathbf{R}$, $x_i \in \mathcal{X}$ and $q\in\mathcal{N}_+$
		(positive definiteness).
		\item[iv.] $\langle k(x,\cdot),k(y,\cdot) \rangle=K(x,y)$.
	\end{itemize}
\end{proposition}
Common examples of reproducing kernels defined on a compact domain $\mathcal{X} \subset \mathrm{R}^n$ are the 
(1) constant kernel: $K(x,y)= m > 0$
(2) linear kernel: $k(x,y)=x\cdot y$
(3) polynomial kernel: $k(x,y)=(1+x\cdot y)^d$ for $d \in \mathbf{N}_+$
(4) Laplace kernel: $k(x,y)=e^{-||x-y||_2/\sigma^2}$, with $\sigma >0$
(5)  Gaussian kernel: $k(x,y)=e^{-||x-y||^2_2/\sigma^2}$, with $\sigma >0$
(6) triangular kernel: $k(x,y)=\max \{0,1-\frac{||x-y||_2^2}{\sigma} \}$, with $\sigma >0$.
(7) locally periodic kernel: $k(x,y)=\sigma^2 e^{-2 \frac{ \sin^2(\pi ||x-y||_2/p)}{\ell^2}}e^{-\frac{||x-y||_2^2}{2 \ell^2}}$, with $\sigma, \ell, p >0$.

\begin{theorem} \label{thm1}
	Let $k:\mathcal{X} \times {\mathcal X} \rightarrow \mathbf{R}$ be a symmetric and positive definite function. Then there
	exists a Hilbert space of functions $\mathcal{H}$ defined on $\mathcal{X}$   admitting $k$ as a reproducing Kernel.
	Conversely, let  $\mathcal{X}$  be a Hilbert space of functions $f: \mathcal{X} \rightarrow \mathbf{R}$ satisfying
	$\forall x \in \mathcal{X}, \exists \kappa_x>0,$ such that $|f(x)| \le \kappa_x \|f\|_{\mathcal{H}},
	\quad \forall f \in \mathcal{H}. $
	Then $\mathcal{H}$ has a reproducing kernel $k$.
\end{theorem}


\begin{theorem}\label{thm4}
	Let $k(x,y)$ be a positive definite kernel on a compact domain or a manifold $X$. Then, there exists a Hilbert
	space $\mathcal{F}$  and a function $\Phi: X \rightarrow \mathcal{F}$ such that
	$$k(x,y)= \langle \Phi(x), \Phi(y) \rangle_{\mathcal{F}} \quad \mbox{for} \quad x,y \in X.$$
	$\Phi$ is called a feature map, and $\mathcal{F}$ a feature space\footnote{The dimension of the feature space can be infinite, for example, in the case of the Gaussian kernel.}.
\end{theorem}

\subsection{Function Approximation in RKHSs: An Optimal Recovery Viewpoint} 
In this section, we review function approximation in RKHSs from the point of view of optimal recovery as discussed in \cite{owhadi_scovel_2019}. 

\paragraph{Problem {\bf P}:} Given input/output data $(x_1, y_1),\cdots , (x_N , y_N ) \in \mathcal{X} \times \mathbb{R}$,  recover an unknown function $u^{\ast}$ mapping $\mathcal{X}$ to $\mathbb{R}$ such that
$u^{\ast}(x_i)=y_i$ for $i \in \{1,...,N\}$.

In the setting of optimal recovery, \cite{owhadi_scovel_2019}  Problem {\bf P} can be turned into a well-posed problem by restricting candidates for $u$ to belong to a Banach space of functions $\mathcal{B}$ endowed with a norm $||\cdot||$ and identifying the optimal recovery as the minimizer of the relative error

\begin{equation} \label{game}
    \mbox{min}_v\mbox{max}_u \frac{||u-v||^2}{||u||^2}, 
\end{equation} 
where the max is taken over $u \in \mathcal{B}$ and the min is taken over candidates in $v \in \mathcal{B}$ such that $v(x_i)=u(x_i)=y_i$. For the validity of the constraints $u(x_i) = y_i$,  $\mathcal{B}^{\ast}$, the dual space of $\mathcal{B}$, must contain delta Dirac functions $\phi_i(\cdot)=\delta(\cdot-x_i)$. This problem can be stated as a game between Players I and II and can then be represented as
  \begin{equation}\label{eqdkjdhkjhffORgameban}
\text{\xymatrixcolsep{0pc}\xymatrix{
\text{(Player I)} & u\ar[dr]_{\max}\in \mathcal{B}    &      &v\ar[ld]^{\min}\in L(\Phi,\mathcal{B}) &\text{(Player II)}\\
&&\frac{\|u-v(u)\|}{\|u\|}\,.& &
}}\,
\end{equation}

If $||\cdot||$ is quadratic, i.e. $||u||^2=[Q^{-1}u,u] $ where $[\phi, u]$ stands for the duality product between $\phi \in \mathcal{B}^{\ast}$ and $u \in \mathcal{B}$ and $Q : \mathcal{B}^{\ast}\rightarrow \mathcal{B}$ is a positive symmetric linear bijection (i.e. such that $[\phi, Q \phi] \ge  0$ and $[\psi, Q \phi ] = [\phi, Q \psi]$ for $\phi,\psi \in \mathcal{B}^{\ast} $). In that case, the optimal solution of (\ref{game}) has the explicit form 
\begin{equation}\label{sol_rep}v^{\ast}=\sum_{i,j=1}^{N}u(x_i) A_{i,j} Q \phi_j, \end{equation}
where   $A=\Theta^{-1}$ and $\Theta \in \RR^{N \times N}$ is a Gram matrix with entries $\Theta_{i,j}=[\phi_i,Q\phi_j]$.

To recover the classical representer theorem, one defines the reproducing kernel $K$ as $$K(X,y)=[\delta(\cdot-x),Q\delta(\cdot-y)]$$ 
In this case, $(\mathcal{B},||\cdot ||)$ can be seen as an RKHS endowed with the norm
$$||u||^2=\mbox{sup}_{\phi \in \mathcal{B}^\ast}\frac{(\int \phi(x) u(x) dx)^2}{(\int \phi(x) K(X,y) \phi(y) dx dy)}$$
and (\ref{sol_rep}) corresponds to the classical representer theorem 
\begin{equation}\label{eqkjelkjefffhb}
v^{\ast}(\cdot) = y^T AK(X,\cdot),
\end{equation} 
 using the vectorial notation $y^T AK(X,\cdot)=\sum_{i,j=1}^{N}y_iA_{i,j}K(X_j,\cdot)$ with $y_i=u(x_i)$, $A=\Theta^{-1}$ and $\Theta_{i,j} =K(X_i,x_j)$.
  
 Now, let us consider the problem of learning the kernel from data. As introduced in \cite{Owhadi19}, the method of KFs is based on the premise that \emph{a kernel is good if there is no significant loss in accuracy in the prediction error if the number of data points is halved}. This led to the introduction of 
 \[\rho=\frac{||v^{\ast}-v^{s} ||^2}{||v^{\ast} ||^2} \]
  which is the relative error between 
  $v^\ast$, the optimal recovery \eqref{eqkjelkjefffhb} of $u^\ast$ based on the full dataset
  $X=\{(x_1,y_1),\ldots,(x_N,y_N)\}$, and
  $v^s$  the optimal recovery  of both $u^\ast$ and $v^\ast$ based on half of the dataset $ X^s=\{(x_i,y_i)\mid i \in \mathcal{S}\}$ ($\operatorname{Card}(\mathcal{S})=N/2$) which admits the representation
  \begin{equation}
v^s=(y^s)^T A^s K(X^s,\cdot)
  \end{equation}
 with $y^s=\{y_i\mid i \in \mathcal{S}\}$,
 $x^s=\{x_i\mid i \in \mathcal{S}\}$,
 $A^s=(\Theta^s)^{-1}$, $\Theta^s_{i,j}=K(X_i^s,x_j^s)$.
 This quantity  $\rho$ is directly related to the game in (\ref{eqdkjdhkjhffORgameban}) where one is minimizing the relative error of $v^{\ast}$ versus $v^s$. 
Instead of using the entire the dataset $X$ one may use random subsets $X^{s_1}$ (of $X$) for $v^{\ast}$ and random subsets $ X^{s_2}$ (of $X^{s_1}$) for $v^s$.  In practice, it is computed as   \cite{Owhadi19}

\begin{equation} \label{rho_scalar_s1_s2}
\rho=1-\frac{{Y_{s_2}}^TK(X^{s_2},X^{s_2})^{-1}Y_{s_2}}{{Y_{s_1}}^TK(X^{s_1},X^{s_1})^{-1}Y_{s_1}}
\end{equation}

Writing $\sigma^2(x)=K(X,x)-K(X,X^f)K(X^f,X^f)^{-1}K(X^f,x)$ we have the pointwise error bound
\begin{equation}\label{error_estimate} |u(x)-v^\ast(x)| \leq  \sigma(x) \|u\|_{\Hc},\end{equation}
 Local error estimates such as (\ref{error_estimate}) are
classical in Kriging \cite{Wu92localerror} (see also \cite{owhadi2015bayesian}[Thm. 5.1] for applications to PDEs). $\|u\|_{\Hc}$ is bounded from below (and, in with sufficient data, can be approximated by) by $\sqrt{Y^{f,T} K(X^f,X^f)^{-1} Y^f} $, i.e., the RKHS norm of the interpolant of $v^\ast$.

\section{Different \texorpdfstring{$\rho$}{rho} functions corresponding to different versions of KFs}

In previous work, we introduced different variants of kernel flows
\begin{itemize}
\item \emph{Lyapunov Exponents based Kernel Flows} and the premise that a kernel is good if there is no significant loss in accuracy if half of the data is used to estimate the maximal Lyapunov exponent from data\footnote{A similar principle can be used with other Lyapunov exponents.}. The following metric is minimized  $$\rho_{L}=|\lambda_{\mbox{max},N}- \lambda_{\mbox{max},N/2}| $$
The goal here is to learn a dynamical system that has similar long term behavior as the underlying system from which the data is coming.

\item \emph{Maximum Mean Discrepancy (MMD-) based Kernel Flows} and the premise that a kernel is good if there is no significant loss in accuracy when estimating the MMD when two different samples, $S_1$ and $S_2$, of the time series are used and  minimize  
$$
  \rho_{ \scriptstyle  \mbox{MMD}}=\mbox{MMD}(S_1,S_2)  
$$
The goal here is to capture the statistical properties of the underlying dynamical system in the spirit of what is done through the Frobenius-Perron operator.

\item \emph{Sparse Kernel Flows:} For an additive base kernel of the form 	$$
		K_{\beta,\theta} (x,y)= \sum_{i=1}^{m} \theta^2_i k_i(x,y;\beta),
$$
(where the $k_i(x,y;\beta)$ are kernels)
we consider the $L_1$ regularization
$$
	\mathcal{L}(\beta,\theta) = 
	\arg\min\limits_{\beta,\theta} \bigg(  1 - \frac{y^\top_c K^{-1}_{\beta,\theta} y_c}{y^\top_b K^{-1}_{\beta,\theta} y_b} +
	{\lambda}\|\theta\|_1 \bigg)
$$ in order to sparsify the base kernel and set as many $\theta_i$ to zero.

\item \emph{Hausdorff-Metric Kernel Flows (HMKFs)}

When the system has an attractor, we will consider a kernel of the form\footnote{For kernels of this form with $m>2$, the method of Kernel Flows can be viewed as a problem of data compression in the context of Algorithmic Information Theory (AIT) \cite{bh_huter_kfs_ait}.}
$$K_{\beta,\theta} (x,y)= \sum_{i=1}^{m} \theta^2_i k_i(x,y;\beta)$$
and find the parameters $\theta$ and $\beta$ using the following metric    $$
  \rho_{\mbox{HD}}=\mbox{HD}(\mathcal{A}_N,\mathcal{A}_{N/2})   $$
  instead of (\ref{kf_original}). This metric is the Hausdorff distance between the attractor reconstruction with $N$ points and its reconstruction with $N/2$ points. We will call this method \emph{Regular Hausdorff metric-based Kernel Flows}.  To improve the performance of this method, we will combine it with the method of Sparse Kernel Flows that we introduced in  \cite{bh_sparse_kfs}, and we will also consider training the kernel by minimizing the following metric 
   $$ 
  \rho_{\mbox{HD}}=  \mbox{HD}({\mathcal{A}}_N,\mathcal{A}_{N/2}) + {\lambda}\|\theta\|_1  $$
with respect to $\beta$, $\theta$. We will call this approach 	\emph{Sparse Hausdorff metric-based Kernel Flows}.  

 \end{itemize}

\bibliographystyle{plain} 
\bibliography{ref_kernel_sos}

\def\cprime{$'$}
\begin{thebibliography}{10}

\bibitem{abarbanel2012analysis}
H.~Abarbanel.
\newblock {\em Analysis of Observed Chaotic Data}.
\newblock Institute for Nonlinear Science. Springer New York, 2012.

\bibitem{abdel2018metaheuristic}
Mohamed Abdel-Basset, Laila Abdel-Fatah, and Arun~Kumar Sangaiah.
\newblock Metaheuristic algorithms: A comprehensive review.
\newblock {\em Computational intelligence for multimedia big data on the cloud
  with engineering applications}, pages 185--231, 2018.

\bibitem{ahmadi2018sum}
Amir~Ali Ahmadi.
\newblock Sum of squares (sos) techniques: an introduction.
\newblock {\em Princeton, Princeton, NJ, USA, Tech. Rep}, pages 1--9, 2018.

\bibitem{ALEXANDER2020132520}
Romeo Alexander and Dimitrios Giannakis.
\newblock Operator-theoretic framework for forecasting nonlinear time series
  with kernel analog techniques.
\newblock {\em Physica D: Nonlinear Phenomena}, 409:132520, 2020.

\bibitem{aronszajn50reproducing}
N.~Aronszajn.
\newblock Theory of reproducing kernels.
\newblock {\em Transactions of the American Mathematical Society},
  68(3):337--404, 1950.

\bibitem{bortz1998simplex}
David~Matthew Bortz and Carl~Tim Kelley.
\newblock The simplex gradient and noisy optimization problems.
\newblock In {\em Computational Methods for Optimal Design and Control:
  Proceedings of the AFOSR Workshop on Optimal Design and Control Arlington,
  Virginia 30 September--3 October, 1997}, pages 77--90. Springer, 1998.

\bibitem{5706920}
Jake Bouvrie and Boumediene Hamzi.
\newblock Balanced reduction of nonlinear control systems in reproducing kernel
  hilbert space.
\newblock In {\em 2010 48th Annual Allerton Conference on Communication,
  Control, and Computing (Allerton)}, pages 294--301, 2010.

\bibitem{bh12}
Jake Bouvrie and Boumediene Hamzi.
\newblock Empirical estimators for stochastically forced nonlinear systems:
  Observability, controllability and the invariant measure.
\newblock {\em Proc. of the 2012 American Control Conference}, pages 294--301,
  2012.
\newblock \url{https://arxiv.org/abs/1204.0563v1}.

\bibitem{bh17}
Jake Bouvrie and Boumediene Hamzi.
\newblock Kernel methods for the approximation of nonlinear systems.
\newblock {\em SIAM J. Control and Optimization}, 2017.
\newblock \url{https://arxiv.org/abs/1108.2903}.

\bibitem{hb17}
Jake Bouvrie and Boumediene Hamzi.
\newblock Kernel methods for the approximation of some key quantities of
  nonlinear systems.
\newblock {\em Journal of Computational Dynamics}, 1, 2017.
\newblock \url{http://arxiv.org/abs/1204.0563}.

\bibitem{Sindy}
Steven~L. Brunton, Joshua~L. Proctor, and J.~Nathan Kutz.
\newblock Discovering governing equations from data by sparse identification of
  nonlinear dynamical systems.
\newblock {\em Proceedings of the National Academy of Sciences},
  113(15):3932--3937, 2016.

\bibitem{CASDAGLI1989}
Martin Casdagli.
\newblock Nonlinear prediction of chaotic time series.
\newblock {\em Physica D: Nonlinear Phenomena}, 35(3):335 -- 356, 1989.

\bibitem{chada2020iterative}
Neil~K Chada, Yuming Chen, and Daniel Sanz-Alonso.
\newblock Iterative ensemble kalman methods: A unified perspective with some
  new variants.
\newblock {\em arXiv preprint arXiv:2010.13299}, 2020.

\bibitem{survey_kf_ann}
Ashesh Chattopadhyay, Pedram Hassanzadeh, Krishna~V. Palem, and Devika
  Subramanian.
\newblock Data-driven prediction of a multi-scale lorenz 96 chaotic system
  using a hierarchy of deep learning methods: Reservoir computing, ann, and
  {RNN-LSTM}.
\newblock {\em CoRR}, abs/1906.08829, 2019.

\bibitem{chen2021solving}
Yifan Chen, Bamdad Hosseini, Houman Owhadi, and Andrew~M. Stuart.
\newblock Solving and learning nonlinear {{PDEs}} with {{Gaussian}} processes.
\newblock {\em Journal of Computational Physics}, 447:110668, December 2021.

\bibitem{cheney2009course}
Elliott~Ward Cheney and William~Allan Light.
\newblock {\em A course in approximation theory}, volume 101.
\newblock American Mathematical Soc., 2009.

\bibitem{CuckerandSmale}
Felipe Cucker and Steve Smale.
\newblock On the mathematical foundations of learning.
\newblock {\em Bulletin of the American Mathematical Society}, 39:1--49, 2002.

\bibitem{bhkfnp}
M.~Darcy, B.~Hamzi, J.~Susiluoto, A.~Braverman, and H.~Owhadi.
\newblock Learning dynamical systems from data: a simple cross-validation
  perspective, part {II}: nonparametric kernel flows.
\newblock {\em Physica D}, 444:133583, 2023.

\bibitem{bhkfsdes}
Matthieu Darcy, Boumediene Hamzi, Giulia Livieri, Houman Owhadi, and Peyman
  Tavallali.
\newblock One-shot learning of stochastic differential equations with data
  adapted kernels.
\newblock {\em Physica D: Nonlinear Phenomena}, 444:133583, 2023.

\bibitem{dorigo2003ant}
Marco Dorigo and Thomas St{\"u}tzle.
\newblock The ant colony optimization metaheuristic: Algorithms, applications,
  and advances.
\newblock {\em Handbook of metaheuristics}, pages 250--285, 2003.

\bibitem{bhcm1}
B.Haasdonk {,}B.Hamzi {,} G.Santin~{,} D.Wittwar.
\newblock Kernel methods for center manifold approximation and a weak
  data-based version of the center manifold theorems.
\newblock {\em Physica D}, 2021.

\bibitem{frazier2018tutorial}
Peter~I Frazier.
\newblock A tutorial on bayesian optimization.
\newblock {\em arXiv preprint arXiv:1807.02811}, 2018.

\bibitem{lyap_bh}
P.~Giesl, B.~Hamzi, M.~Rasmussen, and K.~Webster.
\newblock Approximation of {L}yapunov functions from noisy data.
\newblock {\em Journal of Computational Dynamics}, 2019.
\newblock \url{https://arxiv.org/abs/1601.01568}.

\bibitem{yk4}
R.~González-García, R.~Rico-Martínez, and I.G. Kevrekidis.
\newblock Identification of distributed parameter systems: A neural net based
  approach.
\newblock {\em Computers \& Chemical Engineering}, 22:S965--S968, 1998.
\newblock European Symposium on Computer Aided Process Engineering-8.

\bibitem{yk3}
Ove Grandstrand.
\newblock Nonlinear system identification using neural networks: dynamics and
  instanbilities.
\newblock In A.~B. Bulsari, editor, {\em Neural Networks for Chemical
  Engineers}, chapter~16, pages 409--442. Elsevier, Elsevier, 1995.

\bibitem{bhcm11}
B.~Haasdonk, B.~Hamzi, G.~Santin, and D.~Wittwar.
\newblock Greedy kernel methods for center manifold approximation.
\newblock {\em Proc. of ICOSAHOM 2018, International Conference on Spectral and
  High Order Methods}, (1), 2018.
\newblock \url{https://arxiv.org/abs/1810.11329}.

\bibitem{bh_huter_kfs_ait}
B~Hamzi, M~Hutter, and O~Owhadi.
\newblock A note on learning kernels from data from an algorithmic information
  theoretic point of view, 2023.

\bibitem{hamzi2019kernel}
Boumediene Hamzi and Fritz Colonius.
\newblock Kernel methods for the approximation of discrete-time linear
  autonomous and control systems.
\newblock {\em SN Applied Sciences}, 1(7):674, July 2019.

\bibitem{boumedienehamzi2022note}
Boumediene Hamzi, Amirhossein Jafarian, Houman Owhadi, and L{\'e}o Paillet.
\newblock A {{Note}} on {{Microlocal Kernel Design}} for {{Some Slow-Fast
  Stochastic Differential Equations}} with {{Critical Transitions}} and
  {{Application}} to {{EEG Signals}}.
\newblock {\em Physica D: Nonlinear Phenomena}, 2022.

\bibitem{mmd_kernels_bh}
Boumediene Hamzi, Christian Kuehn, and Sameh Mohamed.
\newblock A note on kernel methods for multiscale systems with critical
  transitions.
\newblock {\em Mathematical Methods in the Applied Sciences}, 42(3):907--917,
  2019.

\bibitem{BHPhysicaD}
Boumediene Hamzi and Houman Owhadi.
\newblock Learning dynamical systems from data: A simple cross-validation
  perspective, part i: Parametric kernel flows.
\newblock {\em Physica D: Nonlinear Phenomena}, 421:132817, 2021.

\bibitem{bh_kfs_missing}
Boumediene Hamzi, Houman Owhadi, and Yannis Kevrekidis.
\newblock Learning dynamical systems from data: A simple cross-validation
  perspective, part iv: Case with partial observations.
\newblock {\em Physica D: Nonlinear Phenomena}, 454:133853, 2023.

\bibitem{yk1}
J.L. Hudson, M.~Kube, R.A. Adomaitis, I.G. Kevrekidis, A.S. Lapedes, and R.M.
  Farber.
\newblock Nonlinear signal processing and system identification: applications
  to time series from electrochemical reactions.
\newblock {\em Chemical Engineering Science}, 45(8):2075--2081, 1990.

\bibitem{kantz97}
Holger Kantz and Thomas Schreiber.
\newblock {\em Nonlinear Time Series Analysis}.
\newblock Cambridge University Press, USA, 1997.

\bibitem{katoch2021review}
Sourabh Katoch, Sumit~Singh Chauhan, and Vijay Kumar.
\newblock A review on genetic algorithm: past, present, and future.
\newblock {\em Multimedia tools and applications}, 80:8091--8126, 2021.

\bibitem{bh2020b}
Stefan Klus, Feliks Nuske, and Boumediene Hamzi.
\newblock Kernel-based approximation of the koopman generator and
  schr{\"o}dinger operator.
\newblock {\em Entropy}, 22, 2020.
\newblock \url{https://www.mdpi.com/1099-4300/22/7/722}.

\bibitem{klus2020data}
Stefan Klus, Feliks N{\"u}ske, Sebastian Peitz, Jan-Hendrik Niemann, Cecilia
  Clementi, and Christof Sch{\"u}tte.
\newblock Data-driven approximation of the koopman generator: Model reduction,
  system identification, and control.
\newblock {\em Physica D: Nonlinear Phenomena}, 406:132416, 2020.

\bibitem{bhks}
Andreas Bittracher{,} Stefan Klus{,} Boumediene Hamzi{,}~Peter Koltai{,} and
  Christof Schutte.
\newblock Dimensionality reduction of complex metastable systems via kernel
  embeddings of transition manifold, 2019.
\newblock https://arxiv.org/abs/1904.08622.

\bibitem{Lasserre}
J.~B. Lasserre.
\newblock A theorem of the alternative in {B}anach lattices.
\newblock {\em Proceedings of the American Mathematical Society}, pages
  189--194, 1998.

\bibitem{lee2021learning}
Jonghyeon Lee, Edward~De Brouwer, Boumediene Hamzi, and Houman Owhadi.
\newblock Learning dynamical systems from data: A simple cross-validation
  perspective, part iii: Irregularly-sampled time series, 2021.

\bibitem{lengyel2024thesis}
Daniel Lengyel.
\newblock {\em Optimal Sample Set Selection for the Simplex Gradient}.
\newblock PhD thesis, Imperial College London, 2024.
\newblock PhD thesis, not yet uploaded to arXiv.

\bibitem{lengyel2023curvature}
Daniel Lengyel, Panos Parpas, Nikolas Kantas, and Nicholas~R Jennings.
\newblock Curvature aligned simplex gradient: Principled sample set
  construction for numerical differentiation.
\newblock {\em arXiv preprint arXiv:2310.12712}, 2023.

\bibitem{marteau2020non}
Ulysse Marteau-Ferey, Francis Bach, and Alessandro Rudi.
\newblock Non-parametric models for non-negative functions.
\newblock {\em Advances in neural information processing systems},
  33:12816--12826, 2020.

\bibitem{marti2003multi}
Rafael Mart{\'\i}.
\newblock Multi-start methods.
\newblock {\em Handbook of metaheuristics}, pages 355--368, 2003.

\bibitem{nemirovski2004interior}
Arkadi Nemirovski.
\newblock Interior point polynomial time methods in convex programming.
\newblock {\em Lecture notes}, 42(16):3215--3224, 2004.

\bibitem{nielsen2019practical}
A.~Nielsen.
\newblock {\em Practical Time Series Analysis: Prediction with Statistics and
  Machine Learning}.
\newblock O'Reilly Media, 2019.

\bibitem{hamzimaulikowhadi}
Boumediene Hamzi {,} Romit Maulik{,}~Houman Owhadi.
\newblock Simple, low-cost and accurate data-driven geophysical forecasting
  with learned kernels.
\newblock {\em Proceedings of the Royal Society A: Mathematical, Physical and
  Engineering Sciences}, 477(2252), 2021.

\bibitem{owhadi_scovel_2019}
H.~Owhadi and C.~Scovel.
\newblock {\em Operator-Adapted Wavelets, Fast Solvers, and Numerical
  Homogenization: from a game theoretic approach to numerical approximation and
  algorithm design}.
\newblock Cambridge Monographs on Applied and Computational Mathematics.
  Cambridge University Press, 2019.

\bibitem{Owhadi19}
H.~Owhadi and G.~R. Yoo.
\newblock Kernel flows: From learning kernels from data into the abyss.
\newblock {\em Journal of Computational Physics}, 389:22--47, 2019.

\bibitem{owhadi2015bayesian}
Houman Owhadi.
\newblock Bayesian numerical homogenization.
\newblock {\em Multiscale Modeling \& Simulation}, 13(3):812--828, 2015.

\bibitem{houman_cgc}
Houman Owhadi.
\newblock Computational graph completion.
\newblock {\em Research in the Mathematical Sciences}, 9(2):27, 2022.

\bibitem{jaideep1}
Jaideep Pathak, Zhixin Lu, Brian~R. Hunt, Michelle Girvan, and Edward Ott.
\newblock Using machine learning to replicate chaotic attractors and calculate
  lyapunov exponents from data.
\newblock {\em Chaos: An Interdisciplinary Journal of Nonlinear Science},
  27(12):121102, 2017.

\bibitem{yk2}
Ramiro Rico-Martinez, K~Krischer, IG~Kevrekidis, MC~Kube, and JL~Hudson.
\newblock Discrete-vs. continuous-time nonlinear signal processing of cu
  electrodissolution data.
\newblock {\em Chemical Engineering Communications}, 118(1):25--48, 1992.

\bibitem{rudi2024finding}
Alessandro Rudi, Ulysse Marteau-Ferey, and Francis Bach.
\newblock Finding global minima via kernel approximations.
\newblock {\em Mathematical Programming}, pages 1--82, 2024.

\bibitem{santinhaasdonk19}
Gabriele Santin and Bernard Haasdonk.
\newblock Kernel methods for surrogate modeling.
\newblock 2019.
\newblock \url{https://arxiv.org/abs/1907.105566}.

\bibitem{smirnov2022mean}
Alexandre Smirnov, Boumediene Hamzi, and Houman Owhadi.
\newblock Mean-field limits of trained weights in deep learning: A dynamical
  systems perspective.
\newblock {\em Dolomites Research Notes on Approximation}, 15(3), 2022.

\bibitem{stengle1974nullstellensatz}
Gilbert Stengle.
\newblock A nullstellensatz and a positivstellensatz in semialgebraic geometry.
\newblock {\em Mathematische Annalen}, 207:87--97, 1974.

\bibitem{wright2001convergence}
Stephen~J Wright.
\newblock On the convergence of the newton/log-barrier method.
\newblock {\em Mathematical programming}, 90:71--100, 2001.

\bibitem{Wu92localerror}
Zongmin Wu and Robert Schaback.
\newblock Local error estimates for radial basis function interpolation of
  scattered data.
\newblock {\em {IMA} J. Numer. Anal}, 13:13--27, 1992.

\bibitem{hmkfs}
Lu~Yang, Boumediene Hamzi, Yannis Kevrekidis, Houman Owhadi, Xiuwen Sun, and
  Naiming Xie.
\newblock Learning dynamical systems from data: A simple cross-validation
  perspective, part vi: Hausdorff metric based training of kernels to learn
  attractors with application to 133 chaotic dynamical systems.
\newblock {\em Physica D: Nonlinear Phenomena}, 2024.

\bibitem{bh_sparse_kfs}
Lu~Yang, Xiuwen Sun, Boumediene Hamzi, Houman Owhadi, and Naiming Xie.
\newblock Learning dynamical systems from data: A simple cross-validation
  perspective, part v: Sparse kernel flows for 132 chaotic dynamical systems.
\newblock {\em Physica D: Nonlinear Phenomena}, 460:134070, 2024.

\bibitem{bh_hamiltonian}
Jalalian Yasamin, Bounediene Hamzi, Houman Owhadi, Tavallali Peyman, and Samir
  Moustafa.
\newblock Learning dynamical systems from data: A simple cross-validation
  perspective, part vii: Hamiltonian systems.
\newblock 2023.

\bibitem{yoo2021deep}
Gene~Ryan Yoo and Houman Owhadi.
\newblock Deep regularization and direct training of the inner layers of
  {{Neural Networks}} with {{Kernel Flows}}.
\newblock {\em Physica D: Nonlinear Phenomena}, 426:132952, November 2021.

\end{thebibliography}

\end{document}